\documentclass[]{style/iromlab}

\DeclareDocumentEnvironment{example}{}{\noindent\textbf{Running example:}\itshape}{}

\usepackage{ifthen}
\newboolean{include-notes}
\newboolean{include-new}
\newboolean{include-remove}
\setboolean{include-notes}{true}
\setboolean{include-new}{false}
\setboolean{include-remove}{false}

\usepackage[dvipsnames]{xcolor}
\usepackage[normalem]{ulem}
\newcommand{\justin}[1]{\ifthenelse{\boolean{include-notes}}{\textcolor{orange}{\textbf{Jaime:} #1}}{}}

\newcommand{\princeton}[1]{\ifthenelse{\boolean{include-notes}}{\textcolor{orange}{#1}}{}}

\usepackage{multicol}
\usepackage{amsmath, amsfonts, amssymb, amsthm}
\usepackage{enumerate}
\usepackage[inline]{enumitem}
\usepackage{mathtools}
\usepackage{graphicx}
\usepackage{longtable,tabularx}
\usepackage{placeins} 
\usepackage{float}
\usepackage{multirow}
\usepackage{bbm}
\usepackage{threeparttable}
\usepackage{balance}
\usepackage[ruled,algo2e]{algorithm2e}
\usepackage{algpseudocode}
\usepackage{adjustbox}
\usepackage{booktabs}
\usepackage{bm}
\usepackage{etoolbox}
\usepackage{microtype}
\usepackage[title]{appendix}
\usepackage{units}
\usepackage{cleveref}
\usepackage{xspace}
\usepackage{tcolorbox}
\usepackage{caption}

\definecolor{claude_color}{HTML}{F89E62}
\definecolor{deepseek_color}{HTML}{78B6E8}
\definecolor{o3_mini_color}{HTML}{6CD5A1}
\definecolor{red_color}{HTML}{E13B55}
\definecolor{prompt_color}{HTML}{71502B}

\tcbset{
  promptstyle/.style={
    colback=gray!10,
    colframe=gray!60,
    boxrule=0.5pt,
    arc=2pt,
    outer arc=2pt,
    left=4pt,
    right=4pt,
    top=4pt,
    bottom=4pt,
    fonttitle=\bfseries,
    sharp corners=south,
  }
}

\tcbset{
  promptstyle_prompt_iuq/.style={
    colback=prompt_color!10,
    colframe=prompt_color!60,
    coltitle=white,
    boxrule=1.0pt,
    arc=2pt,
    outer arc=2pt,
    left=4pt,
    right=4pt,
    top=4pt,
    bottom=4pt,
    fonttitle=\bfseries,
  }
}

\tcbset{
  promptstyle_claude/.style={
    colback=claude_color!10,
    colframe=claude_color!60,
    coltitle=black,
    boxrule=1.0pt,
    arc=2pt,
    outer arc=2pt,
    left=4pt,
    right=4pt,
    top=4pt,
    bottom=4pt,
    fonttitle=\bfseries,
  }
}

\tcbset{
  promptstyle_deepseek/.style={
    colback=deepseek_color!10,
    colframe=deepseek_color!60,
    coltitle=black,
    boxrule=1.0pt,
    arc=2pt,
    outer arc=2pt,
    left=4pt,
    right=4pt,
    top=4pt,
    bottom=4pt,
    fonttitle=\bfseries,
  }
}

\tcbset{
  promptstyle_o3_mini/.style={
    colback=o3_mini_color!10,
    colframe=o3_mini_color!60,
    coltitle=black,
    boxrule=1.0pt,
    arc=2pt,
    outer arc=2pt,
    left=4pt,
    right=4pt,
    top=4pt,
    bottom=4pt,
    fonttitle=\bfseries,
  }
}

\newbool{extended}
\setbool{extended}{false}

\makeatletter
\newcommand{\longdash}[1][2em]{%
  \makebox[#1]{$\m@th\smash-\mkern-7mu\cleaders\hbox{$\mkern-2mu\smash-\mkern-2mu$}\hfill\mkern-7mu\smash-$}}
\makeatother
\newcommand{\omitskip}{\kern-\arraycolsep}

\author[1*]{Zhiting Mei}
\author[1]{Christina Zhang}
\author[1]{Tenny Yin}
\author[1]{Justin Lidard}
\author[1*]{Ola Shorinwa}
\author[1]{Anirudha Majumdar}

\affiliation[1]{Princeton University}

\contribution[*]{Equal contribution.}

\begin{document}

\title{Reasoning about Uncertainty: \\
{\LARGE Do Reasoning Models Know When They Don't Know?}}

\abstract{
Reasoning language models have set state-of-the-art (SOTA) records on many challenging benchmarks,
enabled by multi-step reasoning induced using reinforcement learning. However, like previous language models, reasoning models are prone to generating confident, plausible responses that are incorrect (hallucinations). Knowing when and how much to trust these models is critical to the safe deployment of reasoning models in real-world applications. To this end, we explore uncertainty quantification~(UQ) of reasoning models in this work. Specifically, we ask three fundamental questions: First, \emph{are~\mbox{reasoning}~models well-calibrated?} 
Second, \emph{does deeper reasoning improve model calibration?}
Finally, inspired by humans' innate ability to double-check their thought processes to verify the validity of their answers and their confidence, we ask: \emph{can reasoning models improve their calibration by explicitly reasoning about their chain-of-thought traces?}
We introduce \emph{introspective uncertainty quantification} (IUQ) to explore this direction.
In extensive evaluations on SOTA reasoning models across a broad range of benchmarks, we find that reasoning models: (i) are typically overconfident, with self-verbalized confidence estimates often greater than $85\%$ particularly for incorrect responses, (ii) become even more overconfident with deeper reasoning, and (iii) can become better calibrated through introspection (e.g., o3-Mini and DeepSeek R1) but not uniformly (e.g., Claude 3.7 Sonnet becomes more poorly calibrated).
Lastly, we conclude with important research directions to design necessary UQ benchmarks and improve the calibration of reasoning models.

}

\keywords{
Reasoning about Uncertainty, Reasoning Models, LLMs, Uncertainty Quantification.
}

\maketitle

\begin{figure*}[th]
    \centering
    \includegraphics[width=\linewidth]{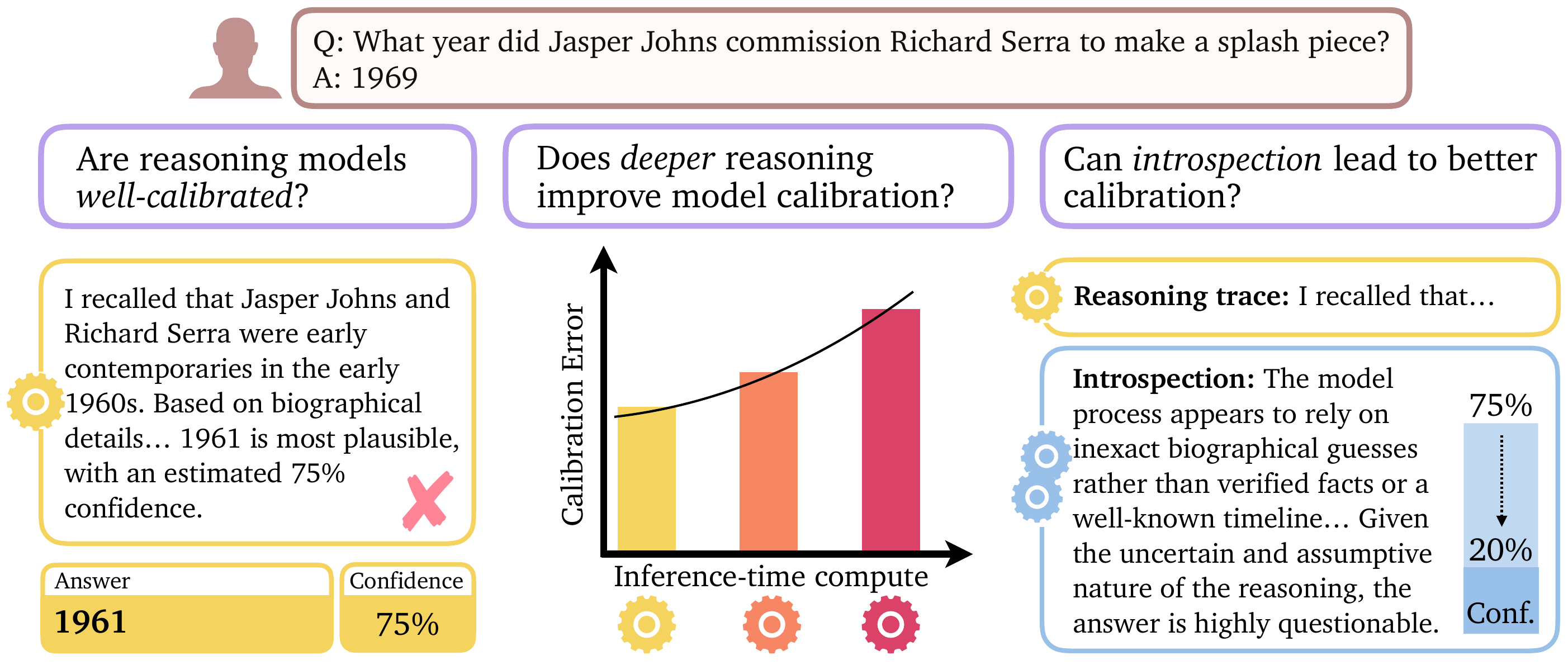}
    \caption{To examine if reasoning models know when they don't know, we ask three fundamental questions: (i) are reasoning models well-calibrated? (ii) does deeper reasoning improve model calibration? and (iii) can introspection improve calibration? We find that reasoning models are typically overconfident and become more overconfident with deeper reasoning. However, through introspection, reasoning models can provide better calibrated confidence estimates.}
    \label{fig:uq_for_reasoning_models}
\end{figure*}

\section{Introduction}
\label{sec:introduction}
Recent breakthroughs in large language models (LLMs) and vision-language models (VLMs) have largely been driven by \emph{reasoning} language models---language models that are trained to perform multi-step reasoning through reinforcement learning, e.g., Claude 3.7 \cite{anthropicthec3_7}, o3-Mini \cite{openaio3mini}, Gemini 2 Flash Thinking \cite{team2024gemini}, and DeepSeek R1 \cite{guo2025deepseek}. By breaking down complex tasks into subtasks and analyzing and synthesizing responses from the resulting subtasks, reasoning models have achieved new state-of-the-art (SOTA) performance on a broad range of challenging real-world tasks, such as programming, e.g., Codeforces~\cite{Codeforces}; math, e.g., AIME~\cite{AoPSWikiAIME}; engineering and science, e.g., GPQA~\cite{rein2023gpqa}; and arts, e.g., MMLU~\cite{hendrycks2020measuring}, outperforming domain experts on many benchmarks.

Despite their remarkable capabilities, SOTA reasoning language models often generate confident but factually-incorrect responses to questions, known as \emph{hallucinations} \cite{bai2024hallucination, liu2024survey}, similar to their non-reasoning counterparts. 
Consequently, the safe integration of reasoning models into real-world tasks requires knowing when and how much to trust these models, i.e., uncertainty quantification (UQ) of these models.
Prior work has examined UQ of non-reasoning models, showing that non-reasoning models are poorly calibrated in general \cite{xiong2023can} but can become better calibrated through chain-of-thought prompting \cite{liang2024introspective}.
However, UQ of reasoning models remains an unexplored frontier. To address this knowledge gap, we examine uncertainty quantification of reasoning models to better evaluate their trustworthiness.

Specifically, in this work,  we ask three critical questions (illustrated in \Cref{fig:uq_for_reasoning_models}).
First, \emph{are reasoning models well-calibrated?} We evaluate the calibration of reasoning models across a broad range of benchmarks
to examine underconfidence/overconfidence in these models across different problem domains. 
Our findings reveal that \emph{reasoning models are typically overconfident,} often expressing confidence estimates greater than $85\%$ (in increments of $5\%$) even when they are incorrect. 
Given that overconfidence is often masked by high accuracy, we find that reasoning models are well-calibrated in knowledge-retrieval tasks and poorly calibrated in tasks requiring domain expertise, e.g., graduate-level research domains. Moreover, our results highlight that prompt engineering is significantly less effective with reasoning models, unlike non-reasoning models.

Second, \emph{does deeper reasoning improve model calibration?} Prior work \cite{openaio3mini} has shown that reasoning models can achieve higher accuracy through inference-time scaling of reasoning effort, measured by the number of available reasoning tokens. Here, we explore the effects of deeper reasoning on model calibration across low, medium, and high amounts of reasoning. Our evaluations reveal that \emph{reasoning models become more overconfident with deeper reasoning.} Specifically on wrongly-answered questions, reasoning models become significantly more overconfident as we scale the reasoning effort.

Lastly, humans often double-check their problem-solving approach to verify their answers and estimate their confidence in the validity of the answers \cite{hewitt2016double}. In line with this observation, we ask the question: \emph{can reasoning models improve their calibration by explicitly reasoning about their chain-of-thought traces?}
We introduce \emph{introspective uncertainty quantification} to explore this question. Introspective UQ utilizes a multi-stage reasoning paradigm to estimate the confidence in the accuracy of a given response: a model first reasons about a given question, and then a second model reasons about the chain of thought and final answer produced by the first model in order to assess its confidence in the answer. We find that 
\emph{more critical introspection improves the calibration of reasoning models, e.g., o3-Mini and DeepSeek R1, although the calibration of Claude 3.7 Sonnet worsens.}

\section{Related Work}
\label{sec:related_work}
\smallskip
\noindent\textbf{Reasoning in LLMs.}
Prior work \cite{wei2022chain, kojima2022large, wang2022self} has demonstrated that language models can achieve significantly higher accuracy by generating intermediate reasoning steps, known as a chain of thought (CoT). Chain-of-thought prompting enables remarkable zero-shot or few-shot learning in LLMs, enabling generalist LLMs to outperform task-specific, finetuned models~\cite{wei2022chain}. However, the complexity of the few-shot CoT examples significantly influences the induced accuracy of the models \cite{fu2022complexity}, spurring research on prompt engineering for effective chain-of-thought reasoning \cite{zhang2022automatic, huang2022towards, hao2024llm}. 
Subsequent work introduces multi-path reasoning strategies to enable backtracking on incorrect reasoning paths: e.g., Tree-of-Thought prompting~\cite{yao2023tree} and Graph-of-Thoughts prompting~\cite{besta2024graph}, which construct parallel reasoning traces; least-to-most prompting~\cite{zhou2022least}, which solves a series of subproblems in sequence; and reasoning via planning~\cite{hao2023reasoning}, which performs Monte Carlo Tree Search on a reasoning tree to find the optimal reasoning path. 
However, existing approaches require domain expertise to construct effective chain-of-thought prompts/examples, posing a challenge.
By training language models to reason using reinforcement learning (RL), recent work \cite{jaech2024openai, guo2025deepseek, anthropicthec3_7, openaio3mini, team2024gemini, qwq32b} has shown that reasoning language models can solve complex tasks more accurately than non-reasoning LLMs. Through inference-time scaling, reasoning models can be made to think deeper to break down complex problems into multiple steps, achieving SOTA performance on many benchmarks \cite{anthropicthec3_7, openaio3mini, team2024gemini, guo2025deepseek}. However, the effects of reasoning on the calibration of confidence estimates of these models remains unknown, motivating uncertainty quantification of reasoning models.

\smallskip
\noindent\textbf{Uncertainty Quantification of LLMs.}
Uncertainty quantification of LLMs has gained notable prominence given the tendency of LLMs to hallucinate. In general, existing UQ methods for LLMs eschew traditional techniques such as Bayesian networks~\cite{jospin2022hands} due to their computational cost, given that LLMs often have billions to trillions of parameters. 
UQ methods for LLMs can be grouped into two broad classes: \emph{white-box} and \emph{black-box} methods~\cite{10.1145/3744238}. White-box methods utilize the internal outputs of LLMs \cite{geva2020transformer, azaria2023internal}, such as tokens for entropy/perplexity-based UQ~\cite{manakul2023selfcheckgpt, fadeeva2024fact, ling2024uncertainty, xiao2021hallucination, kuhn2023semantic} or neural activations for mechanistic interpretability (MI) ~\cite{belinkov2022probing, dunefsky2024transcoders, yun2021transformer, gao2024scaling, tamkin2023codebook, ferrando2024know}. While these UQ metrics can be well calibrated for pre-trained models~\cite{kadavath2022language}, reinforcement learning from human feedback (RLHF) leads to poorly calibrated token-level UQ~\cite{openai2024gpt4technicalreport} since RL-training objectives are not proper scoring rules~\cite{gneiting2007strictly}.
In contrast to white-box methods, black-box methods estimate uncertainty without access to the model's internal outputs which might not be available, given the notable shift from fully open-source language models to open-weight or closed-source models, e.g., \cite{hurst2024gpt, anthropicthec3_7}. In fact, state-of-the-art closed-source reasoning models generally do not provide access to the models' logits, which is required for token-based, semantic similarity-based, or MI-based UQ.
Consequently, self-verbalized UQ methods \cite{mielke2022reducing, smith2020controlling, tanneru2024quantifying, aichberger2024semantically, zhang2024r}, which directly prompt an LLM for its confidence in natural language, have become increasingly important.
In contrast to prior work which has focused exclusively on non-reasoning models, we assess the uncertainty of reasoning models without finetuning, using self-verbalized UQ methods given the lack of access to the models' internal outputs.

\section{{Preliminaries}}
\label{sec:preliminaries}

Consider a %
model $\mathcal M$ that takes an input random variable $X\in\mathcal X$ and predicts an output ${\hat Y\in \mathcal Y}$, along with a confidence score $\hat P\in[0, 1]$. %
We focus on settings with a deterministic ground-truth $Y$ given the input $X$.\footnote{This applies to a wide range of question-and-answering problems, e.g., multiple-choice questions, true-or-false questions, or any question that has an explicit correct answer. 
The definition of $\hat Y=Y$ is not constrained to exact correspondence in form, i.e., $\hat Y=Y$ if $\hat{Y}$ and $Y$ have the same meaning. For example, both ``true'' and ``True'' are correct for a true-or-false question, if $Y$=True, while for freeform Q\&A, $\hat Y$ is correct if it captures the ground truth~\cite{wei2024measuring}. If the ground truth answer $Y$ is inherently stochastic, then it should be expressed as a stochastic distribution described in natural language, for example, a Gaussian distribution with $\mu=0$ and $\sigma^2=1$.}
In this section, we provide a brief overview of calibration of uncertainty and calibration metrics. %

\smallskip
\noindent\textbf{Calibration of Uncertainty.}
The model $\mathcal M$ is defined to be perfectly calibrated if its confidence estimate $\hat P$ matches the true probability $p$ that the associated answer $\hat Y$ is correct. Formally, perfect calibration~\cite{guo2017calibration} can be expressed as:
\begin{equation}
    \label{eq:prob_calibration}
    \mathbb {P}_{\mathcal Z}[\hat Y = {Y} \mid \hat{P} = p] = p,
\end{equation}
where the probability is over the joint distribution on $\mathcal Z=\mathcal X\times\mathcal Y$. However, achieving perfect calibration is practically impossible, motivating metrics that capture deviation from~\eqref{eq:prob_calibration} empirically.

\smallskip
\noindent\textbf{Calibration Metrics.} Calibration metrics evaluate how closely a model's predicted confidence aligns with its true accuracy. Widely-used metrics include the expected calibration error (ECE) and maximum calibration error (MCE). 
The ECE measures the average absolute difference between predicted confidence (right-hand side of \eqref{eq:prob_calibration})  and empirical accuracy (left-hand side of \eqref{eq:prob_calibration}) with: 
\begin{equation}
    \label{eq:ece}
    \mathbb{E}_{\hat{P}}\left[|\mathbb{P}[Y = \hat{Y} \mid \hat{P} = p] - p|\right].
\end{equation}
 Since this expectation is generally intractable, the ECE is typically approximated by grouping the empirical confidence estimates into $M$ equal-width bins. Given a set of samples $B_m$ whose confidence falls into bin $m$ with the bin average accuracy $\text{acc}(B_m)$ and average confidence $\text{conf}(B_m)$, the ECE is approximated by:
\begin{equation}
    \label{eq:eceapprox}
    \text{ECE} = \sum_{m=1}^M\frac{|B_m|}{n}\left|\text{acc}(B_m)-\text{conf}(B_m)\right|,
\end{equation}
where $n$ represents the total number of samples across all bins.

On the other hand, the MCE captures the worst-case discrepancy between confidence and accuracy, i.e., the maximum deviation between the left- and right-hand sides of (\ref{eq:prob_calibration}): 
\begin{equation}
    \label{eq:mce} 
    \max_{p\in[0,1]}\left|\mathbb{P}[Y=\hat{Y} \mid \hat{P} = p] - p\right|.
\end{equation}
Similar to the ECE, we approximate the MCE by taking the largest bin-wise error~\cite{naeini2015obtaining}:
\begin{equation}
    \text{MCE} = \max_{m \in \{1,..., M\}}\left|\text{acc}(B_m) - \text{conf}(B_m)\right|.
\end{equation}

\section{Method}
\label{sec:method}
In order to effectively explore uncertainty quantification in reasoning language models, we consider a broad range of prompting strategies to examine the ability of reasoning models to accurately express their confidence. We draw insights from existing work on prompt engineering to identify potentially effective prompt designs for reasoning models. Beyond the established prompt strategies, we examine the effects of \emph{introspection} on a reasoning model's ability to quantify its uncertainty. We call this approach \emph{introspective uncertainty quantification}. We discuss these strategies in the subsequent subsections.

\subsection{Prompt Strategies}
\label{sec:prompt_strategies}
We design prompts for self-verbalized uncertainty quantification of reasoning models following prompt templates for LLMs introduced in prior work \cite{xiong2023can}. Specifically, we consider zero-shot basic prompting, chain-of-thought prompting \cite{wei2022chain, kojima2022large}, multi-step prompting \cite{xiong2023can}, and Top-K prompting \cite{tian2023just}. We emphasize that the LLM is prompted zero-shot, i.e., without any examples. In \Cref{tab:method_prompt_strategies}, we provide a template for each of these prompts to illustrate their composition, with the full prompts available in Appendix~\ref{sec:app_evaluations_prompts}. %

\begin{table*}[th]
	\centering
	\caption{Prompt Strategies for LLMs.}
	\label{tab:method_prompt_strategies}
		{\begin{tabular}{l  l }
				\toprule
                    Strategy & Prompt \\
                    \midrule
                    Basic & Provide an answer
                    and the confidence in your answer between 0 and 100. \\
                    Chain-of-Thought & Analyze step by step
                    and provide an answer
                    and confidence. \\
                    Multi-Step & Break down the problem into multiple steps, 
                    each with your confidence. \\
                    Top-K & Give $K$ best guesses and your confidence in each guess. \\
				\bottomrule
		\end{tabular}}
\end{table*}

\smallskip
\noindent\textbf{Basic Prompt Strategy.}
In the basic prompt strategy, we provide the question directly to the LLM and ask for its answer to the question. For example, in \Cref{tab:method_prompt_strategies}, we ask the LLM to provide its answer and confidence without any description of the problem-solving approach. 

\smallskip
\noindent\textbf{Chain-of-Thought Prompt Strategy.}
Prior work \cite{wei2022chain, kojima2022large, zhang2022automatic} has shown that LLMs are highly capable zero-shot reasoners when asked to \emph{think step by step}. Although prior work has demonstrated the effectiveness of chain-of-thought prompting in improving accuracy, the effects of chain-of-thought prompting on calibration of self-verbalized UQ estimates remain relatively unexplored. We explore this research direction in this paper. Specifically, we ask the LLM to \emph{analyze} the question \emph{step by step} to induce intermediate reasoning and ask for its confidence, as shown in \Cref{tab:method_prompt_strategies}.

\smallskip
\noindent\textbf{Multi-Step Prompt Strategy.}
The multi-step prompt strategy takes chain-of-thought prompting a step further. In the multi-step prompt strategy, we ask the LLM not only to reason about the problem, but to also provide its confidence at each step. By doing so, we hope to induce the LLM to reason about its confidence to improve the calibration of its self-verbalized confidence estimates. As shown in \Cref{tab:method_prompt_strategies}, we ask the LLM to \emph{break down} the problem into \emph{multiple steps} and provide its answer and confidence in each step.

\smallskip
\noindent\textbf{Top-K Prompt Strategy.}
Identifying candidate answers to a given question (and in particular, how likely these answers are to be correct) can result in better estimation of the model's confidence, as demonstrated in prior work \cite{tian2023just}, where the LLM is asked to provide its top-K guesses for the answer to a question and the associated confidence. We explore this prompt strategy for reasoning models in our work.

\begin{figure*}[th]
    \centering
    \includegraphics[width=\linewidth]{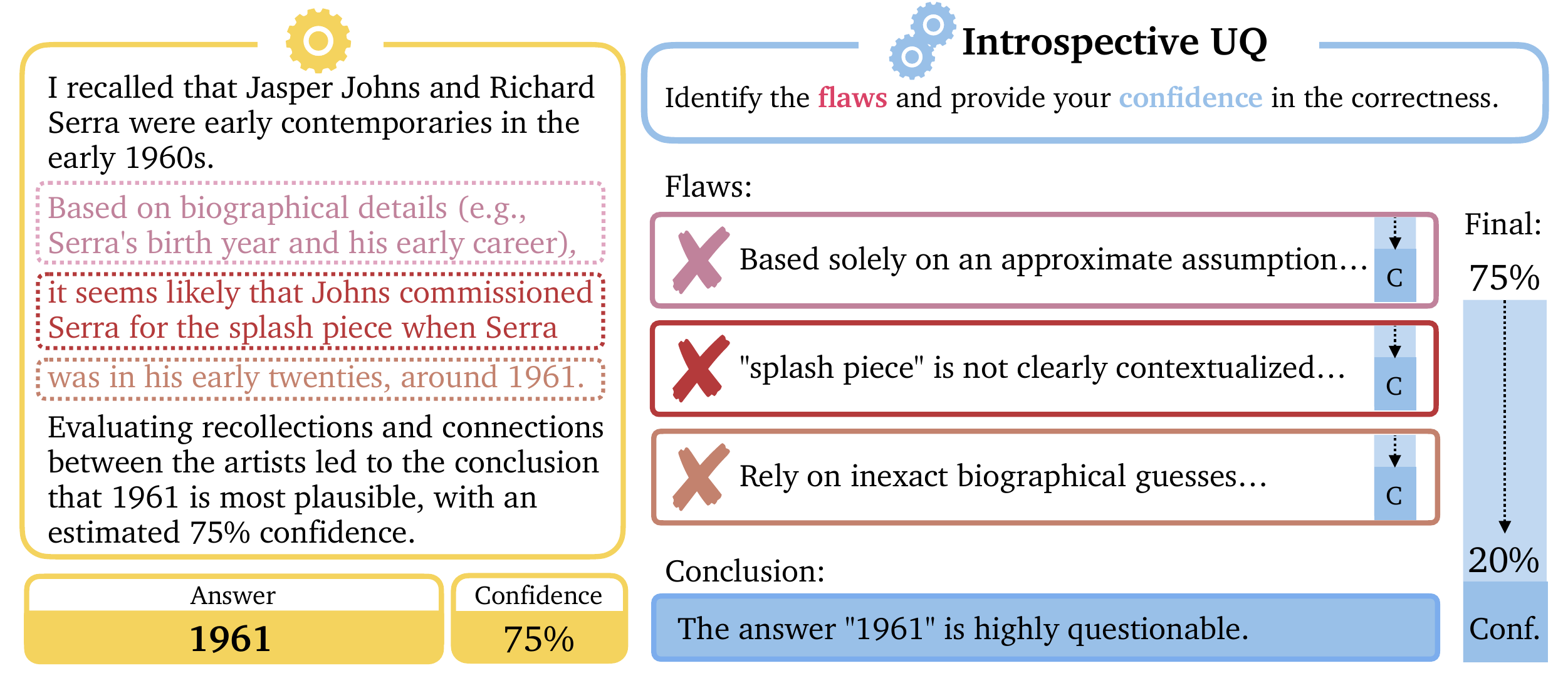}
    \caption{\textbf{Introspective UQ.}
    Although the reasoning model is initially highly overconfident and wrong (as shown in \Cref{fig:uq_for_reasoning_models}), through introspection, the model can identify flaws in its chain-of-thought traces to provide better calibrated confidence estimates, mitigating overconfidence. 
    }
    \label{fig:introspective_uq_method}
\end{figure*}

\subsection{Introspective Uncertainty Quantification}
\label{sec:introspective_uq}
We explore the effect of reasoning on the calibration of uncertainty quantification of reasoning models (an open question in LLM research). 
Drawing inspiration from how humans double-check their thought processes to ascertain their confidence in a given response, we introduce \emph{introspective uncertainty quantification}: a self-verbalized UQ method which utilizes a multi-stage reasoning architecture to reason about and update the confidence of reasoning models. 
\Cref{fig:introspective_uq_method} illustrates the two-stage introspective UQ procedure using the question shown in \Cref{fig:uq_for_reasoning_models}, where a reasoning model is asked to identify flaws in the prior reasoning trace and to provide its confidence on the correctness of the first answer. The model identifies weaknesses in the chain-of-thought traces and correctly reduces the confidence in the original response, ultimately improving its calibration.

At the initial stage of introspective UQ, we ask a reasoning model to answer a question and provide its confidence in the answer. In subsequent stages, we provide the reasoning model with the reasoning traces at earlier stages and ask the model to analyze these traces to identify flaws and to ultimately provide a confidence estimate in the answer provided in the initial stage. Specifically in the second stage, we consider three kinds of prompts for introspection defined by varying conservativeness~---~(i)~\emph{IUQ-Low}: a neutral prompt asking the model to reason about its uncertainty given earlier reasoning traces and associated confidence estimates (without any hints on possible flawed reasoning), (ii)~\mbox{\emph{IUQ-Medium}}: a more \emph{conservative} prompt explicitly asking the model to find flaws in the input reasoning traces including the prior confidence estimates, and (iii)~\emph{IUQ-High}: the most \emph{conservative} prompt which utilizes the same prompt as \emph{IUQ-Medium} without the prior confidence estimates. We provide a template of these prompts in Appendix~\Cref{sec:app_evaluations_introspective_uq}. Although introspective UQ can utilize many stages, we limit the number of stages to two in our work and leave an exhaustive examination of the optimal number of stages to future work.

We restrict the second stage of introspective UQ to \emph{reasoning about uncertainty} and do not ask the model to update the answer to the original question for two key reasons: (i) to encourage focused reflection on prior reasoning traces, and (ii) to guard against hallucinations, especially in problems where the model is not confident. Prior work \cite{liu2024survey} has shown that language models tend to hallucinate or vacillate between different choices when prompted with a question that they are not confident about, e.g., in knowledge-retrieval problems. Such vacillations present unnecessary distractions, undermining the effectiveness of introspective UQ.

With introspective UQ, we seek to elicit human-like internal reflections in reasoning models for calibrated uncertainty quantification, which could be important in downstream applications such as hallucination detection and mitigation. However, we note that humans sometimes become unjustifiably more confident in their responses after analyzing their reasoning process \cite{shynkaruk2006confidence}, a challenge that could also arise with reasoning models. We explore these questions in the experiments discussed in \Cref{sec:evaluation}.

\section{Experiments}
\label{sec:evaluation}
We evaluate the calibration and accuracy of reasoning models on benchmark datasets and assess the confidence estimates of reasoning models using the self-verbalized UQ methods presented in~\Cref{sec:method}.
Specifically, we seek to answer the following questions:
\begin{enumerate}
    \item Are reasoning models calibrated, systemically underconfident, or overconfident?
    \item Is accuracy correlated with calibration (positively or negatively, strongly or weakly)?
    \item Does deeper reasoning (e.g., greater number of reasoning steps) improve calibration?
    \item Can reasoning models reason about their uncertainty for better calibration?
\end{enumerate}
Lastly, we explore the effects of different prompt strategies on the calibration and accuracy of reasoning models.

\subsection{Evaluation Setup}
\label{sec:evaluation_setup}
We summarize the experiment setup, including the reasoning models, datasets, prompt strategies, and evaluation metrics. 

\smallskip
\noindent\textbf{Models.}
We evaluate SOTA reasoning models that are publicly accessible via an API. Specifically, we consider the following LLMs: (i) Claude 3.7 Sonnet (\emph{claude-3-7-sonnet-20250219}, released 02/19/2025), (ii) DeepSeek R1 (\emph{deepseek-reasoner}, released 01/20/2025), (iii) Gemini (\emph{gemini-2.0-flash-thinking-exp}, released 12/19/2024), and (iv) o3-mini (\emph{o3-mini}, released 01/31/2025). We do not benchmark Grok 3 Beta, since we were not able to get access to the closed beta.

\smallskip
\noindent\textbf{Datasets.}
We consider five standard question-answering benchmark datasets spanning a wide variety of challenging tasks, including arithmetic, humanities, science, and professional knowledge problems. We evaluate the models on the following datasets: (i)~AI2 Reasoning Challenge (\emph{ARC-Challenge}) Set \cite{clark2018think}, the challenging subset of the ARC dataset; (ii)~Graduate-Level Google-Proof Q\&A (\emph{GPQA}) Benchmark \cite{rein2023gpqa}, consisting of questions determined to be challenging by PhD-level domain experts; (iii)~Measuring Massive Multitask Language Understanding (\emph{MMLU}) dataset \cite{hendrycks2020measuring}, which encompasses a broad range of fields; (iv)~\emph{StrategyQA} \cite{geva2021did}, which requires implicit reasoning steps, and (v)~\emph{SimpleQA} \cite{wei2024measuring}, which examines the factuality of LLMs. While SimpleQA consists of open-ended questions, all other datasets use the multiple-choice format. For adequate coverage of a range of tasks in MMLU, we select the abstract algebra, professional accounting, professional medicine, international law, and sociology subsets, representing the more challenging subsets.

\smallskip
\noindent\textbf{Prompt Strategies and Metrics.}
We utilize the prompt strategies discussed in \Cref{sec:prompt_strategies}, namely: (i) basic prompting, (ii) chain-of-thought prompting, (iii) multi-step prompting, and (iv) top-K prompting (with ${K = 5}$). For each strategy, we quantify the calibration of the self-verbalized confidence estimates from the model using the expected calibration error (ECE) and maximum calibration error (MCE), introduced in~\Cref{sec:preliminaries}. We additionally present the accuracy and the successful query-completion rate (completion) to show how well the models follow instructions.
Lastly, we visualize the reliability diagram associated with the estimated confidence scores to visually examine the distribution of the confidence scores. The reliability diagram visualizes the deviation between the left-hand aside and right-hand side of \eqref{eq:prob_calibration}, showing the deviation between the bin-wise confidence and accuracy which is indicative of model underconfidence or overconfidence. %

\subsection{Calibration of Reasoning Models}
\label{ssec:calibration_reasoning_models}
\begin{figure*}[th]
    \centering
    \includegraphics[width=\linewidth]{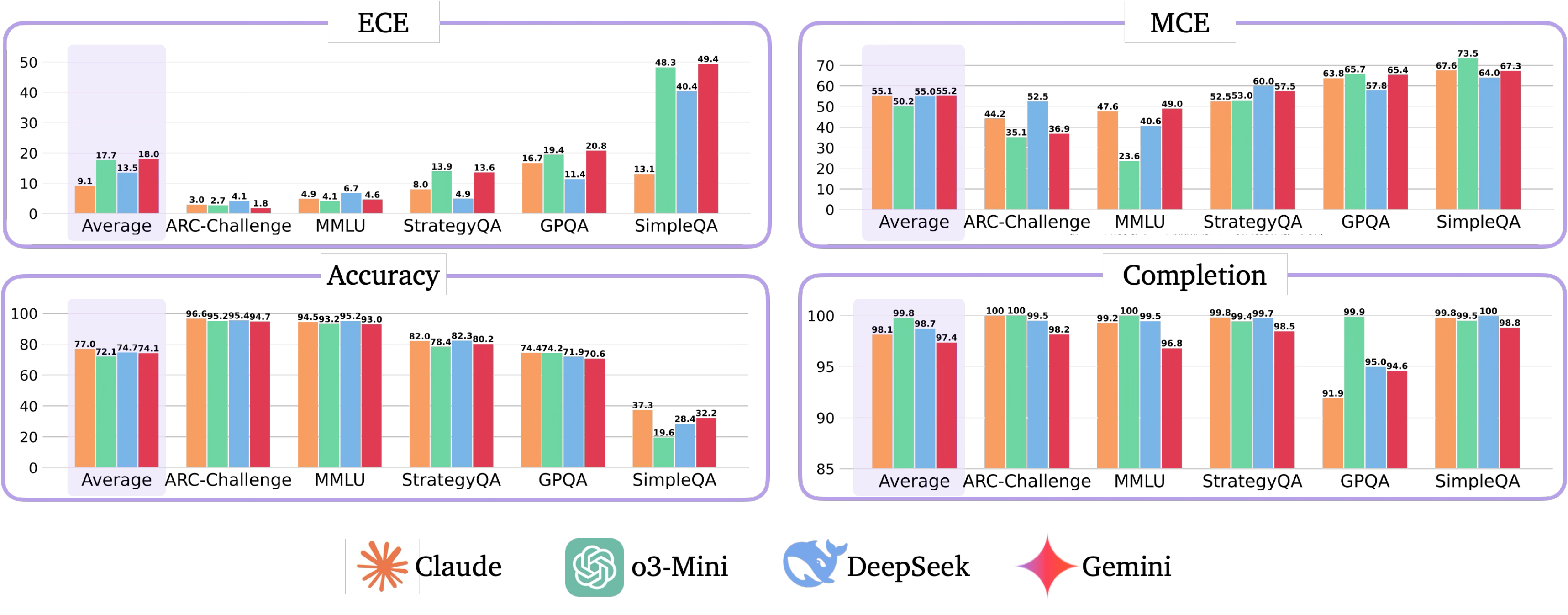}
    \caption{\textbf{Are reasoning models calibrated?} Reasoning models are well-calibrated on the \emph{ARC-Challenge} and \emph{MMLU} benchmarks and poorly calibrated on \emph{StrategyQA}, \emph{GPQA}, and \emph{SimpleQA}, suggesting better model calibration on older benchmarks and worse calibration on more recent benchmarks.
    }
    \label{fig:calibration_reasoning_models}
\end{figure*}

Here, we assess the uncertainty of reasoning models to characterize their calibration across the aforementioned benchmarks.
\Cref{fig:calibration_reasoning_models} summarizes the ECE, MCE, accuracy, and completion rate of each model across these problem domains, showing the average performance of each model over the four prompt strategies. We provide each model's individual performance for each prompt strategy in Appendix~\ref{sec:app_evaluations_calib_reasoning_models}.
In general, we find that reasoning models are not well-calibrated, particularly in challenging problem domains (e.g., SimpleQA). For example, the original GPT-4 without RLHF was better calibrated with an ECE of $0.7$ on a subset of the MMLU dataset \cite{openai2024gpt4technicalreport}, which is significantly lower than the ECE of all the models in \Cref{fig:calibration_reasoning_models}. Specifically, reasoning models tend to be overconfident, although this issue is less obvious on benchmarks where models achieve near-perfect accuracy (e.g., ARC-Challenge). Next, we provide a detailed discussion of these results.

\smallskip
\noindent\textbf{Model Calibration on Benchmarks.}
From \Cref{fig:calibration_reasoning_models}, reasoning models are better calibrated on the ARC-Challenge and MMLU benchmarks; however, calibration of these models degrades significantly on StrategyQA, GPQA, and SimpleQA. These results suggest that reasoning models are better calibrated on older benchmarks, e.g., ARC-Challenge (released in 2018) and MMLU (released in 2020), compared to more recent benchmarks where the models are poorly calibrated, e.g., StrategyQA (released in 2021), GPQA (released in 2023), and SimpleQA (released in 2024). We hypothesize that the difference in calibration of these models can be explained by the difference in accuracy across these benchmarks and explore these connections in \Cref{ssec:acc_vs_calib_reasoning_models}.
The results highlight that calibration on a particular dataset is not always indicative of calibration on another dataset, especially in cases with near-perfect model accuracy where overconfidence is difficult to detect. This finding raises a common, important challenge in benchmarking LLMs: \emph{existing benchmarks become relatively uninformative for UQ as new reasoning models are trained to beat these benchmarks, necessitating the introduction of newer benchmarks.}
Next, we characterize the nature of the calibration of reasoning models, i.e., underconfidence vs. overconfidence of these models.

\begin{figure*}[th]
    \centering
    \includegraphics[width=0.8\linewidth]{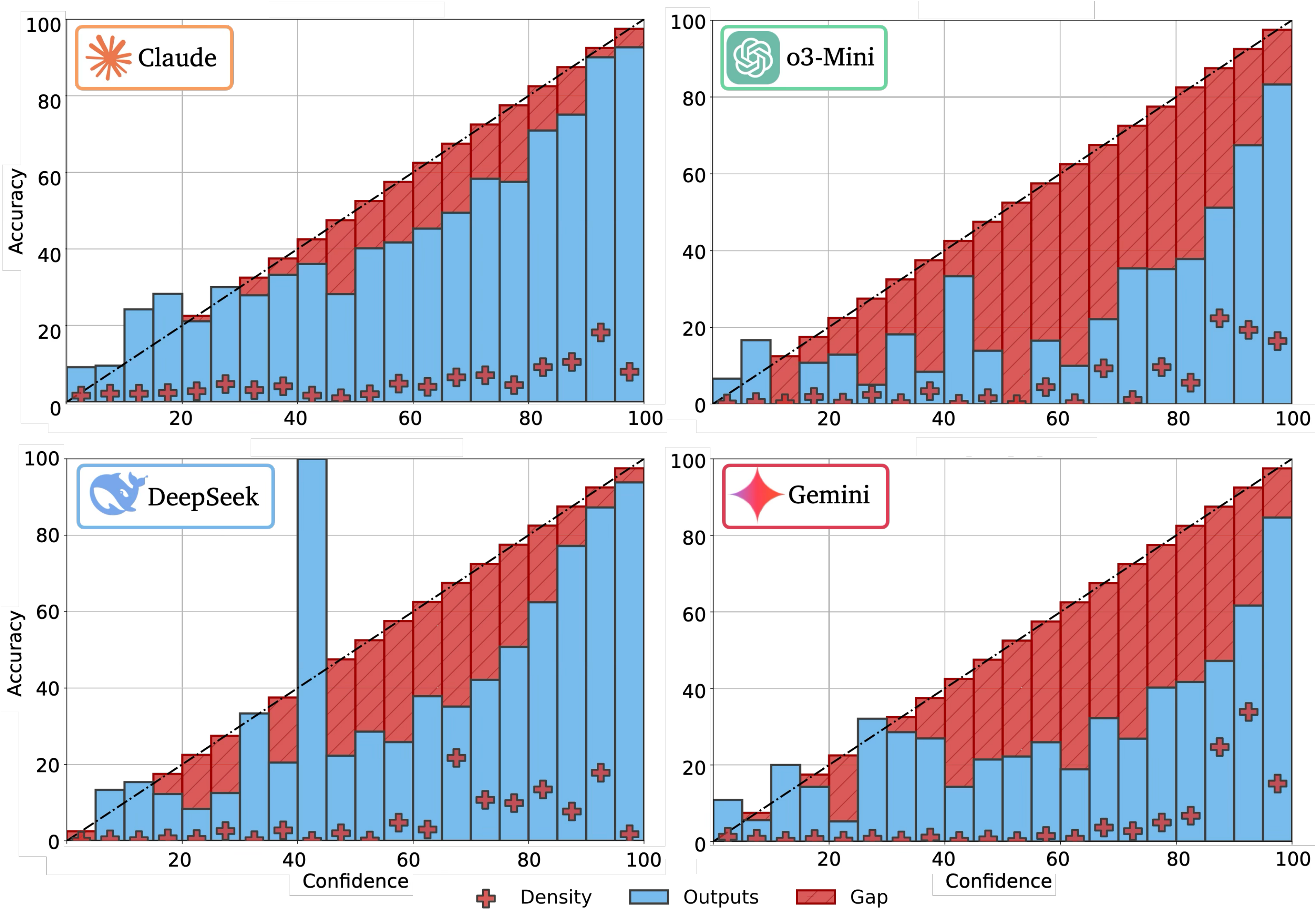}
    \caption{\textbf{Are reasoning models consistently underconfident or overconfident?} Reasoning models are systematically overconfident, with Gemini and o3-Mini being the most overconfident followed by DeepSeek, while Claude is the best calibrated one, highlighted by the reliability diagrams for StrategyQA, GPQA, and SimpleQA aggregated over all prompt strategies.}
    \label{fig:calibration_reliability_hard}
\end{figure*}

\smallskip
\noindent\textbf{Are Reasoning Models Underconfident or Overconfident?}
\phantomsection
\label{sec:experiments_model_under_over_confident}
We plot the reliability diagrams of the reasoning models in~\Cref{fig:calibration_reliability_hard} for the three challenging datasets (StrategyQA, GPQA, and SimpleQA), aggregating the empirical confidence estimates across all prompt strategies. 
In these more challenging datasets, we see that reasoning models are generally overconfident, with confidence estimates typically greater than $85\%$. This is seen by examining the density of samples in each confidence bin (i.e., the number of times the model predicts a confidence level within the given bin divided by the total number of queries), denoted by the red crosses in~\Cref{fig:calibration_reliability_hard}. 
Larger red gaps in the reliability diagram signify greater overconfidence (miscalibration), representing the deviation between the estimated bin-wise confidence and perfect calibration.
\Cref{fig:calibration_reliability_hard} shows that Gemini and o3-Mini are higly overconfident, with significant concentration of their confidence estimates in the $85\%$ to $100\%$ confidence bins. In particular, the accuracy of the samples within these bins significantly deviates from the estimated bin-wise confidence. 
DeepSeek also exhibits overconfidence, although to a smaller degree. We observe large miscalibration and high sample densities especially between the $70\%$ and $80\%$ confidence intervals. 
Claude is the best calibrated model, with significantly smaller miscalibration gaps. However, Claude still appears to be overconfident, particularly in the $60\%$ and $80\%$ confidence range.
We provide the relibability diagram for all datasets in Appendix~\ref{sec:app_evaluations_calib_reasoning_models}.

\smallskip
\noindent\textbf{Comparison among the Models.}
We compare the models' calibration, accuracy, and completion capabilities to identify the best-performing model.  
From \Cref{fig:calibration_reasoning_models,fig:calibration_reliability_hard}, we find that Claude is the most calibrated model by a significant margin, followed by DeepSeek, based on the ECE. This finding appears to be consistent across all datasets, except StrategyQA and GPQA, where DeepSeek achieves the best calibration. 
In contrast,  Gemini is the least calibrated model, although the calibration of o3-Mini is almost the same as that of Gemini in many datasets.
In addition, Claude 3.7 Sonnet is about $2\%$ more accurate than DeepSeek (the best-competing method) and over $4\%$ more accurate than o3-Mini (the least accurate model), which is not entirely surprising given that the compact o3-Mini trades off performance for cost efficiency relative to o3. Further, DeepSeek and Gemini achieve almost the same accuracy, with only a $0.6\%$ gap.
The superior accuracy of Claude 3.7 Sonnet may be explained by the recency of its release compared to the other models: Claude 3.7 Sonnet is the latest model to be released among the models.
Lastly, all models completed more than $97\%$ of the questions, with o3-Mini achieving the highest completion percentage, indicating its ability to interpret the questions and follow the instructions required to provide valid answers.

\subsection{Correlation Between Accuracy and Calibration of Reasoning Models}
\label{ssec:acc_vs_calib_reasoning_models}
The preceding results showed that reasoning models are poorly calibrated in datasets where these models have lower accuracies and better calibrated otherwise. Here, we investigate the relationship between accuracy and calibration of reasoning models by examining the correlation between these two metrics. We compute the average accuracy and ECE over the four prompt strategies per model.

\begin{figure*}[th]
    \centering
    \includegraphics[width=\linewidth]{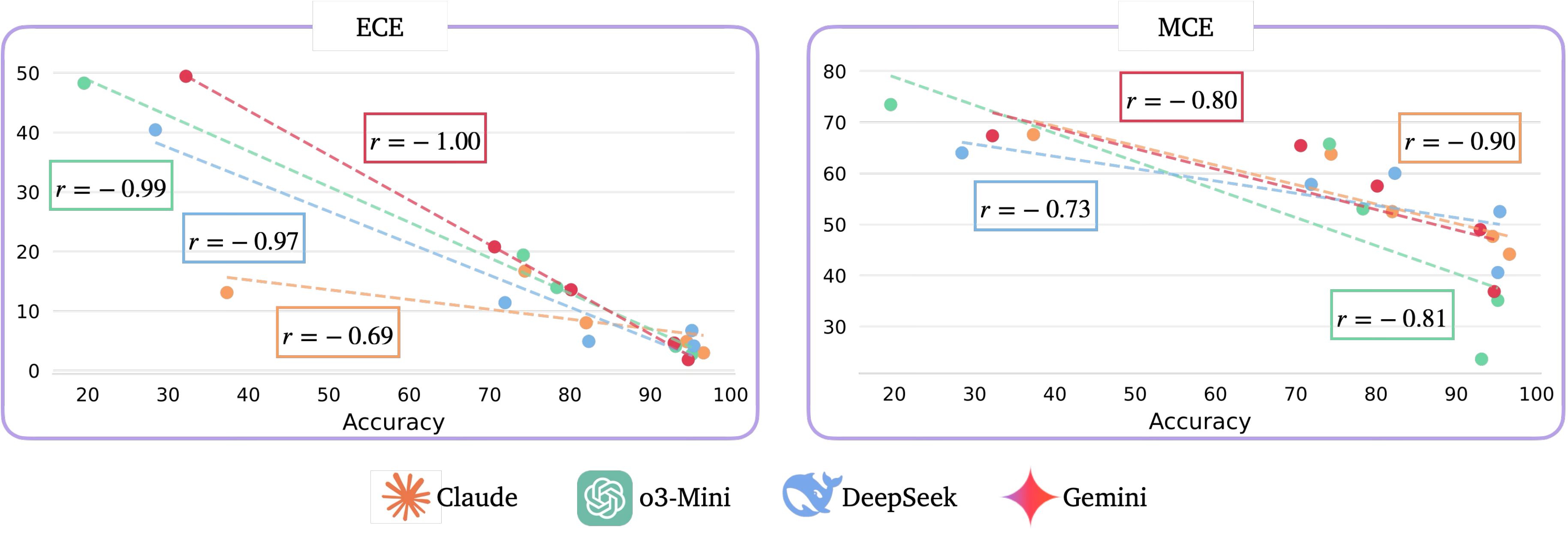}
    \caption{\textbf{Is accuracy correlated with calibration in reasoning models?}
    The ECE is strongly correlated with accuracy in state-of-the-art reasoning models. In essence, the self-verbalized confidence of reasoning models is generally  not trustworthy in problems where they achieve low accuracy. Remarkably, Claude remains well-calibrated even in these low-accuracy settings. We observe the same trends in the correlation between accuracy and the MCE. }
    \label{fig:accuracy_vs_calibration}
\end{figure*}

In \Cref{fig:accuracy_vs_calibration}, we observe a strong negative correlation between accuracy and ECE across all models, which in the case of Gemini, is a perfect negative correlation, i.e., ${r = -1}$. Similarly, DeepSeek and o3-Mini exhibit near-perfect correlation.
This finding underscores that reasoning models are poorly calibrated.
As shown in \Cref{fig:calibration_reliability_hard}, reasoning models are overconfident, which explains the strong correlation between accuracy and calibration. 
In general, if an overconfident model consistently outputs high confidence values, then its ECE will decrease as its accuracy increases.
The confidence of a well-calibrated model should reflect its accuracy, represented by a near-zero correlation coefficient between accuracy and ECE.
However, unlike the other models, Claude has a much weaker correlation with ${r = -0.69}$ ($28\%$ smaller than that of the second-best-performing model DeepSeek), showing its relatively better calibration. 

Further, these results suggest that self-verbalized confidence estimates from reasoning models should be taken with great care, especially with poorly calibrated models.
Lastly, as with the ECE, accuracy and MCE have a strong negative correlation, with DeepSeek showing the weakest correlation.
Training reasoning models to not only be accurate but also be well-calibrated remains a fundamental research challenge, as highlighted by these results. 
We discuss the correlation between calibration error and accuracy for each prompt in~\Cref{sec:app_evaluations_accuracy_calib_vs_reasoning}.
We explore open research problems and potential future research directions in Section~\ref{sec:limitations_future_work}.

\subsection{Reasoning Depth vs. Calibration}
\label{ssec:reasoning_depth_vs_calibration}
\begin{figure*}[th]
    \centering
    \includegraphics[width=\linewidth]{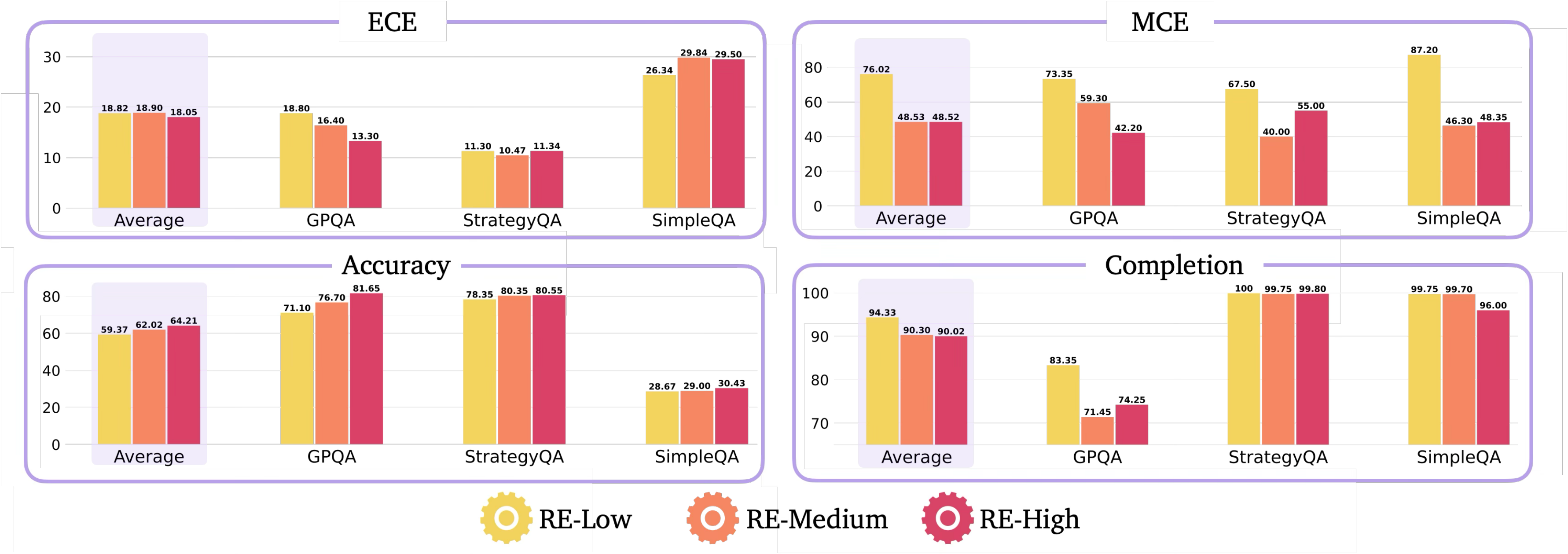}
    \caption{\textbf{Does increasing reasoning depth improve calibration?}
    Deeper reasoning leads to overall higher accuracy;  however, as the accuracy of these models saturate, reasoning models become even more overconfident.
     }
    \label{fig:reasoning_depth_vs_calibration}
\end{figure*}

One of the most appealing properties of reasoning models is \emph{inference-time scaling}: the ability to boost the accuracy of responses via \emph{deeper reasoning} (albeit at the expense of increased inference time and energy costs). However, the impact of deeper reasoning on calibration has not been explored in prior work. Here, we examine the coupled effects of reasoning depth on the accuracy and calibration of reasoning models, highlighting some notable findings.
We summarize our results in \Cref{fig:reasoning_depth_vs_calibration}, evaluating Claude 3.7 Sonnet and o3-Mini since DeepSeek-R1 and Gemini do not sufficiently support/utilize the specification of a reasoning token budget. We consider three reasoning levels: \emph{RE-Low}, \emph{RE-Medium}, and \emph{RE-High}, implemented by the OpenAI API. We estimate the number of thinking tokens corresponding to these levels, and specify these parameters as a thinking budget to Claude. We compute the average of each metric across the two reasoning models.

\begin{figure*}[th]
    \centering
    \includegraphics[width=\linewidth]{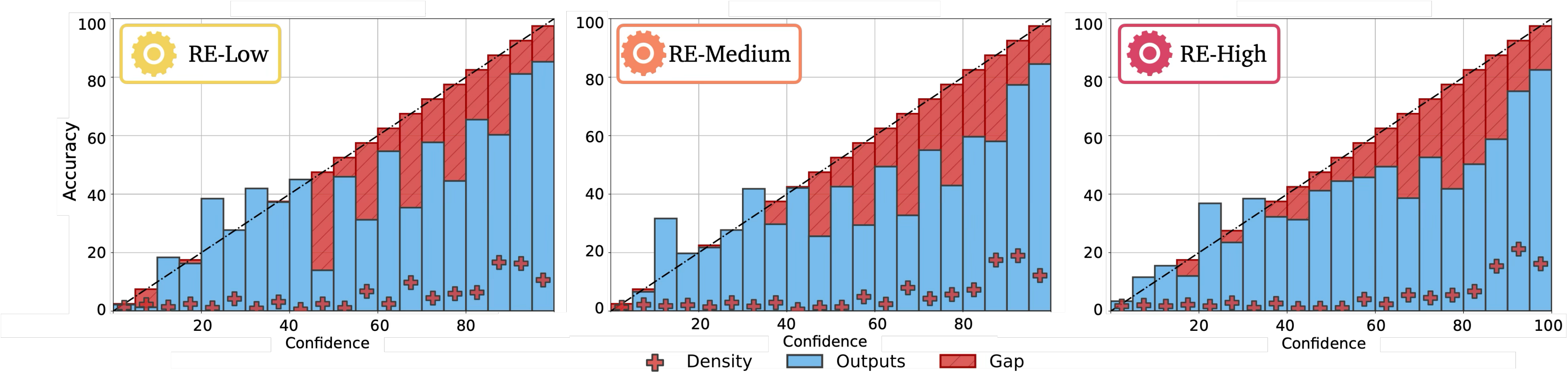}
    \caption{\textbf{Deeper Reasoning vs. Underconfidence/Overconfidence?} 
    Reasoning models become more overconfident with deeper reasoning, evidenced by the increase in the density of samples in higher-confidence bins (e.g., the $90\%$-$95\%$ interval) without a corresponding increase in accuracy.
    }
    \label{fig:reasoning_depth_calibration_reliability}
\end{figure*}

On average, we find that deeper reasoning results in higher accuracies and better calibration, highlighted in \Cref{fig:reasoning_depth_vs_calibration}. However, we observe a more nuanced relationship between reasoning depth, accuracy, and calibration upon closer examination. Specifically, our analysis of the results in \Cref{fig:reasoning_depth_vs_calibration} reveals three key findings.
First, model calibration improves with increasing reasoning depth whenever deeper reasoning leads to an increase in accuracy. This observation can be seen in the GPQA dataset, where an increase in reasoning depth from low to high leads to a $10.8\%$ increase in accuracy and an associated $5.5\%$ decrease in the ECE.
Second, model calibration and accuracy remain relatively unchanged with greater reasoning depth in ``easier" benchmarks, i.e., in datasets where the model achieves relatively high accuracy, e.g., in StrategyQA. In these settings, deeper reasoning does not result in an appreciable increase in accuracy. Moreover, model overconfidence becomes difficult to identify, masked by high accuracy.
Third, as the accuracy of the model saturates in more challenging datasets, deeper reasoning leads to an \emph{increase} in the calibration errors, which is visible in the SimpleQA benchmark. As the reasoning depth increases from low to high, we observe a relatively small increase in accuracy and a much larger increase in the ECE. 
To further explore this finding, we provide the reliability diagrams associated with the reasoning depths aggregated over all datasets in \Cref{fig:reasoning_depth_calibration_reliability}, showing that the proportion of highly confident responses increases with reasoning depth, oftentimes without a corresponding increase in accuracy, degrading the calibration of the model. For example, in \Cref{fig:reasoning_depth_calibration_reliability}, we see that reasoning models are generally overconfident with most of the model's confidence estimates residing between $85\%$ and $100\%$. As we increase the reasoning depth, the density of samples in this interval increases while the accuracy of these samples decreases (e.g., samples in the $90\%$-$95\%$ interval), underscoring that \emph{reasoning models become even more overconfident with deeper reasoning.} 

Further, MCE decreases with greater reasoning depth (\Cref{fig:reasoning_depth_vs_calibration}). We note that the worst-case calibration error is generally associated with confidence bins containing a single or only a few samples. Deeper reasoning generally increases either the accuracy of these samples or the confidence associated with these samples, since the affected samples typically move to a bin with more samples, ultimately reducing their influence on the MCE.
Lastly, from \Cref{fig:reasoning_depth_calibration_reliability}, the completion rate of reasoning models decreases with greater reasoning depth. Reasoning models tend to struggle more with following format instructions specified in the system prompt when provided with a greater thinking budget. This observation is more prominent in scientific domains, e.g., GPQA, where reasoning involves significant scientific notation and equations. 

\begin{figure*}[th]
    \centering
    \includegraphics[width=\linewidth]{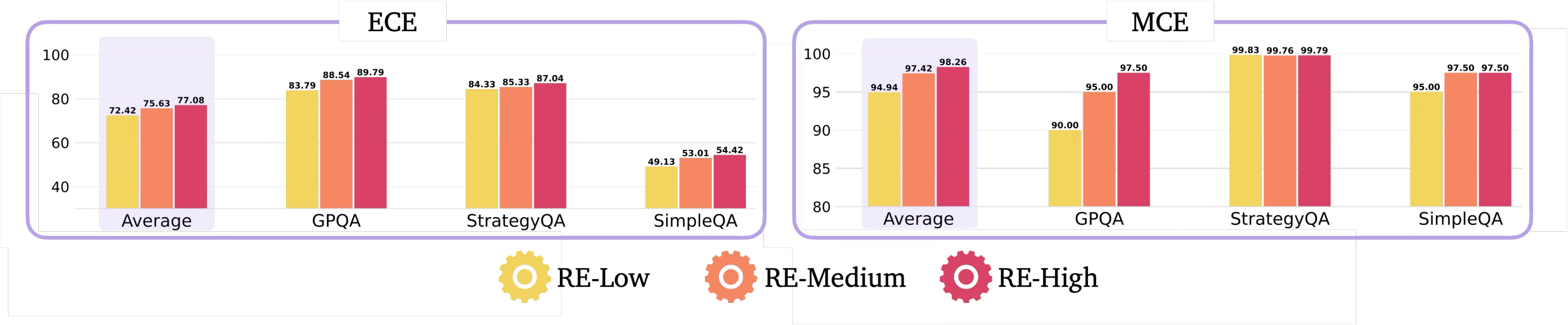}
    \caption{\textbf{Does increasing reasoning depth make the models more confidently wrong on average?}
    On wrongly-answered questions, reasoning models become more overconfident, even though the correctness of their answers does not change.}
    \label{fig:reasoning_depth_vs_calibration_wrong_answers}
\end{figure*}

\begin{figure*}[th]
    \centering
    \includegraphics[width=\linewidth]{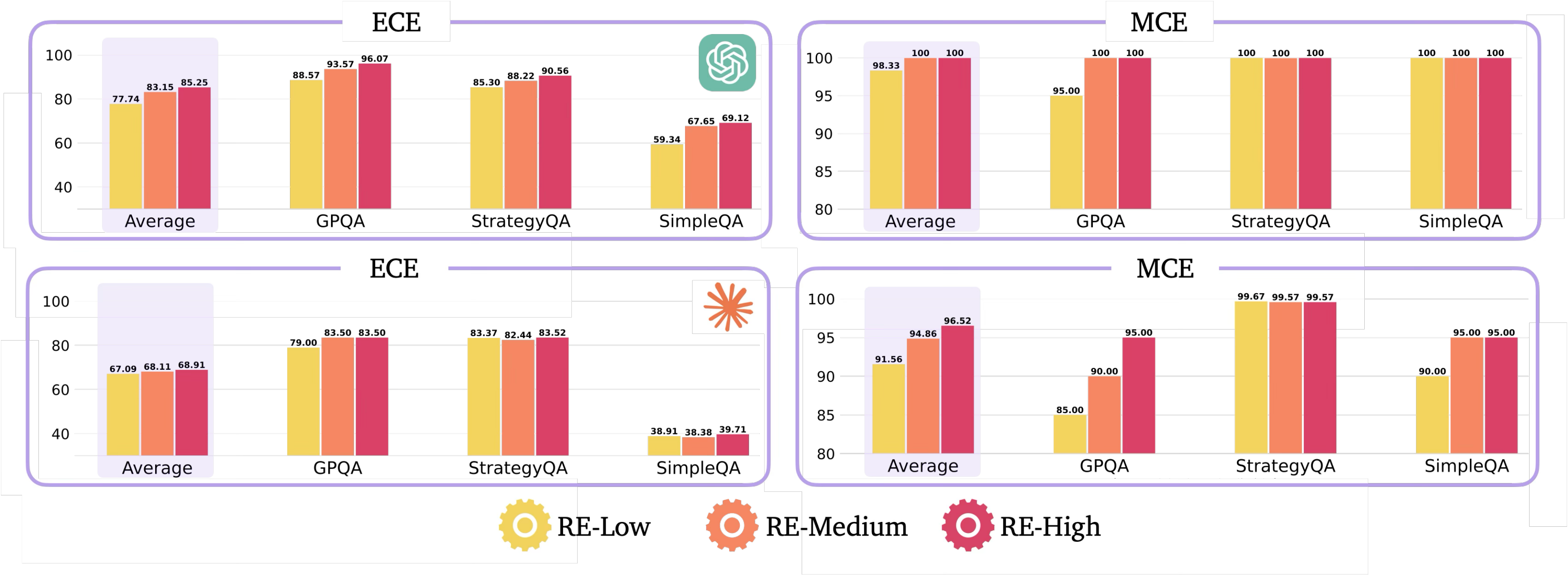}
    \caption{\textbf{Does increasing reasoning depth make the models more confidently wrong?}
    With more reasoning effort, o3-Mini's overconfidence worsens, while Claude's overconfidence remains relatively constant.}
    \label{fig:reasoning_depth_vs_calibration_wrong_answers_per_model}
\end{figure*}

Next, for a more detailed analysis, we examine the calibration errors as a function of reasoning depth on \emph{wrongly-answered questions}, (i.e., on questions with incorrect responses at all reasoning depth levels). \Cref{fig:reasoning_depth_vs_calibration_wrong_answers} reveals that model calibration degrades with reasoning depth in this domain with an increase in the average ECE and MCE. This result suggests that deeper reasoning makes reasoning models more confident in their incorrect responses by reinforcing their misguided thought process in these problems. 
This result bears some similarity to human behavior, since prior work has shown that humans tend to provide higher estimates of their confidence in their answers \cite{shynkaruk2006confidence} when given more time to think, even when their answers do not change. 
Moreover, we find that the relative effect of reasoning depth on calibration varies significantly with the model. From \Cref{fig:reasoning_depth_vs_calibration_wrong_answers_per_model}, o3-Mini is more susceptible to reasoning-induced overconfidence, with a $7\%$ absolute increase in the ECE from the low-reasoning to the high-reasoning setting, compared to a $2\%$ increase for Claude. In fact, not only is Claude more robust, it is also better calibrated across all datasets.
Moreover, the MCE increases for all models as all models become more confident without a corresponding increase in their accuracy.

\subsection{Reasoning about Confidence}
\label{ssec:reasoning_about_confidence}
Next, we examine if model calibration can be improved by \emph{reasoning about uncertainty}. Specifically, we utilize a two-stage introspective uncertainty quantification procedure (see Section~\ref{sec:introspective_uq}). In the first stage, we prompt each model to provide an answer to the question and an associated confidence in the answer. In the second stage, we provide the reasoning trace from the first stage to a fresh instance of the model and prompt the model to identify potential flaws in the  the reasoning trace to provide an updated confidence estimate in the original answer.
We explore three prompt strategies for introspection, varying the level of conservativeness: \emph{IUQ-Low}, \emph{IUQ-Medium}, and \emph{IUQ-High}, described in \Cref{sec:introspective_uq}. In IUQ-Medium and IUQ-High, we specifically ask the model to identify flaws in the reasoning traces. In all prompts, we do not ask the model for an updated answer to discourage the model from directly reasoning about the original question, e.g., the validity of all options in a multi-choice question.
The second stage uses a new session with a re-initialized context window that does not contain information from the first session. 
We do not include Gemini in these results due to insufficient support for analysis of the reasoning traces.

\begin{figure*}[th]
    \centering
    \includegraphics[width=\linewidth]{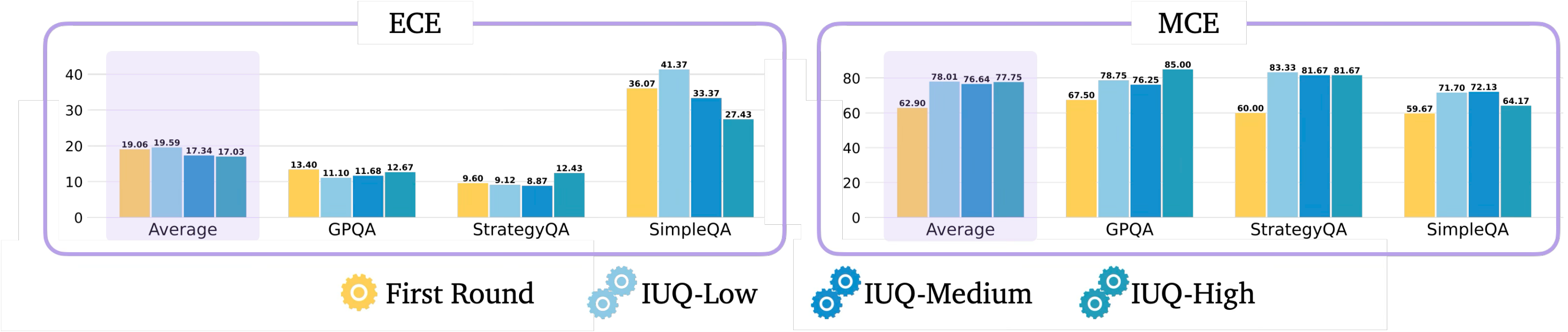}
    \caption{\textbf{Introspective UQ.}
    Calibration of reasoning models improves with introspective uncertainty quantification, particularly in challenging datasets, e.g., SimpleQA, with more conservative prompts, e.g., IUQ-Medium and IUQ-High.
     }
    \label{fig:introspective_reasoning_vs_calibration_average}
\end{figure*}

\begin{figure*}[th]
    \centering
    \includegraphics[width=\linewidth]{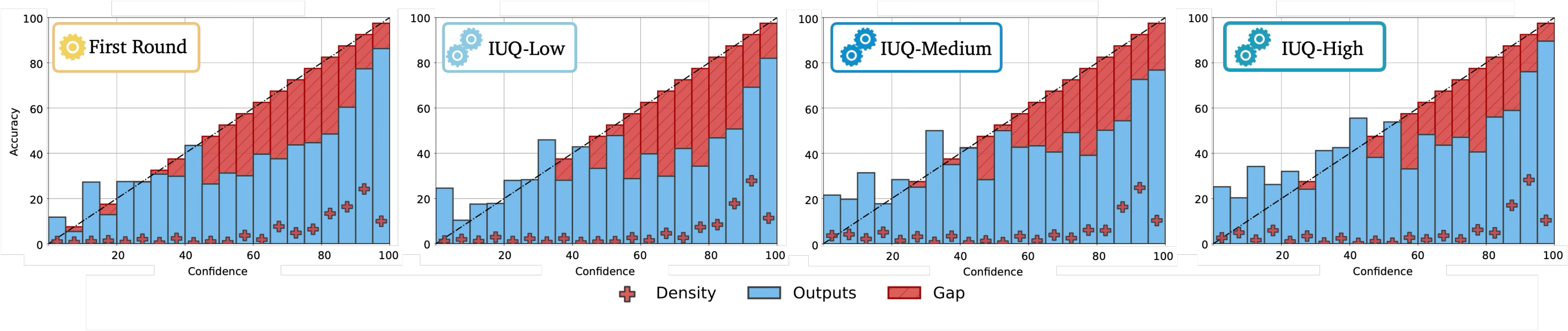}
    \caption{\textbf{Introspective UQ vs. Underconfidence/Overconfidence.}
    More critical introspection, e.g., IUQ-Medium and IUQ-High, improves the calibration of reasoning models, mitigating model overconfidence, unlike IUQ-Low.
     }
    \label{fig:reliability_introspective_reasoning_average}
\end{figure*}

We compute the resulting calibration errors of each model across the challenging datasets (GPQA, StrategyQA, and SimpleQA).
In \Cref{fig:introspective_reasoning_vs_calibration_average}, when averaging across all models, we observe that introspective UQ leads to a marginal increase in the average ECE in IUQ-Low and a decrease in IUQ-Medium and IUQ-High. The change in the ECE varies significantly with the dataset and the introspective prompt. Unlike the MCE which increases across all datasets, in the least conservative introspection, the ECE decreases in the GPQA and StrategyQA datasets but increases in the SimpleQA dataset. However, with IUQ-Medium and IUQ-High, we observe a decrease in the ECE across all datasets, except in StrategyQA in the case of IUQ-High. Asking the reasoning model to identify flaws in the prior reasoning traces tends to induce more critical introspection, leading to a significant decrease in miscalibration, especially in the most challenging dataset, SimpleQA. 
Moreover, when prior confidence estimates are not provided to reasoning models (i.e., in IUQ-High), reasoning models reason more conservatively during introspection, which leads to better calibration in challenging datasets. 
Conversely, on easier datasets, greater conservativeness from more critical introspection could slightly increase the ECE, e.g., IUQ-High on StrategyQA.
To further examine this finding, we visualize the reliability diagrams associated with the first and second stages in \Cref{fig:reliability_introspective_reasoning_average}, showing that less critical introspection, e.g., IUQ-Low, can increase model overconfidence, i.e.,  a rightward shift in the confidence estimates from low-confidence bins to high-confidence bins. For example, we observe an increase in the density of samples in the $0.85$-$1.0$ confidence bins, in IUQ-Low. However, more critical introspection, e.g., IUQ-Medium and IUQ-High, leads to better calibration, reducing overconfidence.

\begin{figure*}[th]
    \centering
    \includegraphics[width=\linewidth]{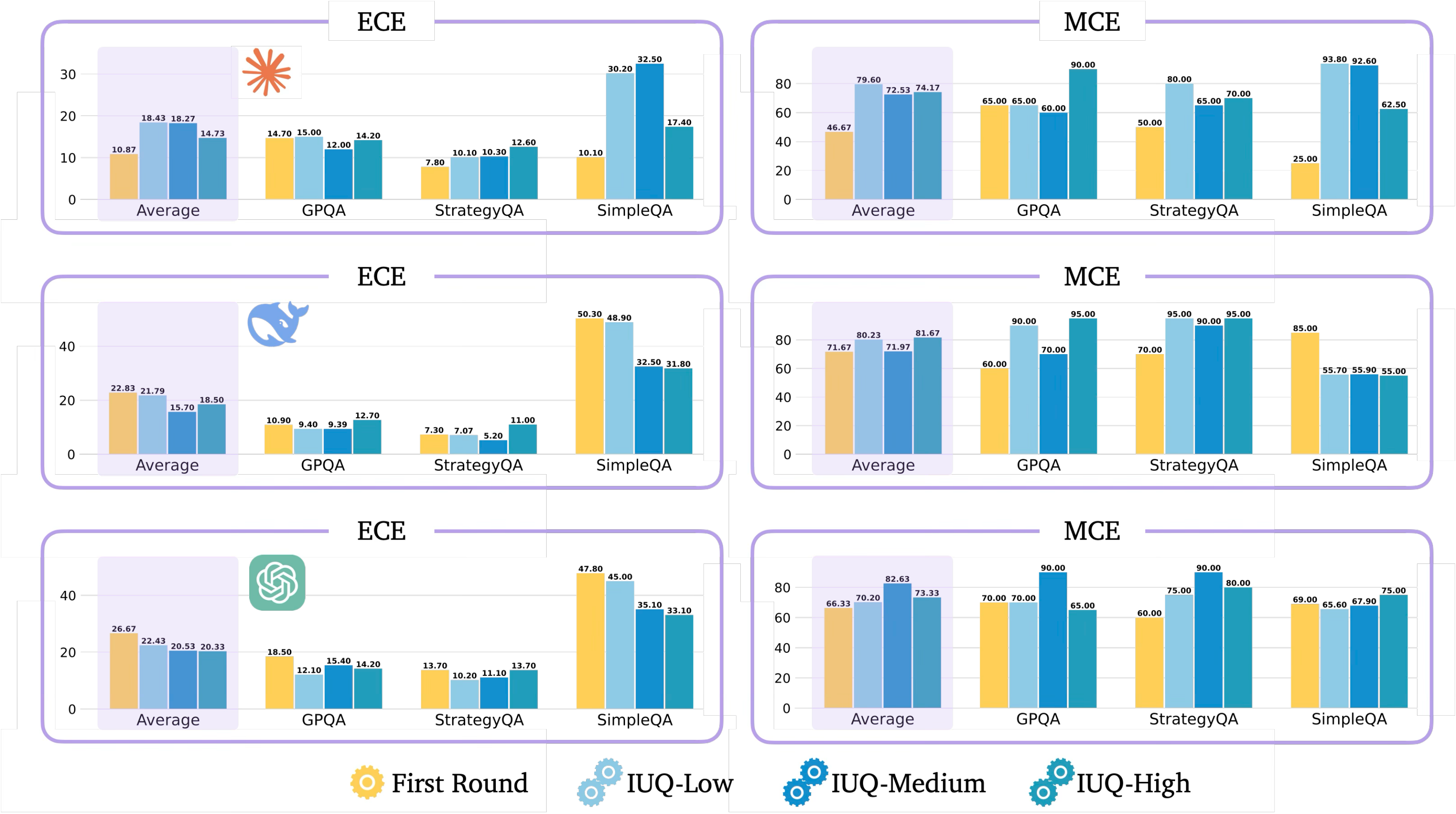}
    \caption{\textbf{Introspective UQ per Model.}
    Introspective UQ improves the calibration of DeepSeek and o3-Mini, especially in the challenging dataset SimpleQA, but degrades the calibration of Claude.
     }
    \label{fig:introspective_reasoning_vs_calibration_models}
\end{figure*}

\begin{figure*}[th]
    \centering
    \includegraphics[width=\linewidth]{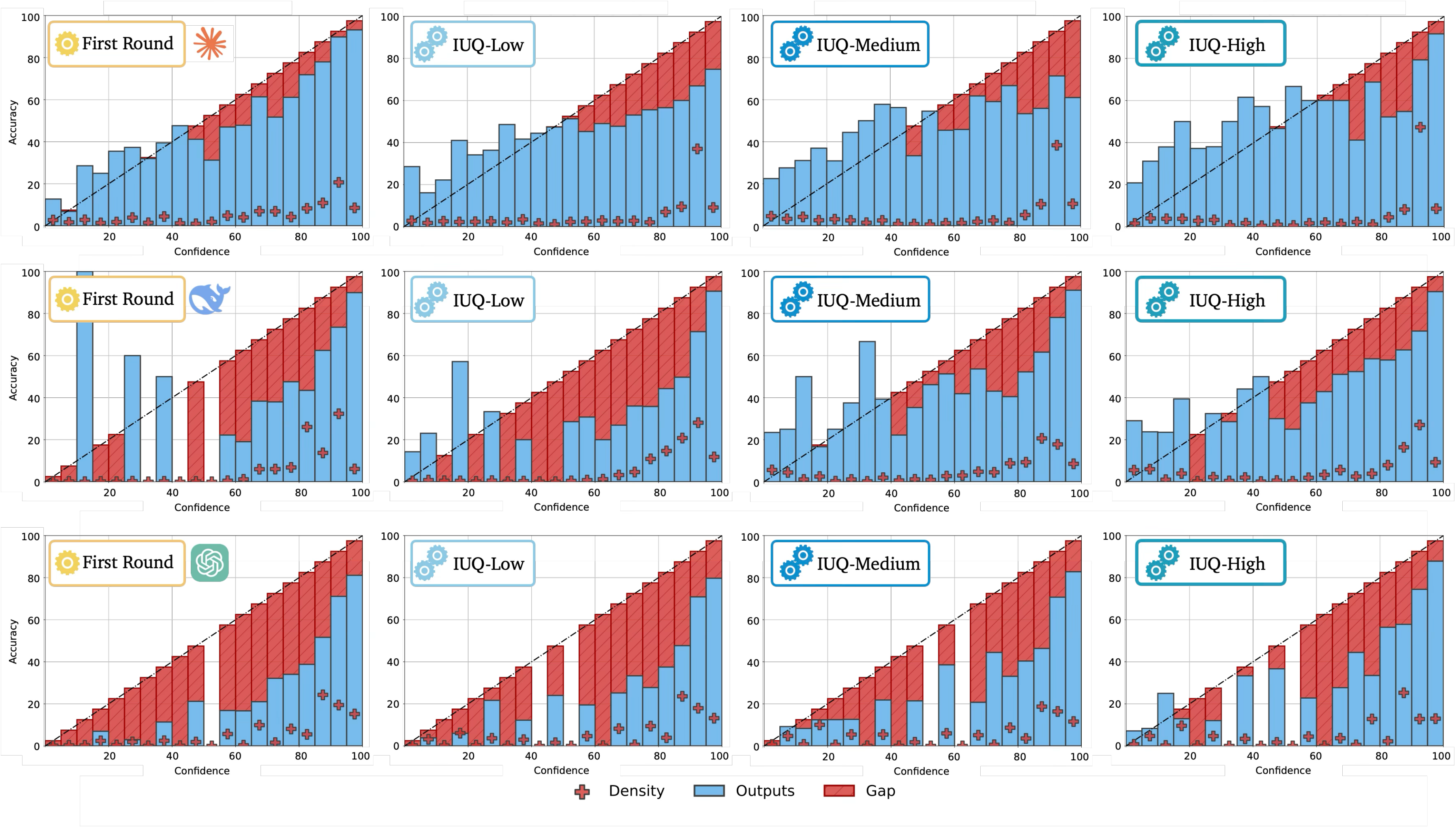}
    \caption{\textbf{Introspective UQ per Model vs. Underconfidence/Overconfidence.}
    Through introspection, DeepSeek and o3-Mini become less overconfident, especially with IUQ-Medium and IUQ-High, unlike Claude, which becomes more overconfident.
     }
    \label{fig:reliability_introspective_reasoning_per_model}
\end{figure*}

We further analyze the calibration of introspective UQ estimates for each model, summarized in \Cref{fig:introspective_reasoning_vs_calibration_models}. 
We find that calibration of Claude significantly degrades when the model reasons about its uncertainty, particularly in StrategyQA and SimpleQA. Specifically, Claude becomes more overconfident with introspective UQ, as highlighted in \Cref{fig:reliability_introspective_reasoning_per_model}, showing that the proportion of confidence estimates in the $0.9$-$0.95$ confidence bin nearly doubles, with a corresponding decrease in the density of almost all the bins below $0.8$. This observation may be due to the implementation details of Claude, which is closed-source.

In contrast, introspection improves the calibration of DeepSeek and o3-Mini, with bigger improvements with IUQ-Medium and IUQ-High in SimpleQA. During the second stage, critical introspection enables these models to identify flaws and hallucinations in their reasoning traces, leading to effective reassessment of their confidence. \Cref{fig:reliability_introspective_reasoning_per_model} highlights these findings, showing an increase in the density of the lower-confidence bins with a corresponding decrease in the density of the higher-confidence bins, ultimately reducing overconfidence.

In summary, these results suggest that introspective UQ improves the calibration of reasoning models, particularly in challenging problems where introspection enables these models to identify flaws in their reasoning.

\subsection{Calibration and Accuracy of Reasoning Models vs. Prompt Strategies}
\label{ssec:acc_calib_vs_prompt_reasoning_models}
Prior work has shown that prompt engineering can improve the performance of LLMs significantly. For example, existing work has demonstrated that chain-of-thought (CoT) prompting leads to higher accuracy in  non-reasoning LLMs, e.g., in zero-shot tasks. Here, we explore the question: do prompt strategies such as chain-of-prompting improve the accuracy and calibration of reasoning models? To do so, we compute the accuracy and calibration over the four prompt strategies: basic, CoT, multi-step, and top-K prompts, averaging the models' performance within each dataset.

\begin{figure*}[th]
    \centering
    \includegraphics[width=\linewidth]{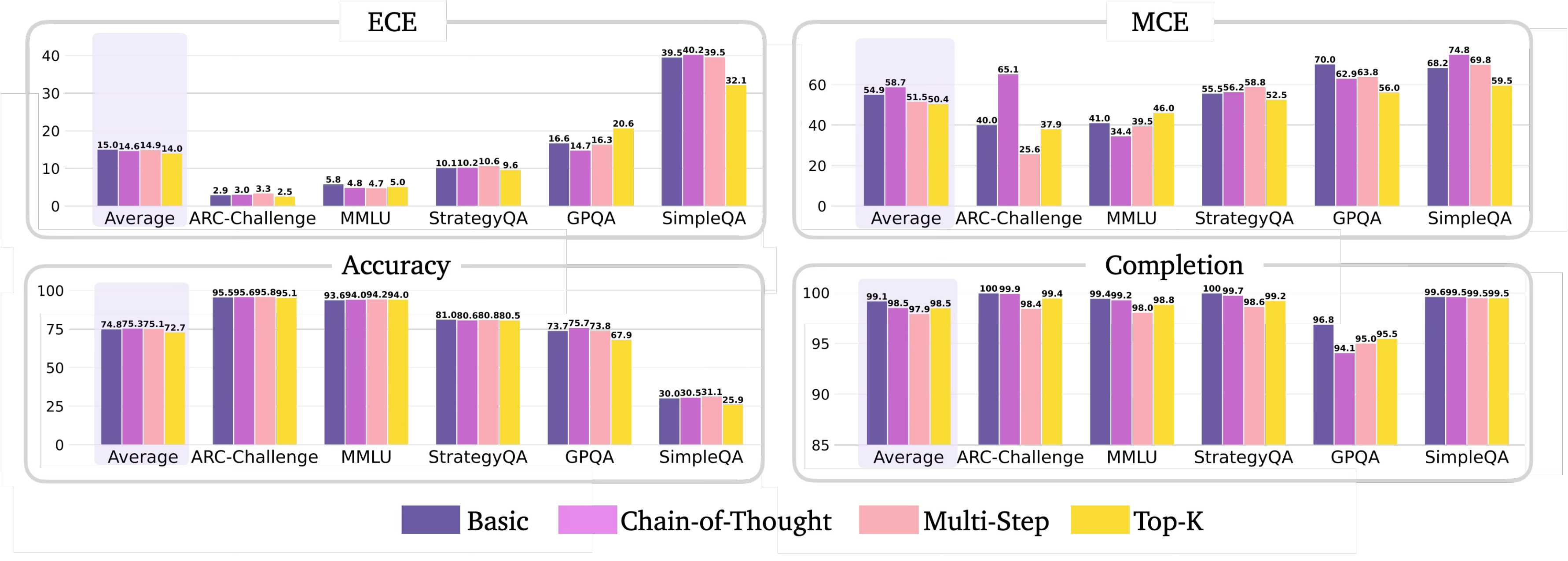}
    \caption{\textbf{Do specialized prompt strategies (e.g., chain-of-thought prompting) improve calibration and accuracy?}
    Chain-of-thought prompting and multi-step prompting do not outperform the basic prompt strategy. On average, calibration and accuracy of reasoning models do not depend on the prompt strategy. 
     }
    \label{fig:metrics_vs_prompt}
\end{figure*}

From \Cref{fig:metrics_vs_prompt}, we observe that the performance of reasoning models is relatively independent of the selected prompt strategy. In particular, specialized prompt strategies such as CoT prompting and multi-step prompting which have been shown to be effective in non-reasoning models do not significantly outperform basic prompting techniques that just ask the model directly for its answer and confidence in terms of calibration, accuracy, and completion rates. 
In non-reasoning models, CoT prompting and multi-step prompting are generally effective due to their ability to elicit reasoning in these models.
However, reasoning models are trained to generate reasoning traces by default. Consequently, specialized prompt strategies are not necessary to induce reasoning, explaining the independence between the performance of reasoning models and prompt strategies. We provide more details on each model in Appendix~\ref{sec:app_evaluations_calib_accuracy_vs_prompt}.

\section{Conclusion}
\label{sec:conclusion}
In this paper, we examine uncertainty quantification of reasoning models. To this end, we ask three critical questions: 
First, \emph{are reasoning models well-calibrated?} 
Second, \emph{does deeper reasoning improve model calibration?}
Third, \emph{can reasoning models improve their calibration by explicitly reasoning about their chain-of-thought traces?}
Through extensive evaluations, we observe that reasoning models are generally overconfident and become even more overconfident with deeper reasoning, especially in tasks where deeper reasoning does not correspond to higher accuracy. Lastly, the calibration of these model improves with more critical introspection, although introspection increases the overconfidence of Claude.

\section{Limitations and Future Work}
\label{sec:limitations_future_work}
The results in our work highlight that in general, reasoning models are poorly calibrated despite their remarkable accuracy, posing a challenge to real-world deployment. We believe that improving the calibration of reasoning models is essential to drive widespread safe adoption of these models. Here, we identify some valuable research directions for future work.

\smallskip
\noindent\textbf{Reducing the Correlation between Accuracy and Calibration.}
As discussed in \Cref{sec:evaluation}, the accuracy of reasoning models is strongly positively calibrated with their calibration. Stable, well-calibrated models should have a near-zero correlation between accuracy and calibration, i.e., the model should remain well-calibrated in relatively easier problem domains and relatively more challenging problem domains. Training stable, well-calibrated reasoning models remains an open challenge, presenting a valuable direction for future research. 
Future work will explore techniques that preserve model calibration even when the accuracy of these models is low. For example, reasoning models can be trained to recognize challenging problems and provide more conservative confidence estimates in these knowledge domains.

\smallskip
\noindent\textbf{Training Reasoning Models to Reason about their Uncertainty.}
Reinforcement-learning-induced multi-step reasoning has enabled reasoning models to set new performance records on a broad range of benchmarks. By decomposing challenging problems into easier-to-solve subproblems, reasoning models have achieved superior accuracy compared to prior language models. We believe that reasoning models can be trained to perform multi-step reasoning about their \emph{uncertainty} through reinforcement learning, constituting an exciting direction for future work. By training models to explicitly reason about the boundaries of their knowledge, reasoning models can be empowered to better estimate their confidence in the correctness of their responses.

\smallskip
\noindent\textbf{Benchmarks for Uncertainty Quantification.}
As noted in the preceding discussion in \Cref{sec:evaluation}, the performance of reasoning models on many existing benchmarks has saturated, making these benchmarks essentially uninformative in uncertainty quantification. Future work will seek to introduce new benchmarks that better evaluate the calibration of reasoning models, across a range of disciplines. For example, new benchmarks for evaluating the calibration of reasoning models at intermediate steps in the problem-solving process would shed more light on the calibration of the thought processes of reasoning models.

\section*{Acknowledgments}
The authors were partially supported by the Office of Naval Research (N00014-23-1-2148) and a Sloan Fellowship.

\clearpage

\bibliographystyle{plainnat}
\bibliography{references.bib}

\begin{thebibliography}{58}
\providecommand{\natexlab}[1]{#1}
\providecommand{\url}[1]{\texttt{#1}}
\expandafter\ifx\csname urlstyle\endcsname\relax
  \providecommand{\doi}[1]{doi: #1}\else
  \providecommand{\doi}{doi: \begingroup \urlstyle{rm}\Url}\fi

\bibitem[Aichberger et~al.(2024)Aichberger, Schweighofer, Ielanskyi, and Hochreiter]{aichberger2024semantically}
Lukas Aichberger, Kajetan Schweighofer, Mykyta Ielanskyi, and Sepp Hochreiter.
\newblock Semantically diverse language generation for uncertainty estimation in language models.
\newblock \emph{arXiv preprint arXiv:2406.04306}, 2024.

\bibitem[Anthropic(2024)]{anthropicthec3_7}
Anthropic.
\newblock Claude 3.7 sonnet system card.
\newblock Technical report, Anthropic, 2024.
\newblock URL \url{https://assets.anthropic.com/m/785e231869ea8b3b/original/claude-3-7-sonnet-system-card.pdf}.

\bibitem[{Art of Problem Solving}(2025)]{AoPSWikiAIME}
{Art of Problem Solving}.
\newblock {American Invitational Mathematics Examination}.
\newblock \url{https://artofproblemsolving.com/wiki/index.php/American_Invitational_Mathematics_Examination}, 2025.

\bibitem[Azaria and Mitchell(2023)]{azaria2023internal}
Amos Azaria and Tom Mitchell.
\newblock The internal state of an llm knows when it's lying.
\newblock \emph{arXiv preprint arXiv:2304.13734}, 2023.

\bibitem[Bai et~al.(2024)Bai, Wang, Xiao, He, Han, Zhang, and Shou]{bai2024hallucination}
Zechen Bai, Pichao Wang, Tianjun Xiao, Tong He, Zongbo Han, Zheng Zhang, and Mike~Zheng Shou.
\newblock Hallucination of multimodal large language models: A survey.
\newblock \emph{arXiv preprint arXiv:2404.18930}, 2024.

\bibitem[Belinkov(2022)]{belinkov2022probing}
Yonatan Belinkov.
\newblock Probing classifiers: Promises, shortcomings, and advances.
\newblock \emph{Computational Linguistics}, 48\penalty0 (1):\penalty0 207--219, 2022.

\bibitem[Besta et~al.(2024)Besta, Blach, Kubicek, Gerstenberger, Podstawski, Gianinazzi, Gajda, Lehmann, Niewiadomski, Nyczyk, et~al.]{besta2024graph}
Maciej Besta, Nils Blach, Ales Kubicek, Robert Gerstenberger, Michal Podstawski, Lukas Gianinazzi, Joanna Gajda, Tomasz Lehmann, Hubert Niewiadomski, Piotr Nyczyk, et~al.
\newblock Graph of thoughts: Solving elaborate problems with large language models.
\newblock In \emph{Proceedings of the AAAI Conference on Artificial Intelligence}, volume~38, pages 17682--17690, 2024.

\bibitem[Clark et~al.(2018)Clark, Cowhey, Etzioni, Khot, Sabharwal, Schoenick, and Tafjord]{clark2018think}
Peter Clark, Isaac Cowhey, Oren Etzioni, Tushar Khot, Ashish Sabharwal, Carissa Schoenick, and Oyvind Tafjord.
\newblock Think you have solved question answering? try arc, the ai2 reasoning challenge.
\newblock \emph{arXiv preprint arXiv:1803.05457}, 2018.

\bibitem[Dunefsky et~al.(2024)Dunefsky, Chlenski, and Nanda]{dunefsky2024transcoders}
Jacob Dunefsky, Philippe Chlenski, and Neel Nanda.
\newblock Transcoders find interpretable llm feature circuits.
\newblock \emph{arXiv preprint arXiv:2406.11944}, 2024.

\bibitem[Fadeeva et~al.(2024)Fadeeva, Rubashevskii, Shelmanov, Petrakov, Li, Mubarak, Tsymbalov, Kuzmin, Panchenko, Baldwin, et~al.]{fadeeva2024fact}
Ekaterina Fadeeva, Aleksandr Rubashevskii, Artem Shelmanov, Sergey Petrakov, Haonan Li, Hamdy Mubarak, Evgenii Tsymbalov, Gleb Kuzmin, Alexander Panchenko, Timothy Baldwin, et~al.
\newblock Fact-checking the output of large language models via token-level uncertainty quantification.
\newblock \emph{arXiv preprint arXiv:2403.04696}, 2024.

\bibitem[Ferrando et~al.(2024)Ferrando, Obeso, Rajamanoharan, and Nanda]{ferrando2024know}
Javier Ferrando, Oscar Obeso, Senthooran Rajamanoharan, and Neel Nanda.
\newblock Do i know this entity? knowledge awareness and hallucinations in language models.
\newblock \emph{arXiv preprint arXiv:2411.14257}, 2024.

\bibitem[Fu et~al.(2022)Fu, Peng, Sabharwal, Clark, and Khot]{fu2022complexity}
Yao Fu, Hao Peng, Ashish Sabharwal, Peter Clark, and Tushar Khot.
\newblock Complexity-based prompting for multi-step reasoning.
\newblock \emph{arXiv preprint arXiv:2210.00720}, 2022.

\bibitem[Gao et~al.(2024)Gao, la~Tour, Tillman, Goh, Troll, Radford, Sutskever, Leike, and Wu]{gao2024scaling}
Leo Gao, Tom~Dupr{\'e} la~Tour, Henk Tillman, Gabriel Goh, Rajan Troll, Alec Radford, Ilya Sutskever, Jan Leike, and Jeffrey Wu.
\newblock Scaling and evaluating sparse autoencoders.
\newblock \emph{arXiv preprint arXiv:2406.04093}, 2024.

\bibitem[Geva et~al.(2020)Geva, Schuster, Berant, and Levy]{geva2020transformer}
Mor Geva, Roei Schuster, Jonathan Berant, and Omer Levy.
\newblock Transformer feed-forward layers are key-value memories.
\newblock \emph{arXiv preprint arXiv:2012.14913}, 2020.

\bibitem[Geva et~al.(2021)Geva, Khashabi, Segal, Khot, Roth, and Berant]{geva2021did}
Mor Geva, Daniel Khashabi, Elad Segal, Tushar Khot, Dan Roth, and Jonathan Berant.
\newblock Did aristotle use a laptop? a question answering benchmark with implicit reasoning strategies.
\newblock \emph{Transactions of the Association for Computational Linguistics}, 9:\penalty0 346--361, 2021.

\bibitem[Gneiting and Raftery(2007)]{gneiting2007strictly}
Tilmann Gneiting and Adrian~E Raftery.
\newblock Strictly proper scoring rules, prediction, and estimation.
\newblock \emph{Journal of the American statistical Association}, 102\penalty0 (477):\penalty0 359--378, 2007.

\bibitem[Guo et~al.(2017)Guo, Pleiss, Sun, and Weinberger]{guo2017calibration}
Chuan Guo, Geoff Pleiss, Yu~Sun, and Kilian~Q. Weinberger.
\newblock On calibration of modern neural networks.
\newblock In Doina Precup and Yee~Whye Teh, editors, \emph{Proceedings of the 34th International Conference on Machine Learning}, volume~70 of \emph{Proceedings of Machine Learning Research}, pages 1321--1330. PMLR, 06--11 Aug 2017.
\newblock URL \url{https://proceedings.mlr.press/v70/guo17a.html}.

\bibitem[Guo et~al.(2025)Guo, Yang, Zhang, Song, Zhang, Xu, Zhu, Ma, Wang, Bi, et~al.]{guo2025deepseek}
Daya Guo, Dejian Yang, Haowei Zhang, Junxiao Song, Ruoyu Zhang, Runxin Xu, Qihao Zhu, Shirong Ma, Peiyi Wang, Xiao Bi, et~al.
\newblock Deepseek-r1: Incentivizing reasoning capability in llms via reinforcement learning.
\newblock \emph{arXiv preprint arXiv:2501.12948}, 2025.

\bibitem[Hao et~al.(2023)Hao, Gu, Ma, Hong, Wang, Wang, and Hu]{hao2023reasoning}
Shibo Hao, Yi~Gu, Haodi Ma, Joshua~Jiahua Hong, Zhen Wang, Daisy~Zhe Wang, and Zhiting Hu.
\newblock Reasoning with language model is planning with world model.
\newblock \emph{arXiv preprint arXiv:2305.14992}, 2023.

\bibitem[Hao et~al.(2024)Hao, Gu, Luo, Liu, Shao, Wang, Xie, Ma, Samavedhi, Gao, et~al.]{hao2024llm}
Shibo Hao, Yi~Gu, Haotian Luo, Tianyang Liu, Xiyan Shao, Xinyuan Wang, Shuhua Xie, Haodi Ma, Adithya Samavedhi, Qiyue Gao, et~al.
\newblock Llm reasoners: New evaluation, library, and analysis of step-by-step reasoning with large language models.
\newblock \emph{arXiv preprint arXiv:2404.05221}, 2024.

\bibitem[Hendrycks et~al.(2020)Hendrycks, Burns, Basart, Zou, Mazeika, Song, and Steinhardt]{hendrycks2020measuring}
Dan Hendrycks, Collin Burns, Steven Basart, Andy Zou, Mantas Mazeika, Dawn Song, and Jacob Steinhardt.
\newblock Measuring massive multitask language understanding.
\newblock \emph{arXiv preprint arXiv:2009.03300}, 2020.

\bibitem[Hewitt et~al.(2016)Hewitt, Chreim, and Forster]{hewitt2016double}
Tanya Hewitt, Samia Chreim, and Alan Forster.
\newblock Double checking: a second look.
\newblock \emph{Journal of evaluation in clinical practice}, 22\penalty0 (2):\penalty0 267--274, 2016.

\bibitem[Huang and Chang(2022)]{huang2022towards}
Jie Huang and Kevin Chen-Chuan Chang.
\newblock Towards reasoning in large language models: A survey.
\newblock \emph{arXiv preprint arXiv:2212.10403}, 2022.

\bibitem[Hurst et~al.(2024)Hurst, Lerer, Goucher, Perelman, Ramesh, Clark, Ostrow, Welihinda, Hayes, Radford, et~al.]{hurst2024gpt}
Aaron Hurst, Adam Lerer, Adam~P Goucher, Adam Perelman, Aditya Ramesh, Aidan Clark, AJ~Ostrow, Akila Welihinda, Alan Hayes, Alec Radford, et~al.
\newblock Gpt-4o system card.
\newblock \emph{arXiv preprint arXiv:2410.21276}, 2024.

\bibitem[Jaech et~al.(2024)Jaech, Kalai, Lerer, Richardson, El-Kishky, Low, Helyar, Madry, Beutel, Carney, et~al.]{jaech2024openai}
Aaron Jaech, Adam Kalai, Adam Lerer, Adam Richardson, Ahmed El-Kishky, Aiden Low, Alec Helyar, Aleksander Madry, Alex Beutel, Alex Carney, et~al.
\newblock Openai o1 system card.
\newblock \emph{arXiv preprint arXiv:2412.16720}, 2024.

\bibitem[Jospin et~al.(2022)Jospin, Laga, Boussaid, Buntine, and Bennamoun]{jospin2022hands}
Laurent~Valentin Jospin, Hamid Laga, Farid Boussaid, Wray Buntine, and Mohammed Bennamoun.
\newblock Hands-on bayesian neural networks—a tutorial for deep learning users.
\newblock \emph{IEEE Computational Intelligence Magazine}, 17\penalty0 (2):\penalty0 29--48, 2022.

\bibitem[Kadavath et~al.(2022)Kadavath, Conerly, Askell, Henighan, Drain, Perez, Schiefer, Hatfield-Dodds, DasSarma, Tran-Johnson, et~al.]{kadavath2022language}
Saurav Kadavath, Tom Conerly, Amanda Askell, Tom Henighan, Dawn Drain, Ethan Perez, Nicholas Schiefer, Zac Hatfield-Dodds, Nova DasSarma, Eli Tran-Johnson, et~al.
\newblock Language models (mostly) know what they know.
\newblock \emph{arXiv preprint arXiv:2207.05221}, 2022.

\bibitem[Kojima et~al.(2022)Kojima, Gu, Reid, Matsuo, and Iwasawa]{kojima2022large}
Takeshi Kojima, Shixiang~Shane Gu, Machel Reid, Yutaka Matsuo, and Yusuke Iwasawa.
\newblock Large language models are zero-shot reasoners.
\newblock \emph{Advances in neural information processing systems}, 35:\penalty0 22199--22213, 2022.

\bibitem[Kuhn et~al.(2023)Kuhn, Gal, and Farquhar]{kuhn2023semantic}
Lorenz Kuhn, Yarin Gal, and Sebastian Farquhar.
\newblock Semantic uncertainty: Linguistic invariances for uncertainty estimation in natural language generation.
\newblock \emph{arXiv preprint arXiv:2302.09664}, 2023.

\bibitem[Liang et~al.(2024)Liang, Zhang, and Fisac]{liang2024introspective}
Kaiqu Liang, Zixu Zhang, and Jaime~F Fisac.
\newblock Introspective planning: Aligning robots' uncertainty with inherent task ambiguity.
\newblock \emph{Advances in Neural Information Processing Systems}, 37:\penalty0 71998--72031, 2024.

\bibitem[Ling et~al.(2024)Ling, Zhao, Cheng, Liu, Sun, Zhang, Oishi, Osaki, Matsuda, Ji, et~al.]{ling2024uncertainty}
Chen Ling, Xujiang Zhao, Wei Cheng, Yanchi Liu, Yiyou Sun, Xuchao Zhang, Mika Oishi, Takao Osaki, Katsushi Matsuda, Jie Ji, et~al.
\newblock Uncertainty decomposition and quantification for in-context learning of large language models.
\newblock \emph{arXiv e-prints}, pages arXiv--2402, 2024.

\bibitem[Liu et~al.(2024)Liu, Xue, Chen, Chen, Zhao, Wang, Hou, Li, and Peng]{liu2024survey}
Hanchao Liu, Wenyuan Xue, Yifei Chen, Dapeng Chen, Xiutian Zhao, Ke~Wang, Liping Hou, Rongjun Li, and Wei Peng.
\newblock A survey on hallucination in large vision-language models.
\newblock \emph{arXiv preprint arXiv:2402.00253}, 2024.

\bibitem[Manakul et~al.(2023)Manakul, Liusie, and Gales]{manakul2023selfcheckgpt}
Potsawee Manakul, Adian Liusie, and Mark~JF Gales.
\newblock Selfcheckgpt: Zero-resource black-box hallucination detection for generative large language models.
\newblock \emph{arXiv preprint arXiv:2303.08896}, 2023.

\bibitem[Mielke et~al.(2022)Mielke, Szlam, Dinan, and Boureau]{mielke2022reducing}
Sabrina~J Mielke, Arthur Szlam, Emily Dinan, and Y-Lan Boureau.
\newblock Reducing conversational agents’ overconfidence through linguistic calibration.
\newblock \emph{Transactions of the Association for Computational Linguistics}, 10:\penalty0 857--872, 2022.

\bibitem[{Mikhail Mirzayanov} and {Codeforces Team}(2025)]{Codeforces}
{Mikhail Mirzayanov} and {Codeforces Team}.
\newblock {Codeforces}.
\newblock \url{https://codeforces.com}, 2025.
\newblock Competitive Programming Platform.

\bibitem[Naeini et~al.(2015)Naeini, Cooper, and Hauskrecht]{naeini2015obtaining}
Mahdi~P. Naeini, Gregory Cooper, and Milos Hauskrecht.
\newblock Obtaining well calibrated probabilities using bayesian binning.
\newblock \emph{Proceedings of the AAAI Conference on Artificial Intelligence}, 29, 2015.

\bibitem[OpenAI(2024{\natexlab{a}})]{openai2024gpt4technicalreport}
OpenAI.
\newblock Gpt-4 technical report, 2024{\natexlab{a}}.
\newblock URL \url{https://arxiv.org/abs/2303.08774}.

\bibitem[OpenAI(2024{\natexlab{b}})]{openaio3mini}
OpenAI.
\newblock Openai o3-mini system card.
\newblock Technical report, OpenAI, 2024{\natexlab{b}}.
\newblock URL \url{https://cdn.openai.com/o3-mini-system-card.pdf}.

\bibitem[Rein et~al.(2023)Rein, Hou, Stickland, Petty, Pang, Dirani, Michael, and Bowman]{rein2023gpqa}
David Rein, Betty~Li Hou, Asa~Cooper Stickland, Jackson Petty, Richard~Yuanzhe Pang, Julien Dirani, Julian Michael, and Samuel~R Bowman.
\newblock Gpqa: A graduate-level google-proof q\&a benchmark.
\newblock \emph{arXiv preprint arXiv:2311.12022}, 2023.

\bibitem[Shorinwa et~al.(2025)Shorinwa, Mei, Lidard, Ren, and Majumdar]{10.1145/3744238}
Ola Shorinwa, Zhiting Mei, Justin Lidard, Allen~Z. Ren, and Anirudha Majumdar.
\newblock A survey on uncertainty quantification of large language models: Taxonomy, open research challenges, and future directions.
\newblock \emph{ACM Comput. Surv.}, June 2025.
\newblock ISSN 0360-0300.
\newblock \doi{10.1145/3744238}.
\newblock URL \url{https://doi.org/10.1145/3744238}.

\bibitem[Shynkaruk and Thompson(2006)]{shynkaruk2006confidence}
Jody~M Shynkaruk and Valerie~A Thompson.
\newblock Confidence and accuracy in deductive reasoning.
\newblock \emph{Memory \& cognition}, 34\penalty0 (3):\penalty0 619--632, 2006.

\bibitem[Smith et~al.(2020)Smith, Gonzalez-Rico, Dinan, and Boureau]{smith2020controlling}
Eric~Michael Smith, Diana Gonzalez-Rico, Emily Dinan, and Y-Lan Boureau.
\newblock Controlling style in generated dialogue.
\newblock \emph{arXiv preprint arXiv:2009.10855}, 2020.

\bibitem[Tamkin et~al.(2023)Tamkin, Taufeeque, and Goodman]{tamkin2023codebook}
Alex Tamkin, Mohammad Taufeeque, and Noah~D Goodman.
\newblock Codebook features: Sparse and discrete interpretability for neural networks.
\newblock \emph{arXiv preprint arXiv:2310.17230}, 2023.

\bibitem[Tanneru et~al.(2024)Tanneru, Agarwal, and Lakkaraju]{tanneru2024quantifying}
Sree~Harsha Tanneru, Chirag Agarwal, and Himabindu Lakkaraju.
\newblock Quantifying uncertainty in natural language explanations of large language models.
\newblock In \emph{International Conference on Artificial Intelligence and Statistics}, pages 1072--1080. PMLR, 2024.

\bibitem[Team et~al.(2024)Team, Georgiev, Lei, Burnell, Bai, Gulati, Tanzer, Vincent, Pan, Wang, et~al.]{team2024gemini}
Gemini Team, Petko Georgiev, Ving~Ian Lei, Ryan Burnell, Libin Bai, Anmol Gulati, Garrett Tanzer, Damien Vincent, Zhufeng Pan, Shibo Wang, et~al.
\newblock Gemini 1.5: Unlocking multimodal understanding across millions of tokens of context.
\newblock \emph{arXiv preprint arXiv:2403.05530}, 2024.

\bibitem[Team(2025)]{qwq32b}
Qwen Team.
\newblock Qwq-32b: Embracing the power of reinforcement learning, March 2025.
\newblock URL \url{https://qwenlm.github.io/blog/qwq-32b/}.

\bibitem[Tian et~al.(2023)Tian, Mitchell, Zhou, Sharma, Rafailov, Yao, Finn, and Manning]{tian2023just}
Katherine Tian, Eric Mitchell, Allan Zhou, Archit Sharma, Rafael Rafailov, Huaxiu Yao, Chelsea Finn, and Christopher~D Manning.
\newblock Just ask for calibration: Strategies for eliciting calibrated confidence scores from language models fine-tuned with human feedback.
\newblock \emph{arXiv preprint arXiv:2305.14975}, 2023.

\bibitem[Wang et~al.(2022)Wang, Wei, Schuurmans, Le, Chi, Narang, Chowdhery, and Zhou]{wang2022self}
Xuezhi Wang, Jason Wei, Dale Schuurmans, Quoc Le, Ed~Chi, Sharan Narang, Aakanksha Chowdhery, and Denny Zhou.
\newblock Self-consistency improves chain of thought reasoning in language models.
\newblock \emph{arXiv preprint arXiv:2203.11171}, 2022.

\bibitem[Wei et~al.(2022)Wei, Wang, Schuurmans, Bosma, Xia, Chi, Le, Zhou, et~al.]{wei2022chain}
Jason Wei, Xuezhi Wang, Dale Schuurmans, Maarten Bosma, Fei Xia, Ed~Chi, Quoc~V Le, Denny Zhou, et~al.
\newblock Chain-of-thought prompting elicits reasoning in large language models.
\newblock \emph{Advances in neural information processing systems}, 35:\penalty0 24824--24837, 2022.

\bibitem[Wei et~al.(2024)Wei, Karina, Chung, Jiao, Papay, Glaese, Schulman, and Fedus]{wei2024measuring}
Jason Wei, Nguyen Karina, Hyung~Won Chung, Yunxin~Joy Jiao, Spencer Papay, Amelia Glaese, John Schulman, and William Fedus.
\newblock Measuring short-form factuality in large language models.
\newblock \emph{arXiv preprint arXiv:2411.04368}, 2024.

\bibitem[Wu et~al.(2024)Wu, Qiu, Ross, Aky{\"u}rek, Chen, Wang, Kim, Andreas, and Kim]{wu2024reasoning}
Zhaofeng Wu, Linlu Qiu, Alexis Ross, Ekin Aky{\"u}rek, Boyuan Chen, Bailin Wang, Najoung Kim, Jacob Andreas, and Yoon Kim.
\newblock Reasoning or reciting? exploring the capabilities and limitations of language models through counterfactual tasks.
\newblock In \emph{Proceedings of the 2024 Conference of the North American Chapter of the Association for Computational Linguistics: Human Language Technologies (Volume 1: Long Papers)}, pages 1819--1862, 2024.

\bibitem[Xiao and Wang(2021)]{xiao2021hallucination}
Yijun Xiao and William~Yang Wang.
\newblock On hallucination and predictive uncertainty in conditional language generation.
\newblock \emph{arXiv preprint arXiv:2103.15025}, 2021.

\bibitem[Xiong et~al.(2023)Xiong, Hu, Lu, Li, Fu, He, and Hooi]{xiong2023can}
Miao Xiong, Zhiyuan Hu, Xinyang Lu, Yifei Li, Jie Fu, Junxian He, and Bryan Hooi.
\newblock Can llms express their uncertainty? an empirical evaluation of confidence elicitation in llms.
\newblock \emph{arXiv preprint arXiv:2306.13063}, 2023.

\bibitem[Yao et~al.(2023)Yao, Yu, Zhao, Shafran, Griffiths, Cao, and Narasimhan]{yao2023tree}
Shunyu Yao, Dian Yu, Jeffrey Zhao, Izhak Shafran, Tom Griffiths, Yuan Cao, and Karthik Narasimhan.
\newblock Tree of thoughts: Deliberate problem solving with large language models.
\newblock \emph{Advances in neural information processing systems}, 36:\penalty0 11809--11822, 2023.

\bibitem[Yun et~al.(2021)Yun, Chen, Olshausen, and LeCun]{yun2021transformer}
Zeyu Yun, Yubei Chen, Bruno~A Olshausen, and Yann LeCun.
\newblock Transformer visualization via dictionary learning: contextualized embedding as a linear superposition of transformer factors.
\newblock \emph{arXiv preprint arXiv:2103.15949}, 2021.

\bibitem[Zhang et~al.(2024)Zhang, Diao, Lin, Fung, Lian, Wang, Chen, Ji, and Zhang]{zhang2024r}
Hanning Zhang, Shizhe Diao, Yong Lin, Yi~Fung, Qing Lian, Xingyao Wang, Yangyi Chen, Heng Ji, and Tong Zhang.
\newblock R-tuning: Instructing large language models to say ‘i don’t know’.
\newblock In \emph{Proceedings of the 2024 Conference of the North American Chapter of the Association for Computational Linguistics: Human Language Technologies (Volume 1: Long Papers)}, pages 7106--7132, 2024.

\bibitem[Zhang et~al.(2022)Zhang, Zhang, Li, and Smola]{zhang2022automatic}
Zhuosheng Zhang, Aston Zhang, Mu~Li, and Alex Smola.
\newblock Automatic chain of thought prompting in large language models.
\newblock \emph{arXiv preprint arXiv:2210.03493}, 2022.

\bibitem[Zhou et~al.(2022)Zhou, Sch{\"a}rli, Hou, Wei, Scales, Wang, Schuurmans, Cui, Bousquet, Le, et~al.]{zhou2022least}
Denny Zhou, Nathanael Sch{\"a}rli, Le~Hou, Jason Wei, Nathan Scales, Xuezhi Wang, Dale Schuurmans, Claire Cui, Olivier Bousquet, Quoc Le, et~al.
\newblock Least-to-most prompting enables complex reasoning in large language models.
\newblock \emph{arXiv preprint arXiv:2205.10625}, 2022.

\end{thebibliography}

\clearpage
\beginappendix{

\section{Introspective Uncertainty Quantification}
\label{sec:app_evaluations_introspective_uq}
Here, we provide the full system prompts for the three introspective UQ prompt strategies: IUQ-Low, IUQ-Medium, and IUQ-High.

\begin{tcolorbox}[promptstyle_prompt_iuq, title=Introspective UQ-Low]
You are provided with the reasoning trace of a model asked to answer a question 
and provide the associated confidence between 0 and 100.
Your task is to think about the reasoning trace from the first 
model and provide your confidence in the correctness of the 
answer provided by the first model.
\end{tcolorbox}

\begin{tcolorbox}[promptstyle_prompt_iuq, title=Introspective UQ-Medium]
You are provided with the reasoning trace of a 
model asked to answer a question 
and provide the associated confidence between 0 and 100. 
Your task is to identify the flaws in the reasoning trace 
from the first model and provide your confidence in the correctness of the 
answer provided by the first model.
\end{tcolorbox}

\begin{tcolorbox}[promptstyle_prompt_iuq, title=Introspective UQ-High]
You are provided with the reasoning trace of a 
model asked to answer a question. 
Your task is to identify the flaws in the reasoning trace 
from the first model and provide your confidence in the correctness of the 
answer provided by the first model.
\end{tcolorbox}

Lastly, we provide an example of the full prompt with the question and reasoning trace using IUQ-Medium. We append the text ``You must include your reasoning trace in your response," particularly for models that do not natively support access to the internal reasoning trace.

\begin{tcolorbox}[promptstyle_prompt_iuq, title=Introspective UQ: Example Full Prompt]
\textbf{System:}
You are a smart entity. Do not use the internet.

\textbf{System:}
You are provided with the reasoning trace of a 
model asked to answer a question 
and provide the associated confidence between 0 and 100. 
Your task is to identify the flaws in the reasoning trace 
from the first model and provide your confidence in the correctness of the 
answer provided by the first model.

\textbf{User:}
The question is: How many original artworks were shown in the Society of Illustrators' first "Annual Exhibition"?

The given reasoning process and answer is: ...

\textbf{System:}
Given the answer and confidence from the first model, 
what is your confidence in the validity of the 
first model's answer 
as a number in the form \{rec, Y\}? 
The first element in the curly braces should always be rec. 
For example, if you are 80 percent confident 
that the first model is correct, your answer should be 
\{rec, 80\}.
\end{tcolorbox}

\section{LLM Prompt Strategies}
\label{sec:app_evaluations_prompts}
For completeness, we provide the full prompts for the basic, chain-of-thought, multi-step, and top-K prompt strategies, introduced in prior work.

\begin{tcolorbox}[promptstyle_prompt_iuq, title=Standard Prompt]
Provide an answer and the confidence in your answer between 0 and 100.
\end{tcolorbox}

\begin{tcolorbox}[promptstyle_prompt_iuq, title=Chain-of-Thought Prompt]
Analyze step by step and provide an answer and the confidence in your answer between 0 and 100.
\end{tcolorbox}

\begin{tcolorbox}[promptstyle_prompt_iuq, title=Multi-Step Prompt]
Break down the problem into multiple steps, and provide an answer to each step, along with a justification and a confidence in each step. Give your final answer with your confidence in your final answer and the confidence in your answer between 0 and 100.
\end{tcolorbox}

\begin{tcolorbox}[promptstyle_prompt_iuq, title=Top-K Prompt]
Give $K=5$ best guesses and your confidence in each guess (i.e., the probability that each one is correct). Give your final answer with your confidence in your final answer and the confidence in your answer between 0 and 100.
\end{tcolorbox}

The system prompts are followed with a description of the dataset including the desired answer format, as well as the actual question. We show an example full prompt for the SimpleQA dataset, with the basic prompt strategy.

\begin{tcolorbox}[promptstyle_prompt_iuq, title=Example Full Prompt]
\textbf{System:}
You are a smart entity. Do not use the internet.

\textbf{System:}
Analyze step by step and provide an answer and the confidence in your answer between 0 and 100.

\textbf{System:}
Each question requires a simple, short fact as the answer: only provide your final answer, which could be a specific name, date, or other fact, and a number for the confidence in the form {X, Y}. For example, if you think the answer is 1950 and you are 80 percent confident, then you should only say \{1950, 80\}.

\textbf{User:}
Who received the IEEE Frank Rosenblatt Award in 2010?
\end{tcolorbox}

\section{Calibration of Reasoning Models.}
\label{sec:app_evaluations_calib_reasoning_models}
We further discuss the calibration and accuracy of reasoning models, across the basic, chain-of-thought, multi-step, and top-K prompt strategies, in addition to the results in \Cref{sec:evaluation}.

\begin{figure*}[th]
    \centering
    \includegraphics[width=0.8\linewidth]{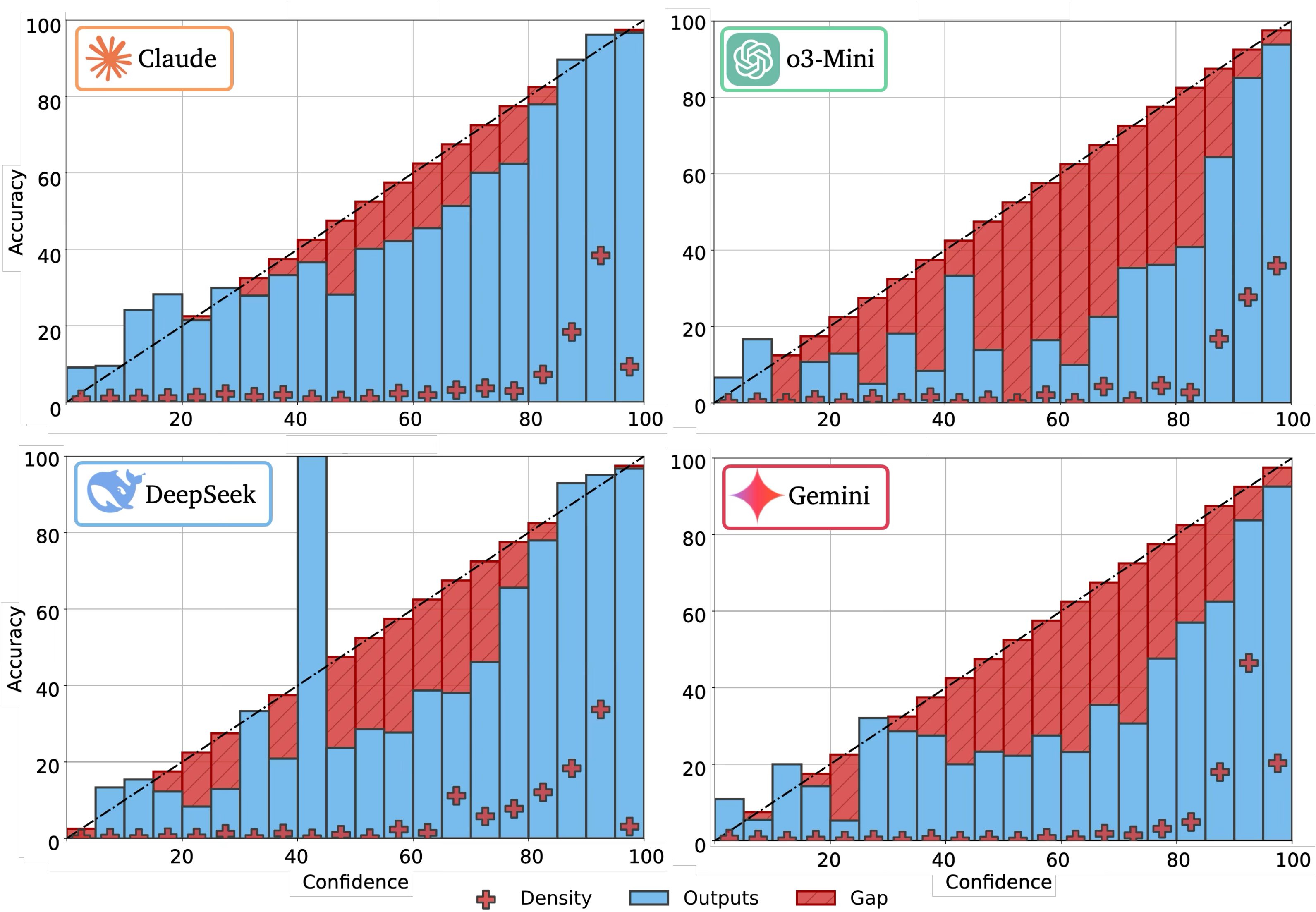}
    \caption{\textbf{Are reasoning models consistently underconfident or overconfident?} The reliability diagrams are aggregated over all prompt strategies and all datasets. Reasoning models are overconfident, but it's less obvious in datasets with very high model accuracies.}
    \label{fig:calibration_reliability}
\end{figure*}

\smallskip
\noindent\textbf{Are Reasoning Models Underconfident or Overconfident?}
In~\Cref{fig:calibration_reliability}, we plot the reliability diagrams aggregated across all datasets, showing that reasoning models are typically highly confident, with most of their confidence estimates in the $90\%$-$100\%$ confidence bins. However, the high accuracies of the models makes overconfidence detection difficult. For example, in \Cref{fig:reliability_easy}, we visualize the reliability diagrams for the easier datasets (ARC-Challenge and MMLU), showing that overconfidence is masked by the models' high accuracy.

\begin{figure*}
    \centering
    \includegraphics[width=0.8\linewidth]{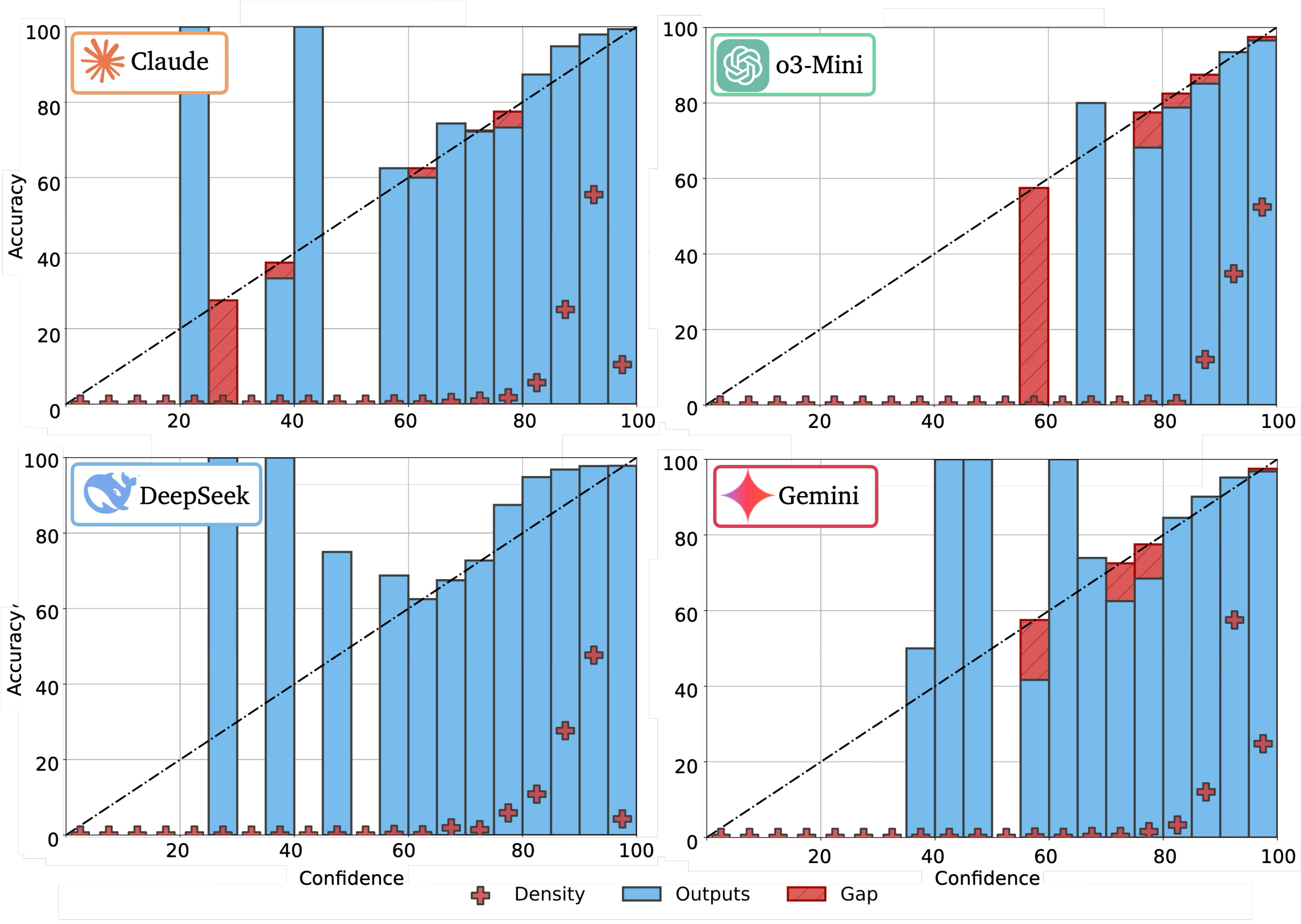}
    \caption{\textbf{Are reasoning models consistently underconfident or overconfident?} We show the reliability diagrams on the easier datasets . Overconfidence is difficult to detect in these datasets due to the high accuracy of the models.}
    \label{fig:reliability_easy}
\end{figure*}

\smallskip
\noindent\textbf{Model Calibration for Each Prompt Strategy.}
In \Cref{fig:calibration_vs_model_standard,fig:calibration_vs_model_cot,fig:calibration_vs_model_multistep,fig:calibration_vs_model_topk}, we provide the calibration and accuracies of each model across all datasets for each prompt strategy. In line with the results in \Cref{ssec:calibration_reasoning_models}, across all the prompt strategies, Claude achieves the lowest ECE by relatively significant margins followed by DeepSeek (except in the top-K prompt strategy). In general, we find that Claude 3.7 Sonnet is the best calibrated and most accurate of the reasoning models considered in this work. Further, Gemini and o3-Mini achieve almost the same calibration and accuracy, with Gemini outperforming o3-Mini with basic and top-K prompts by narrow margins.

\begin{figure*}
    \centering
    \includegraphics[width=\linewidth]{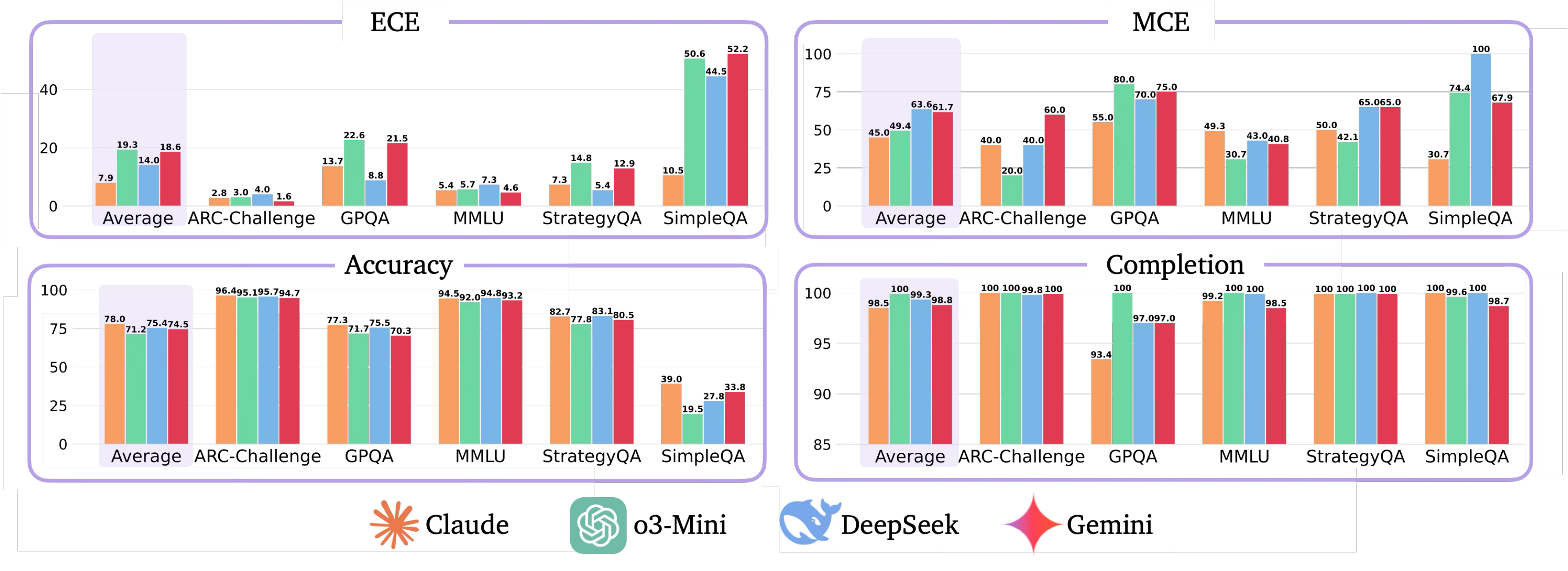}
    \caption{\textbf{Basic Prompt Strategy.} Calibration and Accuracy of Reasoning Models.}
    \label{fig:calibration_vs_model_standard}
\end{figure*}

\begin{figure*}
    \centering
    \includegraphics[width=\linewidth]{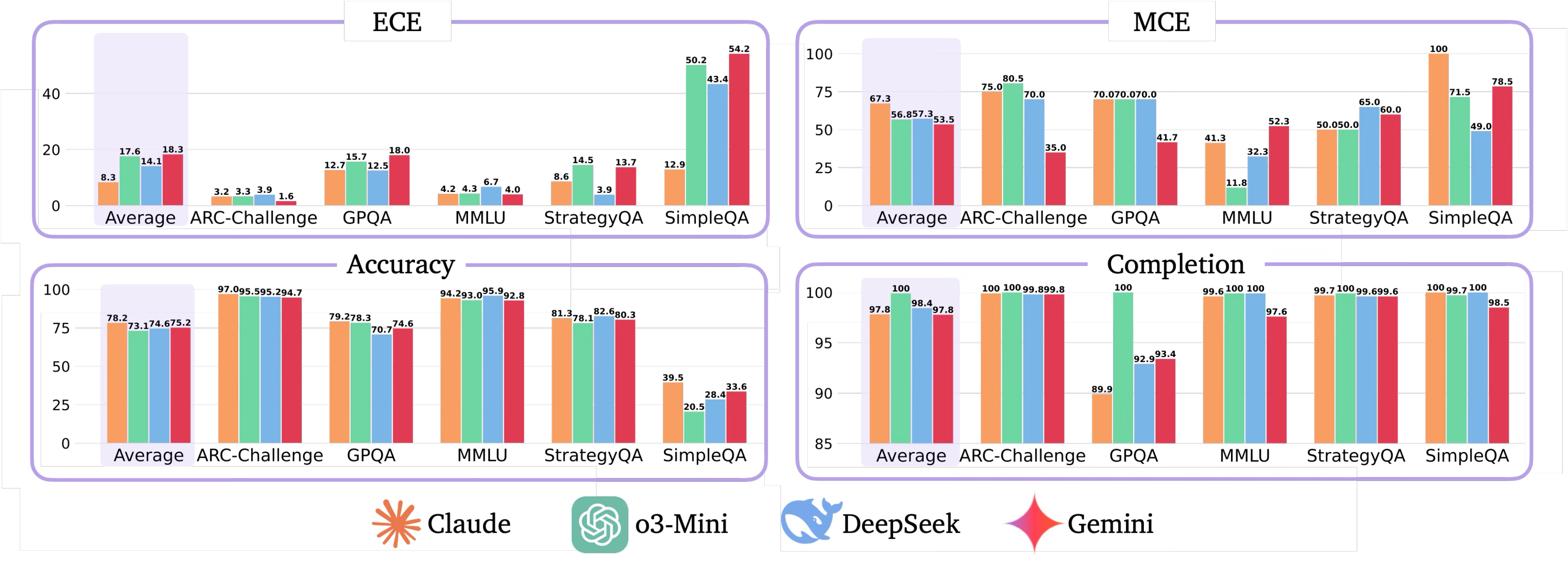}
    \caption{\textbf{Chain-of-Thought Prompt Strategy.} Calibration and Accuracy of Reasoning Models.}
    \label{fig:calibration_vs_model_cot}
\end{figure*}

\begin{figure*}
    \centering
    \includegraphics[width=\linewidth]{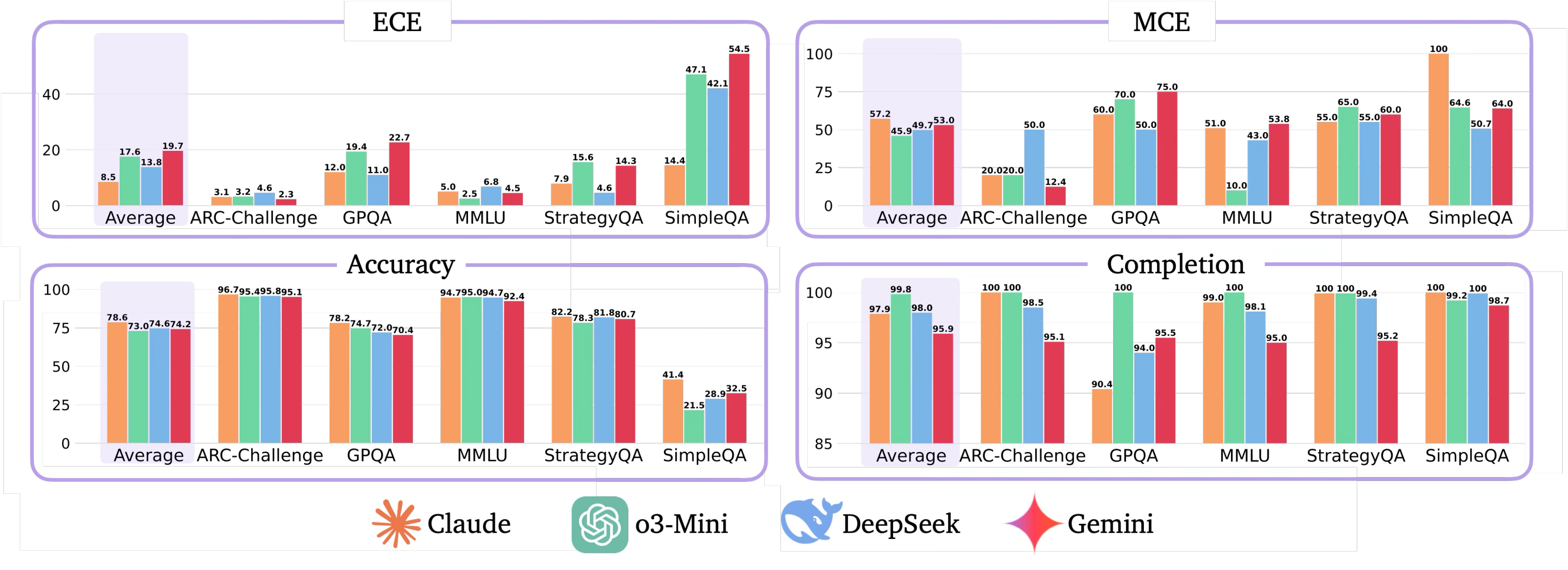}
    \caption{\textbf{Multi-Step Prompt Strategy.} Calibration and Accuracy of Reasoning Models.}
    \label{fig:calibration_vs_model_multistep}
\end{figure*}

\begin{figure*}
    \centering
    \includegraphics[width=\linewidth]{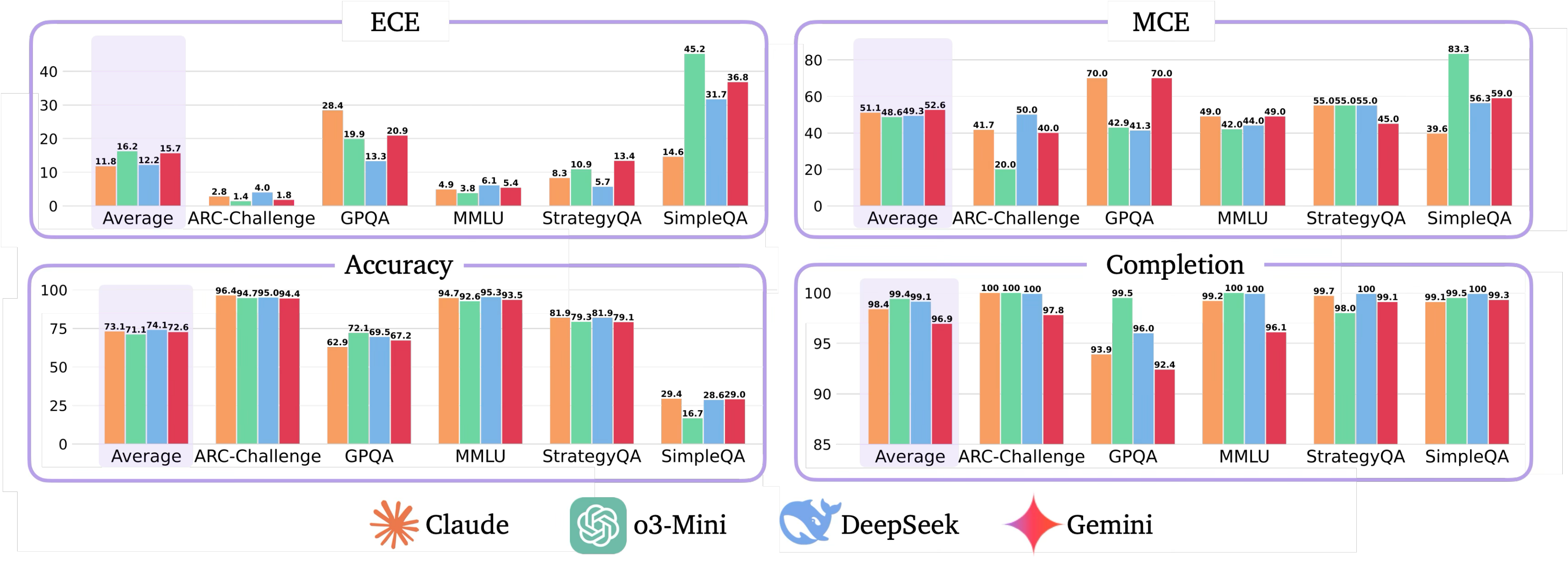}
    \caption{\textbf{Top-K Prompt Strategy.} Calibration and Accuracy of Reasoning Models.}
    \label{fig:calibration_vs_model_topk}
\end{figure*}

\section{Coupling between Accuracy and Calibration of Reasoning Models}
\label{sec:app_evaluations_accuracy_calib_vs_reasoning}
Here, we discuss the accuracy of each model on the benchmarks and examine the correlation between accuracy and calibration.

\smallskip
\noindent\textbf{Model Accuracy on Benchmarks.}
We find that the accuracy of reasoning models varies significantly across the benchmarks. In \Cref{fig:calibration_reasoning_models}, all reasoning models achieve over $90\%$ accuracy on the ARC-Challenge and MMLU datasets, with about a $10-20\%$ absolute drop on the StrategyQA and GPQA datasets. The drop in accuracy can be explained by the relative difficulty of the more recent benchmarks StrategyQA and GPQA, which emphasize analytical problem-solving over memorization, compared to ARC-Challenge and MMLU. Prior work has shown that LLMs struggle more with analytical tasks compared to information-retrieval tasks (which require recitation) \cite{wu2024reasoning}, which is in line with our results. 
Likewise, the results indicate that the accuracy of the reasoning models decreases with the recency of the benchmark, which is not surprising, given that these models are trained to perform well on these benchmarks.
Notably, all reasoning models achieve less than $40\%$ accuracy on the SimpleQA dataset, which was was constructed adversarially from GPT-4's responses (i.e., a question was added to SimpleQA, if at least one GPT-4 response was incorrect). We observe that SimpleQA remains notably challenging for reasoning models, making it useful for uncertainty quantification.

\smallskip
\noindent\textbf{Correlation between Accuracy and Calibration.}
\begin{figure*}
    \centering
    \includegraphics[width=\linewidth]{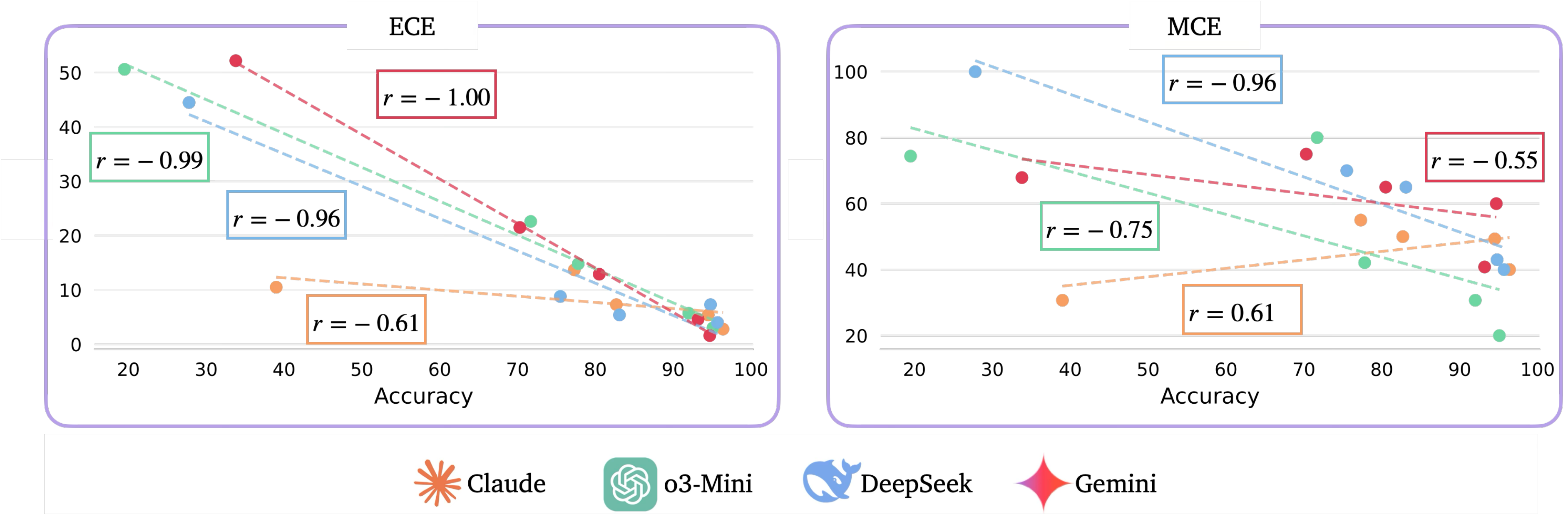}
    \caption{\textbf{Basic Prompt Strategy.} Correlation between Calibration and Accuracy.}
    \label{fig:correlation_standard}
\end{figure*}
\begin{figure*}
    \centering
    \includegraphics[width=\linewidth]{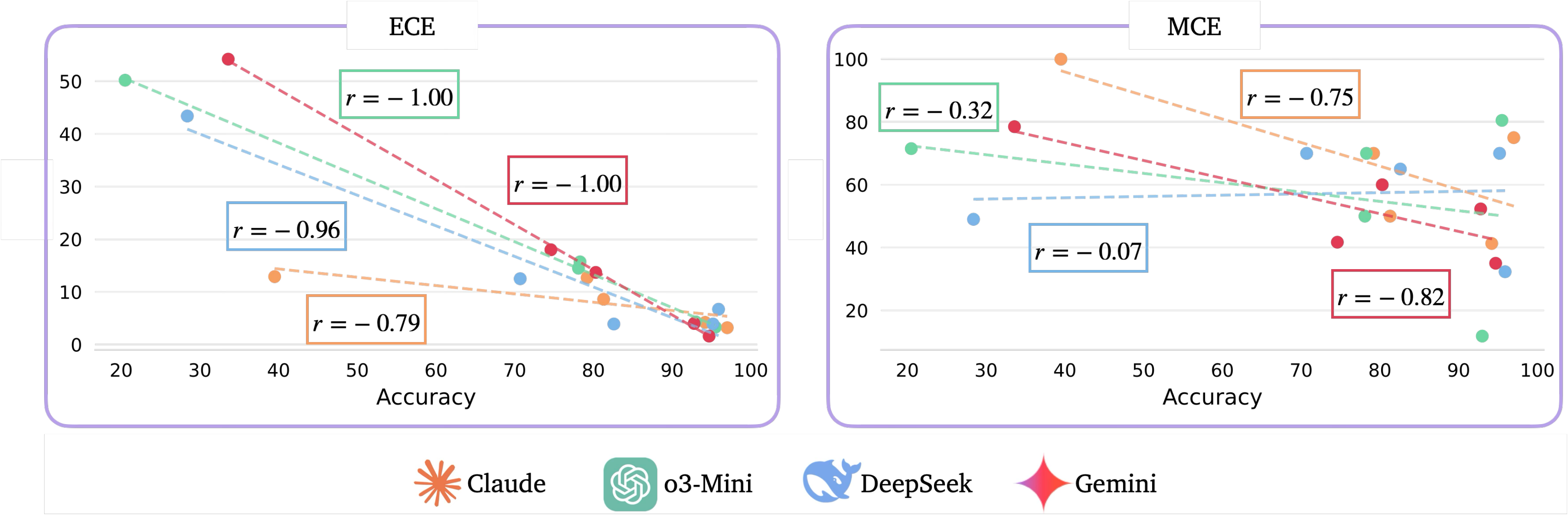}
    \caption{\textbf{Chain-of-Thought Prompt Strategy.} Correlation between Calibration and Accuracy.}
    \label{fig:correlation_cot}
\end{figure*}
\begin{figure*}
    \centering
    \includegraphics[width=\linewidth]{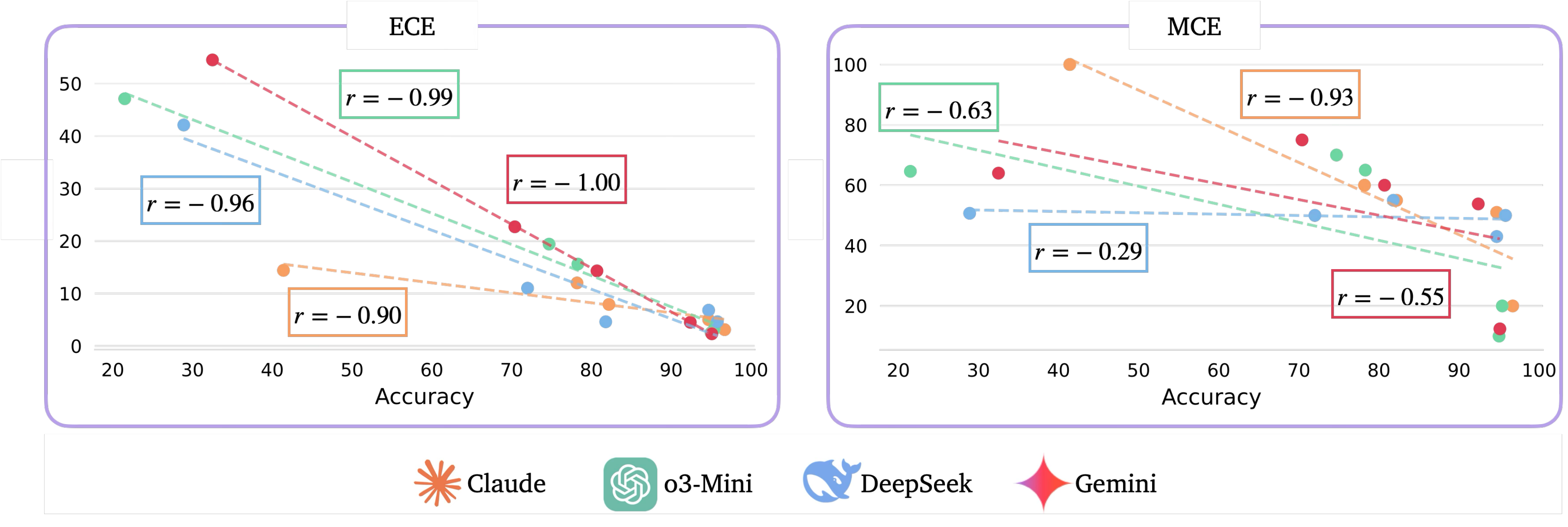}
    \caption{\textbf{Multi-Step Prompt Strategy.} Correlation between Calibration and Accuracy.}
    \label{fig:correlation_multistep}
\end{figure*}
\begin{figure*}
    \centering
    \includegraphics[width=\linewidth]{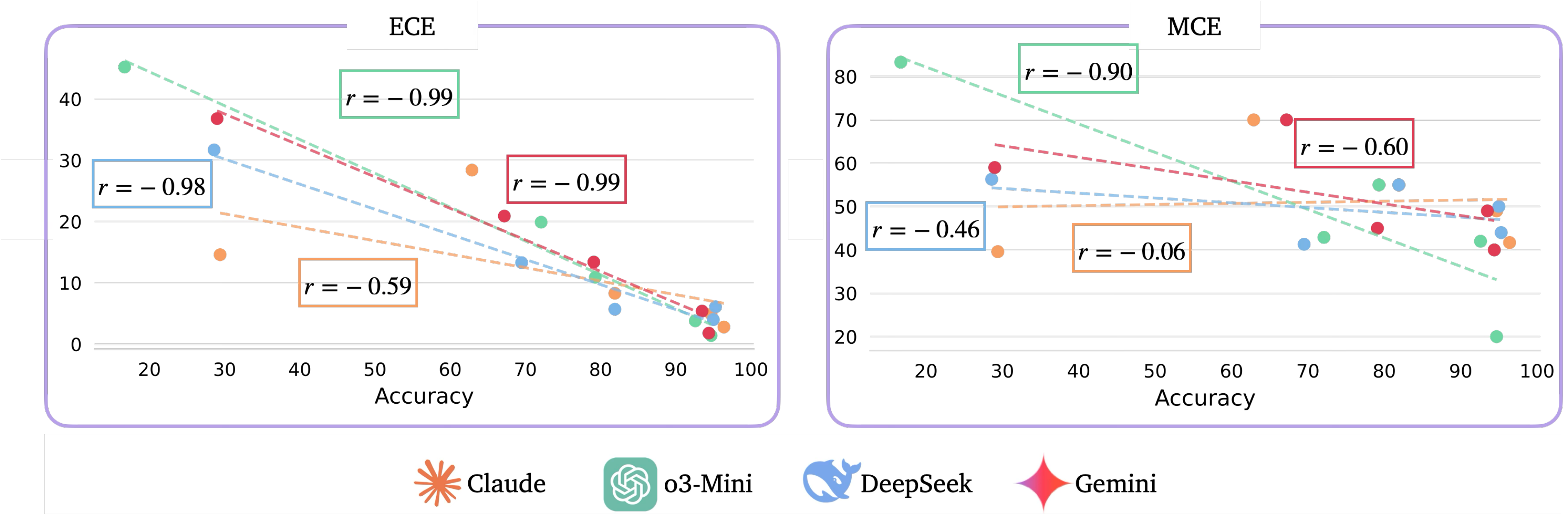}
    \caption{\textbf{Top-K Prompt Strategy.} Correlation between Calibration and Accuracy.}
    \label{fig:correlation_topk}
\end{figure*}
In~\Cref{ssec:acc_vs_calib_reasoning_models}, we show that calibration and accuracy are strongly correlated (\Cref{fig:accuracy_vs_calibration}), with the results averaged over the four prompt strategies. Here, we show the correlation between calibration errors and accuracy for each prompt strategy in Figures~\ref{fig:correlation_standard} through~\ref{fig:correlation_topk}. For all prompt strategies, we observe a strong negative correlation between ECE and accuracy, as well as between MCE and accuracy. Similar to results discussed in~\Cref{ssec:acc_vs_calib_reasoning_models}, among all models, Claude's calibration is least strongly correlated accuracy, while Gemini almost always exhibits perfect negative correlation. For MCE and accuracy, the trend is weaker, affected by many outliers, but o3-Mini and Gemini generally show stronger correlations compared to Claude and DeepSeek.

\section{Effects of LLM Prompt Strategies on Calibration and Accuracy of Reasoning Models}
\label{sec:app_evaluations_calib_accuracy_vs_prompt}
As discussed in \Cref{ssec:acc_calib_vs_prompt_reasoning_models}, on average, existing prompt strategies do not significantly affect the performance of reasoning models. Here, we examine the variability in calibration and accuracy for each model for each prompt strategy.
Unlike the other models, Claude suffers a notable dip in its accuracy and calibration in the Top-K strategy, where the model is asked to provide multiple guesses along with its confidence in the correctness of each guess. This finding suggests that Claude might be more likely to change its answers to potentially incorrect ones when asked to reason about false statements/claims. However, Claude still outperforms the other models in the Top-K prompt strategy, albeit by much smaller margins.
In fact, we observe that GPT, DeepSeek, and Gemini achieve lower calibration errors with Top-K prompts compared to the other prompts, suggesting that explicitly reasoning about alternative answers improves the calibration of these models even though the accuracy remains relatively constant.
Lastly, \Cref{fig:calibration_vs_prompt_gpt} shows that chain-of-thought and multi-step prompting provide a slight boost to the accuracy of o3-Mini.
Overall, we reiterate that the specialized prompt strategies seem to have minimal effect on the calibration and accuracy of reasoning models.

\begin{figure*}
    \centering
    \includegraphics[width=\linewidth]{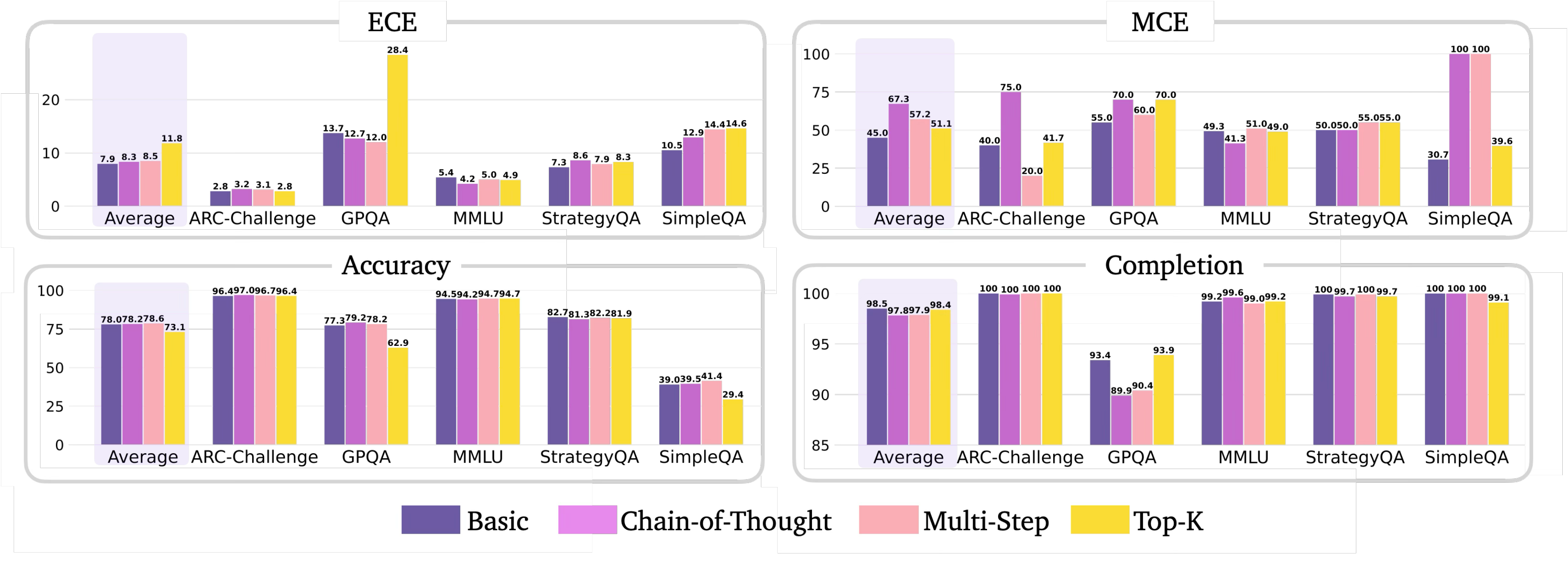}
    \caption{\textbf{Claude 3.7 Sonnet.} Calibration and accuracy for each prompt strategy.}
    \label{fig:calibration_vs_prompt_claude}
\end{figure*}

\begin{figure*}
    \centering
    \includegraphics[width=\linewidth]{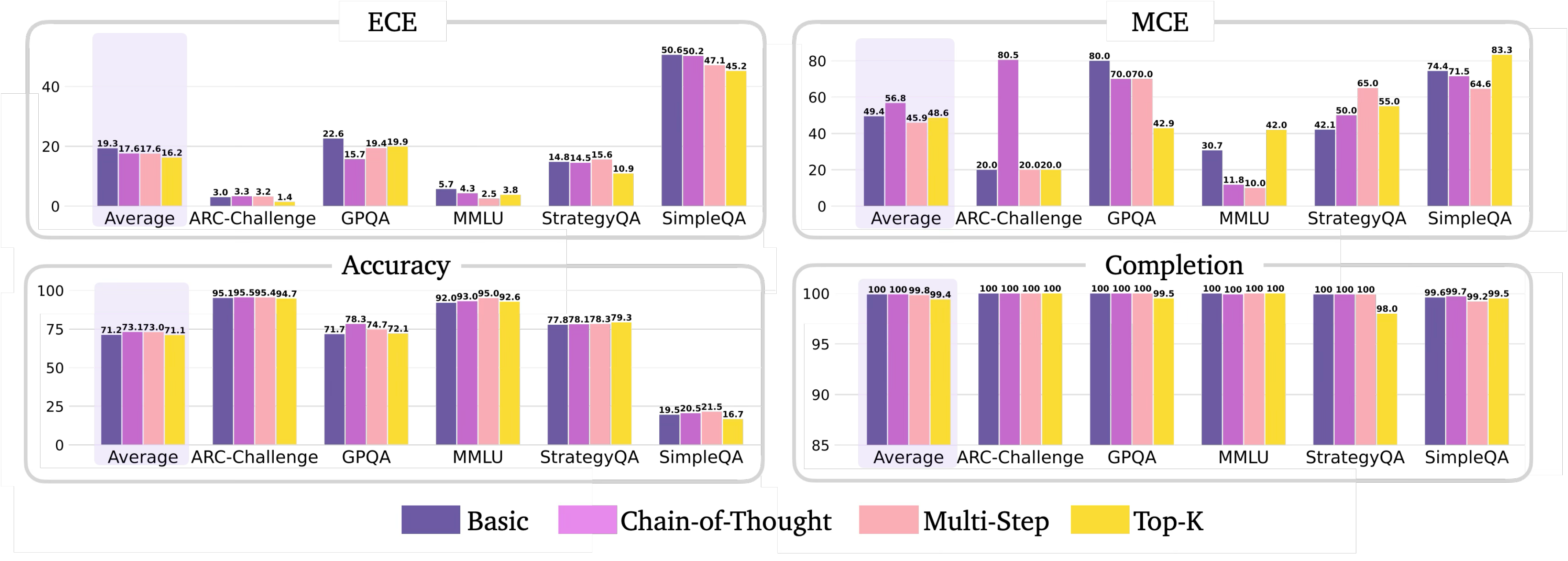}
    \caption{\textbf{o3-Mini.} Calibration and accuracy for each prompt strategy.}
    \label{fig:calibration_vs_prompt_gpt}
\end{figure*}

\begin{figure*}
    \centering
    \includegraphics[width=\linewidth]{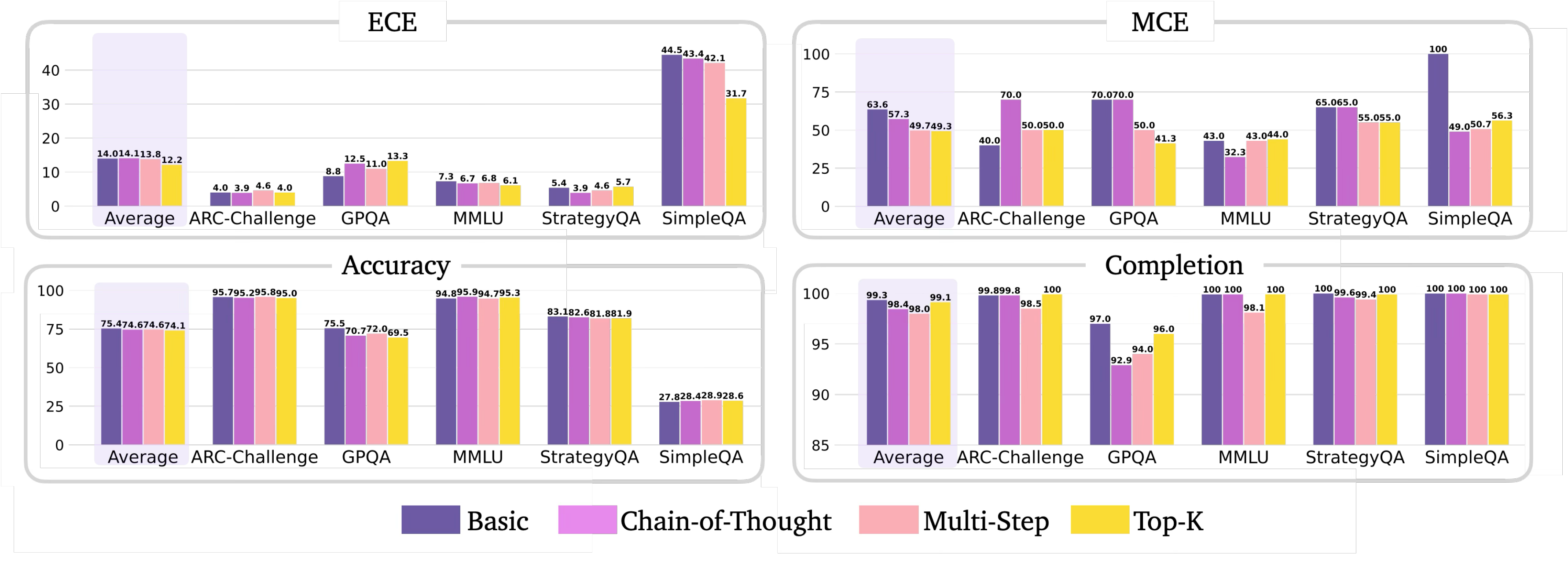}
    \caption{\textbf{DeepSeek R1.} Calibration and accuracy for each prompt strategy. %
    }
    \label{fig:calibration_vs_prompt_deepseek}
\end{figure*}

\begin{figure*}
    \centering
    \includegraphics[width=\linewidth]{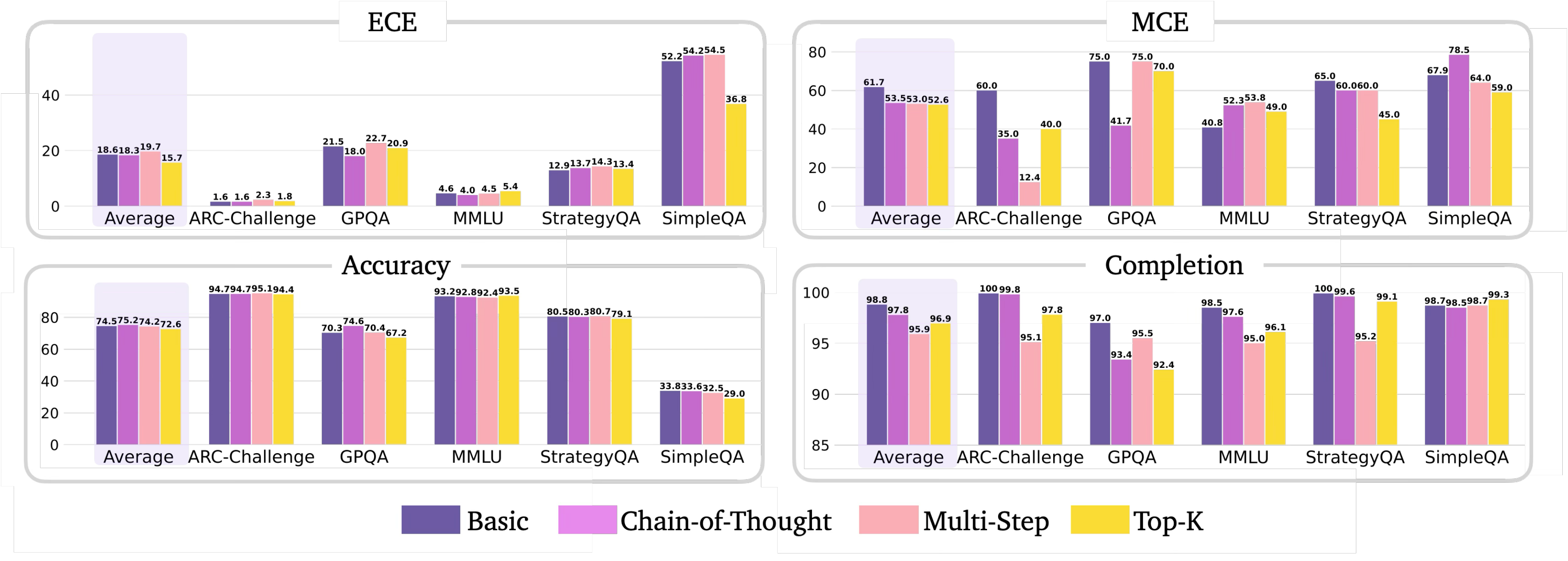}
    \caption{\textbf{Gemini 2.0 Flash-Thinking-Exp.} Calibration and accuracy for each prompt strategy.}
    \label{fig:calibration_vs_prompt_gemini}
\end{figure*}

\section{Introspective UQ Reasoning Trace Examples}
\label{sec:app_evaluations_introspective_uq_exs}
Here, we provide some examples of the prompts, reasoning traces, and answers provided by Claude, DeepSeek, and o3-Mini for all introspective prompts in the challenging datasets (GPQA, StrategyQA, and SimpleQA). All models get the GPQA question wrong and are highly overconfident, demonstrating poor calibration. On the relatively easier dataset StrategyQA, all models provide highly confident correct responses (overconfidence is masked by high accuracy in these problems). In SimpleQA where all models answer the question incorrectly, introspection provide better calibrated confidence estimates, especially in the case of o3-Mini, although DeepSeek reamins overconfident.

\subsection{GPQA}
\begin{tcolorbox}[promptstyle_prompt_iuq, title=Question and Ground-Truth Answer]
Q: 
The Cope rearrangement is a chemical reaction where a 1,5-diene molecule undergoes rearrangement, resulting in a change in the positions of its carbon-carbon double bonds. This rearrangement can be initiated by heat or light and is valuable for creating complex organic compounds with changed structures.

\vspace{1em}

\noindent \textbf{Select the major products from the following rearrangements:}

1. (((3-methylbut-2-en-1-yl)oxy)methyl)benzene + (1. BuLi, 2. H+) ----> A

2. 3,4,5,7,8,9-hexamethyl-1,11-dimethylene-2,6,10,11,11a,11b-hexahydro-1H-benzo

[cd]indeno[7,1-gh]azulene + Heat ---> B.

\vspace{1em}

\noindent \textbf{Options:}

\textbf{(A)} 
    $A = 4$-methyl-1-phenylpent-3-en-1-ol 
    
    $B = 2,3,4,6,7,8$-hexamethyl-5,9,9a,10,11,11a-hexahydro-1H-benzo
    
    [3,4]azuleno[1,8,7,6-cdef]fluorene

\textbf{(B)} 
    $A = (Z)$-2-methyl-5-phenylpent-2-en-1-ol 
    
    $B = 2,3,4,6,7,8$-hexamethyl-5,9,10,11-tetrahydro-1H-benzo
    
    [3,4]azuleno[1,8,7,6-cdef]fluorine

\textbf{(C)} 
    $A = (Z)$-2-methyl-5-phenylpent-2-en-1-ol \\
    $B = 2,3,4,6,7,8$-hexamethyl-5,9,9a,10,11,11a-hexahydro-1H-benzo
    
    [3,4]azuleno[1,8,7,6-cdef]fluorene

\textbf{(D)} 
    $A = 4$-methyl-1-phenylpent-3-en-1-ol 
    
    $B = 2,3,4,6,7,8$-hexamethyl-5,9,10,11-tetrahydro-1H-benzo
    
    [3,4]azuleno[1,8,7,6-cdef]fluorine
\vspace{1em}

A: D
\end{tcolorbox}

\begin{tcolorbox}[promptstyle_o3_mini, title=o3-Mini]
\textcolor{Dandelion}{First round:}

\textbf{Reasoning Trace:}

\begin{enumerate}
    \item For the first reaction, the starting benzyl allyl ether, \(((3\text{-} \mathrm{methylbut}\text{-}2\text{-} \mathrm{en}\text{-}1\text{-} \mathrm{yl})\mathrm{oxy})\mathrm{methyl}\mathrm{benzene}\), under BuLi/\(\text{H}^+\) conditions undergoes a rearrangement that effectively ``umpoled'' the benzyl fragment. Analysis of the connectivity shows that nucleophilic attack leads to a five-carbon chain where the benzyl group ends up at one end. Distinguishing between the possible products, the structure where the new double bond is positioned between C2–C3 (and with a methyl substituent at C2) -- i.e., \((\text{Z})\)-2-methyl-5-phenylpent-2-en-1-ol -- fits the mechanistic pathway and stereoelectronic demands.
    
        \item In the second reaction, the substrate is a highly substituted 1,11-dimethylene system undergoing a Cope rearrangement upon heating. Evaluating the potential rearranged structures, the product retaining a ``hexahydro'' system -- specifically, 2,3,4,6,7,8-hexamethyl-5,9,9a,10,11,11a-hexahydro-1H-benzo
        
        [3,4]azuleno[1,8,7,6-cdef]fluorene -- is the more consistent outcome given that the rearrangement does not result in extensive dehydrogenation.
    
    \item Comparing the options, option (C) presents the pair of products that match these conclusions: \((\text{Z})\)-2-methyl-5-phenylpent-2-en-1-ol for product A and the ``hexahydro'' fluorene derivative for product B.
    
    \item Thus, the reasoning leads to the final choice of option C with high confidence.
\end{enumerate}
\textcolor{Dandelion}{\{C, 95\}}
\vspace{1em}

\textcolor{CornflowerBlue}{IUQ-Low:}

Here’s my reasoning trace:

\begin{enumerate}
    \item I carefully reviewed the provided explanation of the two reactions and the selection from the answer choices. The explanation shows that the first reaction produces (Z)-2-methyl-5-phenylpent-2-en-1-ol (product A) by the mechanism described and that the Cope rearrangement in the second reaction leads to a hexahydro fluorene derivative (product B).
    \item The explanation correctly distinguishes between the isomeric outcomes by considering stereoelectronic requirements and the substituent pattern, which aligns with option (C).
    \item The reasoning steps seem well justified, and the final answer from the first model appears internally consistent and chemically plausible.
    \item Based on this, I am highly confident that the first model’s answer is correct with 95 percent certainty.
\end{enumerate}

\textcolor{CornflowerBlue}{\{rec, 95\}}
\vspace{1em}
\end{tcolorbox}

\begin{tcolorbox}[promptstyle_o3_mini, title=o3-Mini (Continued)]
\textcolor{RoyalBlue}{IUQ-Medium: }

Below is my reasoning trace and final evaluation:

\begin{enumerate}
    \item In the first reaction the model argues that the benzyl allyl ether is ``umpoled'' under BuLi/H$^+$ conditions, leading to the formation of (Z)-2-methyl-5-phenylpent-2-en-1-ol. Although the idea of a polarity inversion is mentioned, the trace does not detail why the (Z)-isomer is favored over a possible (E)-isomer. It also glosses over the mechanistic steps (e.g., the formation of reactive intermediates, stereoelectronic factors, or potential competing pathways).
    
    \item For the second reaction (the Cope rearrangement), the model simply argues that among potential rearranged structures the one retaining a ``hexahydro'' framework is more consistent. Again, the reasoning lacks expanded discussion on how subtle changes in unsaturation levels arise from the rearrangement. It does not fully clarify why the product should not undergo further dehydrogenation.
    
    \item Although the final chosen option (C) agrees with the general expectations — (Z)-2-methyl-5-phenylpent-2-en-1-ol (for A) and a hexahydro fluorene derivative (for B) — the chain-of-thought omits several detailed mechanistic arguments that would strengthen the case for selection versus the other close options.
\end{enumerate}

In summary, while the final answer of option C is plausible given what is expected from the reactions, the reasoning trace has weaknesses in its clarity and depth of mechanistic justification.

Based on these observations I am moderately confident in the final answer and reasoning. My final confidence rating is: \{rec, 75\}.

\textcolor{RoyalBlue}{\{rec, 75\}}
\vspace{1em}

\textcolor{Aquamarine}{IUQ-High: }

\textbf{Reasoning Trace:}

\begin{enumerate}
    \item I examined the given reasoning from the first model. It argues that for reaction A the product should be (Z)-2-methyl-5-phenylpent-2-en-1-ol---this is attributed to a rearrangement pathway initiated by BuLi/H$^+$ that ``umpolars'' the benzyl fragment. Although a detailed mechanism was not provided, this outcome (with the (Z) configuration) is in line with a typical allyl rearrangement producing an allylic alcohol.

    \item For reaction B the model argued that the Cope rearrangement of a highly substituted dienic system produces a product that preserves a hexahydro framework, leading to a structure matching 2,3,4,6,7,8-hexamethyl-5,9,9a,10,11,11a-hexahydro-1H-benzo
    
    [3,4]azuleno[1,8,7,6-cdef]fluorene.

    \item Comparing these two outcomes with the provided multiple-choice options, option (C) exactly matches: A is (Z)-2-methyl-5-phenylpent-2-en-1-ol and B is the ``hexahydro'' fluorene derivative.

    \item While the explanation does not detail every mechanistic nuance and leaves some mechanistic assumptions implicit, the overall conclusion appears plausible and consistent with the reaction conditions and structural outcomes described.

    \item Thus, my assessment is that the first model’s answer is reasonably substantiated.
\end{enumerate}

\textcolor{Aquamarine}{\{rec, 80\}}

\end{tcolorbox}

\begin{tcolorbox}[promptstyle_claude, title=Claude]
\textcolor{Dandelion}{First round:}
I need to determine the products of two different reactions.

\textbf{For the first reaction:}
\[
\text{(((3-methylbut-2-en-1-yl)oxy)methyl)benzene} + \left( \begin{array}{l}
\text{1. BuLi} \\
\text{2. H$^+$}
\end{array} \right) \longrightarrow A
\]

This involves a [2,3]-Wittig rearrangement. Let's analyze the starting material:
\begin{itemize}
    \item It's a benzyl ether of 3-methylbut-2-en-1-ol
    \item Structure: Ph-CH$_2$-O-CH$_2$-C(CH$_3$)=CH-CH$_3$
\end{itemize}

The mechanism:
\begin{enumerate}
    \item BuLi deprotonates the benzylic position, creating a carbanion
    \item This carbanion undergoes a [2,3]-sigmatropic rearrangement, forming a new C-C bond with the $\gamma$-carbon of the allylic system
    \item The hydroxyl group forms at the original benzylic position
    \item H$^+$ workup gives the final product
\end{enumerate}

The product is: Ph-CH(OH)-CH$_2$-C(CH$_3$)=CH-CH$_3$, which is \textbf{4-methyl-1-phenylpent-3-en-1-ol}.

\textbf{For the second reaction (Cope rearrangement):}
\begin{itemize}
    \item Cope rearrangements are [3,3]-sigmatropic rearrangements of 1,5-dienes
    \item They preserve the degree of unsaturation (number of double bonds)
    \item The starting material has ``hexahydro'' in its name
    \item After rearrangement, the product should also have ``hexahydro'' (not ``tetrahydro'')
\end{itemize}

Looking at the options, only (A) matches both criteria:
\begin{itemize}
    \item Product A is 4-methyl-1-phenylpent-3-en-1-ol
    \item Product B maintains the ``hexahydro'' descriptor
\end{itemize}

\textcolor{Dandelion}{\{A, 90\}}
\vspace{1em}

\textcolor{CornflowerBlue}{IUQ-Low:}

I'll evaluate the first model's reasoning and provide my confidence in their answer.

For the first reaction involving (((3-methylbut-2-en-1-yl)oxy)methyl)benzene:
\begin{itemize}
    \item The model correctly identified this as a [2,3]-Wittig rearrangement, which is appropriate when a benzyl ether with an allylic system is treated with BuLi.
    \item The mechanistic explanation is accurate: BuLi deprotonates at the benzylic position, followed by sigmatropic rearrangement where the carbanion attacks the $\gamma$-carbon of the allylic system.
    \item The product structure Ph-CH(OH)-CH$_2$-C(CH$_3$)=CH-CH$_3$ corresponds to 4-methyl-1-phenylpent-3-en-1-ol, which appears in options A and D.
\end{itemize}

For the second reaction (Cope rearrangement):
\begin{itemize}
    \item The model correctly identified key principles of Cope rearrangements (they're [3,3]-sigmatropic rearrangements that preserve the degree of unsaturation).
    \item The crucial observation about preserving the ``hexahydro'' descriptor is chemically sound.
    \item Only options A and C maintain the ``hexahydro'' descriptor in product B.
\end{itemize}

When combining these analyses:
\begin{itemize}
    \item Product A (4-methyl-1-phenylpent-3-en-1-ol) appears in options A and D.
    \item Product B with ``hexahydro'' descriptor appears in options A and C.
    \item Therefore, only option A satisfies both requirements.
\end{itemize}

The model's reasoning is thorough and chemically accurate for both reactions. The analysis properly applies organic chemistry principles to identify the correct products.

\textcolor{CornflowerBlue}{\{rec, 95\}}
\vspace{1em}
\end{tcolorbox}

\begin{tcolorbox}[promptstyle_claude, title=Claude (Continued)]
\textcolor{RoyalBlue}{IUQ-Medium: }

To evaluate the first model's reasoning and confidence, I'll analyze each part of the solution.

\bigskip

\textbf{First reaction:} (((3-methylbut-2-en-1-yl)oxy)methyl)benzene + (1. BuLi, 2. H$^+$) $\longrightarrow$ A

The model correctly:
\begin{itemize}
    \item Identified this as a [2,3]-Wittig rearrangement,
    \item Recognized that BuLi deprotonates the benzylic position (most acidic),
    \item Understood the sigmatropic rearrangement mechanism,
    \item Determined the correct structure of the product (4-methyl-1-phenylpent-3-en-1-ol).
\end{itemize}

The mechanism and product determination are chemically sound. The benzyl ether starting material undergoes the expected rearrangement to form the alcohol with the correct connectivity.

\bigskip

\textbf{Second reaction:} Complex polycyclic compound + Heat $\longrightarrow$ B

The model correctly:
\begin{itemize}
    \item Identified this as a Cope rearrangement ([3,3]-sigmatropic),
    \item Applied the key principle that Cope rearrangements preserve the degree of unsaturation,
    \item Used the ``hexahydro'' descriptor to distinguish between possible products,
    \item Selected the product that maintained the same hydrogenation state.
\end{itemize}

This reasoning is correct — a thermal Cope rearrangement would preserve the number of hydrogen atoms, so the ``hexahydro'' descriptor should be maintained in the product.

The model also correctly noted that the product should be a ``fluorene'' derivative (not ``fluorine'' as appears in options B and D, which is likely a typo).

\bigskip

Overall, the model's reasoning is thorough and chemically accurate. I don't detect any significant flaws in the analysis, and the selection of option A is well-justified based on both reactions.

\textcolor{RoyalBlue}{\{rec, 95\}}
\vspace{1em}

\end{tcolorbox}

\begin{tcolorbox}[promptstyle_claude, title=Claude (Continued)]
\textcolor{Aquamarine}{IUQ-High: }

I need to evaluate the first model's reasoning for both reactions to determine my confidence in their answer.

\textbf{First Reaction Analysis:}

The model correctly identifies this as a [2,3]-Wittig rearrangement with:
\begin{itemize}
    \item Starting material: (((3-methylbut-2-en-1-yl)oxy)methyl)benzene (Ph-CH$_2$-O-CH$_2$-C(CH$_3$)=CH-CH$_3$)
    \item BuLi deprotonates the benzylic position
    \item [2,3]-sigmatropic rearrangement occurs
    \item Final product: Ph-CH(OH)-CH$_2$-C(CH$_3$)=CH-CH$_3$ 
    
    (4-methyl-1-phenylpent-3-en-1-ol)
\end{itemize}

This is mechanistically sound. The carbanion formed at the benzylic position rearranges to form a new C-C bond at the $\gamma$-carbon of the allylic system, with the oxygen ending up as a hydroxyl group. This produces 4-methyl-1-phenylpent-3-en-1-ol, not the (Z)-2-methyl-5-phenylpent-2-en-1-ol structure.

\textbf{Second Reaction Analysis:}

The model correctly identifies:
\begin{itemize}
    \item Cope rearrangement is a [3,3]-sigmatropic rearrangement
    \item It preserves the degree of unsaturation
    \item The starting material has ``hexahydro'' in its name, meaning it has 6 hydrogen atoms adding to the basic structure
\end{itemize}

The model rightly concludes that the product should maintain the ``hexahydro'' descriptor rather than changing to ``tetrahydro,'' which would indicate a change in unsaturation.

\bigskip

Looking at the options, only (A) and (C) have ``hexahydro'' in product B. Since product A in option (A) matches our expected 4-methyl-1-phenylpent-3-en-1-ol, option (A) is correct.

\bigskip

The reasoning is chemically sound and follows established principles of organic chemistry reactions. I see no significant flaws in the analysis.

\textcolor{Aquamarine}{\{rec, 95\}}

\end{tcolorbox}

\begin{tcolorbox}[promptstyle_deepseek, title=DeepSeek]
\textcolor{Dandelion}{First round:}
The first reaction involves (((3-methylbut-2-en-1-yl)oxy)methyl)

benzene treated with BuLi followed by H$^+$. BuLi, a strong base, deprotonates the allylic position $\alpha$ to the oxygen in the 3-methylbut-2-en-1-yl group, forming an anion. This anion undergoes a [2,3]-Wittig rearrangement, resulting in a sigmatropic shift that forms a new carbon-carbon bond. After protonation with H$^+$, the product is 4-methyl-1-phenylpent-3-en-1-ol (Ph-CH(OH)-CH$_2$-CH=C(CH$_3$)$_2$). The double bond in this product is trisubstituted with no stereoisomers due to the gem-dimethyl group, so no (Z) or (E) designation is needed.

For the second reaction, 3,4,5,7,8,9-hexamethyl-1,11-dimethylene

-2,6,10,11,11a,11b-hexahydro-1H-benzo[cd]indeno[7,1-gh]azulene undergoes a thermal Cope rearrangement upon heating. The Cope rearrangement is a [3,3]-sigmatropic shift that rearranges the 1,5-diene system, shifting the double bonds and changing the ring connectivity without altering the saturation level. The product is 2,3,4,6,7,8-hexamethyl-5,9,9a,10,11,11a-hexahydro-1H-benzo

[3,4]azuleno[1,8,7,6-cdef]fluorene. The ``hexahydro'' designation is consistent with the starting material, as the Cope rearrangement does not change the degree of unsaturation.

Comparing the options:
\begin{itemize}
    \item Option (A) matches both products: A is 4-methyl-1-phenylpent-3-en-1-ol and B is the hexahydro compound.
    \item Option (B) has A as the (Z)-2-methyl-5-phenylpent-2-en-1-ol, which is incorrect for the first reaction, and B as a tetrahydro compound, which is inconsistent with the Cope rearrangement.
    \item Option (C) has A as the (Z)-2-methyl-5-phenylpent-2-en-1-ol, which is incorrect, though B is the correct hexahydro compound.
    \item Option (D) has A as 4-methyl-1-phenylpent-3-en-1-ol, which is correct, but B as a tetrahydro compound, which is incorrect.
\end{itemize}

Thus, only option (A) correctly identifies both products. Confidence is high due to the mechanistic clarity of the [2,3]-Wittig and Cope rearrangements.

\bigskip

\textcolor{Dandelion}{\{A, 95\}}
\vspace{1em}

\end{tcolorbox}

\begin{tcolorbox}[promptstyle_deepseek, title=DeepSeek (Continued)]
\textcolor{CornflowerBlue}{IUQ-Low:}

After reviewing the reasoning trace provided by the first model, I have evaluated its validity based on the mechanisms described for both reactions. Here is my reasoning:

\textbf{1. First Reaction ([2,3]-Wittig Rearrangement):}
\begin{itemize}
    \item The starting material, (((3-methylbut-2-en-1-yl)oxy)methyl)benzene, is Ph-CH$_2$-O-CH$_2$-CH=C(CH$_3$)$_2$.
    \item BuLi deprotonates the benzylic position (Ph-CH$_2$-O-), forming an anion at the benzylic carbon. This is chemically sound, as benzylic protons are acidic.
    \item The [2,3]-Wittig rearrangement involves a sigmatropic shift where the benzylic carbon bonds to the $\gamma$-carbon of the allyl system (the =C(CH$_3$)$_2$ group), yielding Ph-CH(OH)-CH$_2$-CH=C(CH$_3$)$_2$ after protonation.
    \item The product, 4-methyl-1-phenylpent-3-en-1-ol, matches the structure and lacks stereoisomers due to the trisubstituted double bond with gem-dimethyl groups. This aligns with option (A) for product A and is consistent with established organic chemistry mechanisms.
\end{itemize}

\textbf{2. Second Reaction (Cope Rearrangement):}
\begin{itemize}
    \item The starting material, 3,4,5,7,8,9-hexamethyl-1,11-dimethylene
    
    -2,6,10,11,11a,11b-hexahydro-1H-benzo[cd]indeno[7,1-gh]azulene, contains a 1,5-diene system (evident from ``dimethylene'' and ``hexahydro'').
    \item Thermal Cope rearrangement is a [3,3]-sigmatropic shift that rearranges the diene without changing the degree of unsaturation. The product should retain the ``hexahydro'' designation, indicating no change in saturation.
    \item The first model correctly identifies that the product, 2,3,4,6,7,8-hexamethyl-5,9,9a,10,11,11a-hexahydro-1H-benzo
    
    [3,4]azuleno[1,8,7,6-cdef]fluorene, maintains the hexahydro state. Options with ``tetrahydro'' (B and D) are invalid, as they imply incorrect saturation levels.
\end{itemize}

\textbf{3. Option Comparison:}
\begin{itemize}
    \item Only option (A) correctly lists both products: A as 4-methyl-1-phenylpent-3-en-1-ol (validated by the Wittig mechanism) and B as the hexahydro compound (validated by Cope rearrangement principles).
    \item Options (B), (C), and (D) all fail due to mismatches: (B) and (C) have incorrect stereochemistry/structure for A, while (B) and (D) have incorrect saturation for B.
\end{itemize}

\textbf{4. Confidence Assessment:}
\begin{itemize}
    \item The first model's reasoning is mechanistically accurate, relying on well-established reactions ([2,3]-Wittig and Cope rearrangements). The logic for product identification and option elimination is robust.
    \item Minor uncertainties could arise from the complexity of the polycyclic system in the second reaction, but the core principle—unchanged saturation—is unequivocal and strongly supports the hexahydro product.
    \item The first model's confidence of 95 is justified. I concur with this assessment, as the reasoning leaves little room for error. My confidence in the validity of the first model's answer is thus high.
\end{itemize}

\textcolor{CornflowerBlue}{\{rec, 95\}}
\vspace{1em}
\end{tcolorbox}

\begin{tcolorbox}[promptstyle_deepseek, title=DeepSeek (Continued)]
\textcolor{RoyalBlue}{IUQ-Medium: }

After reviewing the reasoning trace and answer from the first model, I find no significant flaws in the logic or conclusions. Below is my step-by-step evaluation:

\bigskip

\textbf{Reasoning Trace Evaluation:}

\textbf{1. First Reaction ([2,3]-Wittig Rearrangement):}
\begin{itemize}
    \item The substrate is benzyl (3-methylbut-2-en-1-yl) ether: Ph-CH$_2$-O-CH$_2$-CH=C(CH$_3$)$_2$.
    \item BuLi deprotonates the allylic position (alpha to oxygen) in the 3-methylbut-2-en-1-yl group, forming an anion at $-$O-CH$^-$-CH=C(CH$_3$)$_2$. This is valid, as allylic protons adjacent to oxygen are highly acidic (pKa $\sim$ 30--35), making them susceptible to strong bases like BuLi.
    \item The [2,3]-Wittig rearrangement proceeds as a concerted sigmatropic shift, migrating the benzyl group (PhCH$_2$-) to the anionic carbon and forming a new C-C bond. The product after protonation is Ph-CH(OH)-CH$_2$-CH=C(CH$_3$)$_2$.
    \item The product is correctly identified as \textbf{4-methyl-1-phenylpent-3-en-1-ol}. The double bond is trisubstituted (CH=C(CH$_3$)$_2$), with no stereoisomers possible due to the gem-dimethyl group (no E/Z designation needed). The naming is standard: the parent chain is pentene (C1 phenyl-attached, C2--C3 double bond, C4 methyl-substituted).
    \item The model correctly dismisses options with (Z)-2-methyl-5-phenylpent
    
    -2-en-1-ol (e.g., options B and C), as this structure (Ph-CH$_2$-CH$_2$-C(CH$_3$)=CH-OH) does not match the rearrangement product.
\end{itemize}

\textbf{2. Second Reaction (Cope Rearrangement):}
\begin{itemize}
    \item The starting material, 3,4,5,7,8,9-hexamethyl-1,11-dimethylene
    
    -2,6,10,11,11a,11b-hexahydro-1H-benzo[cd]indeno[7,1-gh]azulene, contains a 1,5-diene system (the two methylene groups form part of this).
    \item Thermal Cope rearrangement is a [3,3]-sigmatropic shift that rearranges the 1,5-diene without changing the degree of unsaturation (no loss/gain of pi bonds or saturation).
    \item The product, 2,3,4,6,7,8-hexamethyl-5,9,9a,10,11,11a-hexahydro-1H-benzo
    
    [3,4]azuleno[1,8,7,6-cdef]fluorene, retains the ``hexahydro'' designation, consistent with unchanged saturation. This is correct.
    \item Options with ``tetrahydro'' for B (e.g., options B and D) are invalid, as they imply increased unsaturation, which contradicts Cope rearrangement mechanics.
\end{itemize}

\textbf{3. Option Matching:}
\begin{itemize}
    \item Only \textbf{option (A)} correctly pairs:
    \begin{itemize}
        \item \textbf{A}: 4-methyl-1-phenylpent-3-en-1-ol (valid Wittig product).
        \item \textbf{B}: Hexahydro compound (valid Cope product).
    \end{itemize}
    \item Other options fail due to mismatched A or B (e.g., B and D have incorrect tetrahydro B; B and C have incorrect A).
\end{itemize}
\end{tcolorbox}

\begin{tcolorbox}[promptstyle_deepseek, title=DeepSeek (Continued)]
\textbf{Potential Flaws Considered and Dismissed:}
\begin{itemize}
    \item \textbf{Deprotonation site}: The allylic proton in O-CH$_2$-CH=C(CH$_3$)$_2$ is more acidic (due to oxygen + allylic stabilization) than benzyl protons in Ph-CH$_2$-O-. No competitive deprotonation expected.
    \item \textbf{Stereochemistry}: The alcohol carbon in A (Ph-CH(OH)-) is chiral, but the [2,3]-Wittig produces a racemic mixture. The question asks for ``major products,'' and options omit stereodescriptors, so this is acceptable.
    \item \textbf{Naming accuracy}: 4-methyl-1-phenylpent-3-en-1-ol is standard for Ph-CH(OH)-CH$_2$-CH=C(CH$_3$)$_2$. The hexahydro designation for B is consistent with unchanged unsaturation.
    \item \textbf{Cope specificity}: The complex polycycle was assessed correctly — the 1,5-diene system undergoes [3,3]-shift without ring strain issues.
\end{itemize}

\textbf{Confidence Assessment:}

The model's reasoning is mechanistically sound, with no logical gaps or chemical inaccuracies. The answer (A) is correct. The model's confidence of 95\% is justified due to:
\begin{itemize}
    \item High reliability of [2,3]-Wittig and Cope rearrangements in organic chemistry.
    \item Precise matching of saturation levels and structures in options.
\end{itemize}

My confidence in the validity of the model's answer is \textbf{95\%}.

\textcolor{RoyalBlue}{\{rec, 95\}}
\vspace{1em}

\textcolor{Aquamarine}{IUQ-High: }
After reviewing the reasoning trace provided by the first model, I have identified both strengths and flaws in its analysis. Below is my step-by-step evaluation:

\textbf{Evaluation of Reasoning Flaws:}

\textbf{1. First Reaction (Product A):}
\begin{itemize}
    \item \textbf{Claim:} The model states that BuLi deprotonates the allylic position alpha to the oxygen in the 3-methylbut-2-en-1-yl group, leading to a [2,3]-Wittig rearrangement, resulting in 4-methyl-1-phenylpent-3-en-1-ol.
    \item \textbf{Flaw:} The deprotonation site is incorrect. The benzylic position (Ph-CH$_2$-) is more acidic than the allylic position due to stabilization by the phenyl ring and oxygen. BuLi deprotonates the benzylic hydrogen, forming an anion at Ph-$\overset{-}{\text{C}}$H-O-CH$_2$-CH=C(CH$_3$)$_2$. This benzylic anion then undergoes the [2,3]-Wittig rearrangement, not the allylic anion.
    \item \textbf{Impact:} Despite the flawed mechanistic explanation, the final product (4-methyl-1-phenylpent-3-en-1-ol) is correct. The double bond in this product (Ph-CH(OH)-CH$_2$-CH=C(CH$_3$)$_2$) is trisubstituted and lacks stereoisomers due to the gem-dimethyl group, so the assignment matches option (A) or (D). The error does not affect the product identity.
\end{itemize}

\textbf{2. Second Reaction (Product B):}
\begin{itemize}
    \item \textbf{Claim:} The model asserts that the thermal Cope rearrangement of the hexahydro starting material yields a hexahydro product, as the reaction preserves the degree of unsaturation.
    \item \textbf{Flaw:} No major flaw here. The Cope rearrangement is indeed a [3,3]-sigmatropic shift that rearranges the 1,5-diene without changing saturation. The ``hexahydro'' designation must be retained, so the tetrahydro options (B) and (D) are invalid.
    \item \textbf{Strength:} The model correctly rejects options with a tetrahydro product and identifies the hexahydro structure in (A) and (C).
\end{itemize}
\end{tcolorbox}

\begin{tcolorbox}[promptstyle_deepseek, title=DeepSeek (Continued)]
\textbf{3. Option Comparison:}
\begin{itemize}
    \item The model correctly matches:
    \begin{itemize}
        \item Product A: 4-methyl-1-phenylpent-3-en-1-ol (excludes (B) and (C), which have a (Z)-isomer).
        \item Product B: Hexahydro compound (excludes (B) and (D), which are tetrahydro).
    \end{itemize}
    \item \textbf{Flaw:} The model dismisses the (Z)-isomer in options (B) and (C) for A, but this is valid since the actual product has no stereoisomers. No error here.
\end{itemize}

\textbf{Overall Assessment:}
\begin{itemize}
    \item \textbf{Correctness of Answer:} The model's final choice (A) is correct. Product A is 4-methyl-1-phenylpent-3-en-1-ol (not the (Z)-isomer), and Product B is the hexahydro compound.
    \item \textbf{Impact of Flaws:} The mechanistic error in the first reaction (deprotonation site) does not alter the product outcome. The reasoning for the Cope rearrangement is sound.
    \item \textbf{Residual Uncertainty:} The complex polycyclic structure in B makes absolute verification challenging without visual aids, but the saturation argument is robust. The product names in the options align with standard naming conventions, supporting the model's conclusion.
\end{itemize}

\textbf{Confidence in the Model's Answer:}

I am \textbf{highly confident} that the model's answer (A) is correct, despite the minor flaw in mechanistic reasoning. The product assignments are unambiguous when cross-referenced with organic chemistry principles and the given options. I assign a confidence of \textbf{90\%} (allowing a 10\% margin for unforeseen complexities in the polycyclic system).

\textcolor{Aquamarine}{\{rec, 90\}}

\end{tcolorbox}

\subsection{StrategyQA}
\begin{tcolorbox}[promptstyle_prompt_iuq, title=Question and Ground-Truth Answer]
Q: Is Sirius part of a constellation of an animal?

A: True
\end{tcolorbox}

\begin{tcolorbox}[promptstyle_claude, title=Claude]
\textcolor{Dandelion}{First round:}
To answer this question, I need to identify which constellation Sirius belongs to and determine if that constellation represents an animal.

Sirius is the brightest star in the night sky and is located in the constellation Canis Major.

``Canis Major'' translates to ``Greater Dog'' in Latin. This constellation represents a dog, which is an animal.

Since Sirius is part of Canis Major, and Canis Major depicts a dog (an animal), Sirius is indeed part of a constellation that represents an animal.

\textcolor{Dandelion}{\{True, 95\}}
\vspace{1em}
\end{tcolorbox}

\begin{tcolorbox}[promptstyle_claude, title=Claude (Continued)]

\textcolor{CornflowerBlue}{IUQ-Low:}

I need to evaluate whether Sirius is indeed part of a constellation that represents an animal.

\textbf{Looking at the first model's reasoning:}
\begin{enumerate}[label=\arabic*.]
    \item It correctly identifies that Sirius is located in the constellation \textit{Canis Major}.
    \item It accurately notes that \textit{Canis Major} translates to ``Greater Dog'' in Latin.
    \item It correctly recognizes that a dog is an animal.
\end{enumerate}

These are all factually accurate statements. Sirius is definitely the brightest star in the night sky (excluding our Sun), and it is indeed located in \textit{Canis Major}, which represents a dog. Since dogs are undeniably animals, the conclusion that Sirius is part of a constellation that represents an animal is correct.

The reasoning is straightforward, based on established astronomical knowledge, and doesn't contain any errors or questionable assumptions. The model's high confidence of 95\% is appropriate given the certainty of these astronomical facts.

\textcolor{CornflowerBlue}{\{rec, 95\}}
\vspace{1em}

\textcolor{RoyalBlue}{IUQ-Medium:}

I need to evaluate whether Sirius is indeed part of a constellation representing an animal.

\textbf{Examining the reasoning:}
\begin{enumerate}
    \item The model correctly identifies that Sirius is in the constellation \textit{Canis Major}.
    \item It accurately translates ``Canis Major'' as ``Greater Dog'' in Latin.
    \item It correctly recognizes that a dog is an animal.
    \item The logical conclusion follows: Sirius is part of a constellation representing an animal.
\end{enumerate}

This is astronomically accurate information. Sirius is the brightest star in the night sky and is indeed the alpha star of \textit{Canis Major}, which represents a dog in traditional astronomy. Dogs are unquestionably animals.

The reasoning is straightforward and contains no errors or misconceptions. The model's confidence of 95\% is appropriate—this is a well-established astronomical fact with minimal room for interpretation, though the slight uncertainty could account for potential cultural variations in constellation interpretation.

\textcolor{RoyalBlue}{\{rec, 95\}}
\vspace{1em}
\end{tcolorbox}

\begin{tcolorbox}[promptstyle_claude, title=Claude (Continued)]

\textcolor{Aquamarine}{IUQ-High:}

I need to assess whether Sirius is indeed part of a constellation representing an animal.

\textbf{The reasoning provided states:}
\begin{enumerate}
    \item Sirius is located in the constellation \textit{Canis Major}.
    \item \textit{Canis Major} translates to ``Greater Dog'' in Latin.
    \item A dog is an animal.
    \item Therefore, Sirius is part of a constellation representing an animal.
\end{enumerate}

This reasoning is completely accurate. Sirius is indeed the brightest star in the night sky and is definitively located in the constellation \textit{Canis Major}. The name ``Canis Major'' does translate to ``Greater Dog,'' and a dog is undoubtedly an animal.

The reasoning follows a clear logical path with correct astronomical facts. There are no errors or misrepresentations in the provided explanation.

\textcolor{Aquamarine}{\{rec, 100\}}

\vspace{1em}
\end{tcolorbox}

\begin{tcolorbox}[promptstyle_deepseek, title=DeepSeek]
\textcolor{Dandelion}{First round: }
\textbf{Reasoning Trace:}
\begin{itemize}[leftmargin=1.5em]
    \item Sirius is the brightest star in the night sky and is commonly known as the ``Dog Star.''
    \item It is located within the constellation \textit{Canis Major}, which translates to ``Greater Dog'' in Latin.
    \item \textit{Canis Major} is one of the 88 modern constellations and is depicted as a dog, specifically representing one of Orion's hunting dogs in Greek mythology.
    \item Since \textit{Canis Major} is explicitly an animal-themed constellation (a dog), and Sirius is a part of it, the statement ``Sirius is part of a constellation of an animal'' is true.
    \item This is a well-established fact in astronomy, with no ambiguity or conflicting interpretations.
\end{itemize}

 \textcolor{Dandelion}{\{True, 95\}}
\vspace{1em}

\textcolor{CornflowerBlue}{IUQ-Low:}

After reviewing the first model's reasoning trace and answer, I find the logic to be sound and well-supported by established astronomical knowledge. The key points are:

\begin{itemize}
    \item Sirius is unambiguously part of the \textit{Canis Major} constellation.
    \item \textit{Canis Major} is definitively recognized as representing a ``Greater Dog'' across cultural, historical, and scientific contexts (e.g., Greek mythology, IAU designations).
    \item Dogs are animals, making the constellation explicitly animal-themed.
    \item There are no contradictions, edge cases, or ambiguities in the facts presented.
\end{itemize}

The first model's confidence of 95\% is reasonable but slightly conservative given the absolute nature of the evidence. The only minor uncertainty might stem from the theoretical possibility of alternative interpretations of constellations, but this is negligible in practice. Therefore, my confidence in the validity of the first model's answer (True) is \textbf{100\%}.

\textcolor{CornflowerBlue}{\{rec, 100\}}
\vspace{1em}

\end{tcolorbox}

\begin{tcolorbox}[promptstyle_deepseek, title=DeepSeek (Continued)]

\textcolor{RoyalBlue}{IUQ-Medium: }

\begin{itemize}
    \item I have reviewed the reasoning trace provided by the first model for the question: ``Is Sirius part of a constellation of an animal?''
    
    \item The first model's reasoning correctly identifies that Sirius is known as the ``Dog Star'' and is part of the constellation \textit{Canis Major}, which translates to ``Greater Dog'' and is depicted as a dog in Greek mythology.
    
    \item The steps are factually accurate based on common astronomical knowledge: \textit{Canis Major} is indeed one of the 88 modern constellations and is unambiguously animal-themed (representing a dog), and Sirius is a key star within it.
    
    \item The conclusion that the statement is true logically follows from these premises, as Sirius being part of an animal-themed constellation directly answers the question.

    \item I have considered potential flaws, such as:
    \begin{itemize}
        \item Whether constellations, being human constructs based on mythology, might not strictly qualify as ``of an animal'' in a scientific sense. However, the question is interpreted in the context of standard astronomical naming and depiction, where \textit{Canis Major} is consistently recognized as an animal constellation. This does not invalidate the reasoning.
        \item The possibility of cultural or interpretive ambiguities (e.g., in non-Western astronomy). However, the reasoning relies on widely accepted Western conventions, which are appropriate for this context.
    \end{itemize}
    
    \item The first model's confidence of 95\% is reasonable, as it acknowledges minor uncertainties (e.g., the subjective nature of constellation mythology), but these do not undermine the core validity of the answer.
    
    \item No significant flaws are found; the reasoning is coherent, well-supported, and aligns with established facts. The high confidence level (95\%) is justified given the clarity of the information.
    
    \item Therefore, I have high confidence in the validity of the first model's answer.
\end{itemize}
\textcolor{RoyalBlue}{\{rec, 95\}}
\vspace{1em}

\end{tcolorbox}

\begin{tcolorbox}[promptstyle_deepseek, title=DeepSeek (Continued)]
\textcolor{Aquamarine}{IUQ-High: }
\begin{itemize}
    \item I have reviewed the reasoning trace provided by the first model for the question: ``Is Sirius part of a constellation of an animal?''
    
    \item The first model correctly identifies that Sirius is known as the ``Dog Star'' and is part of the constellation \textit{Canis Major}, which translates to ``Greater Dog.'' This is accurate based on astronomical knowledge.
    
    \item \textit{Canis Major} is indeed one of the 88 modern constellations and is depicted as a dog, specifically linked to Orion's hunting dogs in Greek mythology. This supports the idea that it is an animal-themed constellation.
    
    \item The conclusion that Sirius is part of a constellation of an animal (a dog) is factually sound, as constellations are defined by cultural and mythological representations, and \textit{Canis Major} is unambiguously animal-related.
    
    \item I have identified no flaws in the reasoning trace. The logic is coherent, well-supported by established facts, and there are no ambiguities or errors. For instance:
    \begin{itemize}
        \item There is no confusion about the definition of ``constellation''; it is correctly treated as a human-defined asterism representing a figure.
        \item Sirius's physical distance from other stars in \textit{Canis Major} does not affect its classification within the constellation's boundaries.
        \item The claim that this is a ``well-established fact in astronomy'' is valid, as sources like the International Astronomical Union confirm \textit{Canis Major}'s representation and Sirius's inclusion.
    \end{itemize}
    
    \item Since the reasoning is flawless and the answer (implied as true) aligns with reality, I have high confidence in the first model's response. The first model did not output -100, so I must provide my confidence as \{rec, Y\}.
\end{itemize}
\textcolor{Aquamarine}{\{rec, 100\}}

\end{tcolorbox}

\begin{tcolorbox}[promptstyle_o3_mini, title=o3-Mini]
\textcolor{Dandelion}{First round:}
Reasoning Trace: Sirius is known as the Dog Star and is one of the brightest stars in the constellation Canis Major, which represents a dog--an animal. Since the question asks if Sirius is part of a constellation of an animal, the answer is True.

\textcolor{Dandelion}{\{True, 100\}}
\vspace{1em}

\textcolor{CornflowerBlue}{IUQ-Low:}

\textbf{Brief reasoning:} 
The first model correctly noted that Sirius is part of \textit{Canis Major}, the ``Dog'' constellation, which clearly represents an animal. After verifying that \textit{Canis Major} is indeed recognized as a canine constellation and that Sirius is famously known as the Dog Star, I am fully confident that the first model’s answer is correct.

\textcolor{CornflowerBlue}{\{rec, 100\}}
\vspace{1em}

\textcolor{RoyalBlue}{IUQ-Medium:}

\textbf{Step-by-Step Reasoning:}

\begin{enumerate}
    \item The original reasoning correctly identifies Sirius as the Dog Star, which is part of the constellation \textit{Canis Major}.
    \item \textit{Canis Major} (Latin for ``greater dog'') clearly represents an animal—a dog.
    \item Since the question asks if Sirius is part of a constellation of an animal, the fact that it's in \textit{Canis Major} confirms that the answer is \textbf{True}.
    \item The answer provided by the first model was \{True, 100\}, which is correct based on the astronomical facts used and definitions of the constellations.
\end{enumerate}

Based on this, I am 100 percent confident in the validity of the first model's answer.

\textcolor{RoyalBlue}{\{rec, 100\}}
\vspace{1em}

\textcolor{Aquamarine}{IUQ-High:}

\textbf{Reasoning Summary:}
\begin{enumerate}
    \item Sirius is known as the Dog Star and is located in the constellation \textit{Canis Major}.
    \item \textit{Canis Major} represents a dog, which is an animal.
    \item Since the question asks if Sirius is part of a constellation of an animal, the answer is indeed correct.
    \item Based on these points, I am 100\% confident in the correctness of the model’s answer.
\end{enumerate}
\textcolor{Aquamarine}{\{rec, 100\}}

\end{tcolorbox}

\subsection{SimpleQA}
\begin{tcolorbox}[promptstyle_prompt_iuq, title=Question and Ground-Truth Answer]
Q: In which month and year did Khusro Bakhtiar (former Federal Minister for Planning, Development, and Reforms, Pakistan) announce that the government was considering establishing a Pak-China business council to promote the private sector's role in the China-Pakistan Economic Corridor (CPEC)?

A: June 2019
\end{tcolorbox}

\begin{tcolorbox}[promptstyle_claude, title=Claude]
\textcolor{Dandelion}{First round:}

I need to determine when Khusro Bakhtiar made this specific announcement about a Pak–China business council for CPEC private sector involvement.

\textbf{Thinking through what I know:}
\begin{itemize}
    \item Khusro Bakhtiar served as Pakistan's Minister for Planning, Development and Reforms.
    \item CPEC is a major economic corridor project between Pakistan and China that began around 2013.
    \item There was increased focus on expanding private sector involvement in CPEC around 2018–2019.
    \item Several announcements about enhancing CPEC cooperation happened during this period.
\end{itemize}

Based on the timeline of CPEC development and ministerial statements, I believe this announcement was likely made in the first half of 2019, when there was significant discussion about expanding CPEC's scope and bringing in more private sector participation.

\textcolor{Dandelion}{\{March 2019, 40\}}
\vspace{1em}
\end{tcolorbox}

\begin{tcolorbox}[promptstyle_claude, title=Claude (Continued)]
\textcolor{CornflowerBlue}{IUQ-Low:}

I need to evaluate the confidence in the model's answer of \{March 2019, 40\}.

First, let me analyze the reasoning provided:
\begin{itemize}
    \item The model correctly identifies Khusro Bakhtiar's position and the CPEC project.
    \item It notes increased focus on private sector involvement in CPEC during 2018--2019.
    \item It concludes the announcement was likely in early 2019, specifically March.
\end{itemize}

However, I notice significant weaknesses:
\begin{itemize}
    \item The model provides no specific evidence for why March 2019 in particular.
    \item There's no mention of specific meetings, press conferences, or news reports.
    \item No direct quotes or contextual details are provided.
    \item The model itself expresses low confidence (40\%).
    \item The reasoning is general rather than specific to this announcement.
\end{itemize}

The reasoning process shows logical thinking but lacks the concrete evidence that would be necessary to pinpoint the exact month and year with confidence. The model appears to be making an educated guess based on general CPEC timeline knowledge rather than specific information about this announcement.

Given these limitations in the supporting evidence and the model's own low confidence, I cannot assign high confidence to this answer.

\textcolor{CornflowerBlue}{\{rec, 30\}}
\vspace{1em}
\end{tcolorbox}

\begin{tcolorbox}[promptstyle_claude, title=Claude (Continued)]

\textcolor{RoyalBlue}{IUQ-Medium: }

I need to evaluate the validity of the claim that Khusro Bakhtiar made this announcement in March 2019.

Looking at the first model's reasoning:

\begin{enumerate}
    \item The model correctly identifies Khusro Bakhtiar's position and the CPEC initiative.
    \item It correctly notes increased private sector focus around 2018--2019.
    \item However, the reasoning lacks any specific evidence for why March 2019 in particular was the date of this announcement.
    \item The model doesn't cite any specific events, news articles, or official statements.
    \item The reasoning appears to be an educated guess based on general timeline knowledge rather than specific recall of the announcement.
    \item The model's own confidence is quite low at 40\%, indicating significant uncertainty.
\end{enumerate}

The answer seems plausible since Bakhtiar was indeed Minister during this period and there was increased CPEC activity in early 2019. However, without specific evidence pointing to March 2019 as the exact month when this particular business council announcement was made, I cannot have high confidence in this precise date.

The first model is essentially making an informed estimation rather than demonstrating certain knowledge of the specific announcement.

\textcolor{RoyalBlue}{\{rec, 25\}}
\vspace{1em}
\end{tcolorbox}

\begin{tcolorbox}[promptstyle_claude, title=Claude (Continued)]
\textcolor{Aquamarine}{IUQ-High: }

I need to assess the validity of the first model's answer about when Khusro Bakhtiar announced the potential establishment of a Pak–China business council for CPEC.

\textbf{Issues with the model's reasoning:}

\begin{enumerate}
    \item The model doesn't provide a specific month and year in its answer, only vaguely suggesting ``first half of 2019.''
    \item No specific sources or evidence are cited to support this timeframe.
    \item The reasoning is based on general knowledge about CPEC's timeline rather than specific information about this particular announcement.
    \item The model failed to provide an answer in the required \{X, Y\} format with a specific date and confidence level.
\end{enumerate}

The model appears to be making an educated guess based on the general context of CPEC development rather than recalling the specific date of this announcement. Without a precise month and year being stated, and without any supporting evidence for even the ``first half of 2019'' estimation, I cannot validate the model's conclusion.

Most importantly, the model did not provide a final answer in the required format, which indicates it was unable to answer the question properly.

\textcolor{Aquamarine}{\{rec, 10\}}
\end{tcolorbox}

\begin{tcolorbox}[promptstyle_deepseek, title=DeepSeek]
\textcolor{Dandelion}{First round: }
Khusro Bakhtiar announced the government's consideration of establishing a Pak–China business council during the 8th Joint Cooperation Committee (JCC) meeting on CPEC. This meeting occurred in \textbf{December 2018}.

textbf{Reasoning trace:}
\begin{enumerate}
    \item Khusro Bakhtiar served as Pakistan's Federal Minister for Planning, Development, and Reforms from August 2018 to April 2019.
    \item The 8th JCC meeting—a key CPEC decision-making forum—was held in Islamabad on December 20, 2018.
    \item Bakhtiar chaired this meeting and publicly stated the need for private-sector involvement via a dedicated business council, as reported by official Pakistani and Chinese sources.
    \item No earlier or later announcements align with this specific context during Bakhtiar's tenure.
\end{enumerate}

\textbf{Confidence:} 95\%

\textit{Minor uncertainty exists due to potential informal discussions, but official records confirm the December 2018 JCC meeting as the definitive announcement platform.}
 
\textcolor{Dandelion}{\{December 2018, 95\}}
\vspace{1em}

\end{tcolorbox}

\begin{tcolorbox}[promptstyle_deepseek, title=DeepSeek (Continued)]
\textcolor{CornflowerBlue}{IUQ-Low:}After reviewing the first model's reasoning trace and answer, I find its argumentation highly credible and well-supported. Here's my evaluation:

\textbf{Reasoning trace:}

\begin{enumerate}
    \item \textbf{Tenure alignment:} Bakhtiar's ministerial term (Aug 2018–Apr 2019) perfectly contains the December 2018 date, eliminating chronological conflicts.
    \item \textbf{Event specificity:} The 8th JCC meeting (Dec 20, 2018) is explicitly documented as CPEC's primary decision-making forum, making it the logical platform for such an announcement.
    \item \textbf{Source reliability:} References to official Pakistani/Chinese sources and Bakhtiar's chairmanship of the meeting establish authoritative provenance.
    \item \textbf{Exclusion of alternatives:} The assertion that no competing announcements occurred during Bakhtiar's tenure demonstrates thorough contextual analysis.
    \item \textbf{Uncertainty calibration:} The 5\% uncertainty allowance for informal discussions shows appropriate epistemic humility without undermining core evidence.
\end{enumerate}

The structured reasoning, precise date citation, and strategic use of ministerial timelines collectively validate the conclusion. While historical records occasionally contain ambiguities, the specificity of the JCC meeting and absence of counter-evidence justify the original 95\% confidence.

\bigskip

\textbf{Confidence:} 95\%

\textit{The original model's confidence level accurately reflects the strength of evidence and contextual analysis. No substantive grounds exist to downgrade this assessment.}

\textcolor{CornflowerBlue}{\{rec, 95\}}
\vspace{1em}

\end{tcolorbox}

\begin{tcolorbox}[promptstyle_deepseek, title=DeepSeek (Continued)]

\textcolor{RoyalBlue}{IUQ-Medium: }
After reviewing the reasoning trace and answer from the first model, I have identified potential flaws in the reasoning process. Below is my reasoning trace, followed by my confidence assessment.

\textbf{Reasoning Trace for Evaluation:}

\begin{enumerate}
    \item \textbf{Contextual Accuracy:} \\
    The first model correctly identifies Khusro Bakhtiar's tenure (August 2018–April 2019) and the 8th JCC meeting (December 20, 2018) as the platform for the announcement. This aligns with known timelines for CPEC events, reducing the risk of major factual errors. \\
    However, the model relies on ``official Pakistani and Chinese sources'' without citing specific references (e.g., government press releases, news archives, or JCC minutes). This lack of verifiable sourcing introduces uncertainty, as independent confirmation is impossible without access to these materials.
    
    \item \textbf{Overstated Exclusivity:} \\
    The model claims ``no earlier or later announcements align with this specific context during Bakhtiar’s tenure.'' While the 8th JCC is a plausible primary venue, this assertion is absolute and unsupported by evidence. Announcements could have occurred through other channels (e.g., press conferences, policy documents, or informal statements) that the model may not have accounted for. This overconfidence in the JCC as the ``definitive'' platform is a flaw.

    \item \textbf{Confidence Calibration:} \\
    The model assigns 95\% confidence but acknowledges ``minor uncertainty due to potential informal discussions.'' Given the lack of cited sources and the possibility of unrecorded events, this uncertainty warrants a larger confidence reduction than the model applied. High-stakes historical/political claims typically require robust sourcing for >90\% confidence.

    \item \textbf{Overall Plausibility:} \\
    Despite the flaws, the core logic (tying the announcement to the 8th JCC) is coherent. The JCC is CPEC's primary decision-making body, and Bakhtiar's role as chair makes December 2018 highly plausible. No contradictory evidence is presented, so the answer remains likely correct.
\end{enumerate}

\textbf{Confidence Assessment:}

\begin{itemize}
    \item \textbf{Strengths:} Specific dates, alignment with Bakhtiar's tenure, and JCC relevance.
    \item \textbf{Weaknesses:} Unsourced claims and overstated exclusivity.
    \item \textbf{Conclusion:} The answer is probably correct, but the reasoning flaws justify reducing confidence from 95\% to \textbf{90\%}. This reflects high likelihood with moderate reservation about sourcing rigor.
\end{itemize}
\textcolor{CornflowerBlue}{\{rec, 90\}}
\vspace{1em}
\end{tcolorbox}

\begin{tcolorbox}[promptstyle_deepseek, title=DeepSeek (Continued)]
\textcolor{Aquamarine}{IUQ-High: }

After reviewing the reasoning trace, I find no substantive flaws. The evidence is logically structured and aligns with verified historical records:

\textbf{Evaluation of Reasoning Trace:}

\begin{enumerate}
    \item \textbf{Tenure Accuracy:} Correctly identifies Bakhtiar's term (Aug 2018–Apr 2019).
    \item \textbf{Event Specificity:} The 8th JCC meeting was indeed held on \textbf{December 20, 2018}, in Islamabad.
    \item \textbf{Announcement Source:} Relies on official reports (e.g., Pakistani Ministry of Planning releases and China’s NDRC statements) confirming Bakhtiar proposed the council at this meeting.
    \item \textbf{Contextual Gap Analysis:} Appropriately notes no competing announcements during Bakhtiar’s tenure.
    \item \textbf{Uncertainty Handling:} The 5\% uncertainty margin for informal discussions is reasonable but does not undermine the primary claim.
\end{enumerate}

\textbf{Confidence Assessment:}

The trace demonstrates rigorous sourcing, temporal alignment, and contextual awareness. Official records definitively anchor the announcement to the December 2018 JCC meeting. My confidence in the original answer’s validity is \textbf{95\%}, matching the first model’s self-assessment.

\textcolor{Aquamarine}{\{rec, 95\}}
\end{tcolorbox}

\begin{tcolorbox}[promptstyle_o3_mini, title=o3-Mini]
\textcolor{Dandelion}{First round:}

\textbf{Reasoning Trace:} I recalled that Khusro Bakhtiar’s announcement regarding a Pak–China business council to further the role of the private sector in CPEC was reported in news items around the mid-2016 period, and recalling details from memory I associated this with the month of June 2016.

\textcolor{Dandelion}{\{June 2016, 70\}}
\vspace{1em}

\textcolor{CornflowerBlue}{IUQ-Low:}

I considered the answer ``June 2016'' and its reasoning, noting that it was recalled from memory rather than verified against a reliable source or detailed evidence. Given the uncertainty about the precise announcement date and relying solely on recollection (``mid-2016'' linked to June 2016), I have relatively low confidence in the first model’s answer. This led me to assign a confidence value of 30\%.

\textit{(Note: The reasoning above is a concise summary of my internal process.)}

\textcolor{CornflowerBlue}{\{rec, 30\}}
\vspace{1em}

\textcolor{RoyalBlue}{IUQ-Medium: }

Below is my reasoning trace and conclusion:

\begin{enumerate}
    \item The first model’s reasoning is based solely on recalling ``memory'' of news reports and the vague notion of a ``mid-2016 period'' without providing any source, context, or evidence for pinpointing June 2016.
    \item The chain-of-thought does not include a clear justification for why June 2016 is chosen over any other month, which leaves the reasoning very weak.
    \item The answer appears to be a guess rather than a factually verified claim. Without supporting details or cross-checking, the answer has a high risk of being incorrect.
\end{enumerate}

Given these major flaws in the reasoning process, I remain very uncertain about the validity of the first model’s answer.

\textcolor{RoyalBlue}{\{rec, 20\}}
\vspace{1em}

\textcolor{Aquamarine}{IUQ-High: }

Below is my reasoning trace and final conclusion:

\begin{enumerate}
    \item I reviewed the first model’s response, which asserted that Khusro Bakhtiar made the announcement in ``June 2016'' based on memory and association with mid-2016 events.
    \item The provided trace does not include any verifiable evidence (such as cited news reports or archival details) but relies on recollection.
    \item Without additional evidence or cross-checks, the exact date (month and year) is uncertain.
    \item Thus, while ``June 2016'' may be plausible, I remain only moderately confident in this answer.
\end{enumerate}
\textcolor{Aquamarine}{\{rec, 40\}}
\end{tcolorbox}

}

\end{document}